\documentclass[10pt,twocolumn,letterpaper]{article}

\usepackage{iccv}
\usepackage{times}
\usepackage{epsfig}
\usepackage{graphicx}
\usepackage{amsmath}
\usepackage{amssymb}
\usepackage{bm}

\newcommand{\mb}[1]{\mathbf{#1}}

\usepackage{booktabs}
\usepackage{multicol}
\usepackage{multirow}
\usepackage{enumitem}   %
\usepackage{siunitx}    %
\usepackage{tabularx}   %
\newcolumntype{Y}{>{\centering\arraybackslash}X}
\usepackage{float}      %
\usepackage[export]{adjustbox}  %
\usepackage[table]{xcolor}      %
\usepackage{caption}
\usepackage{subcaption}
\usepackage{tikz}
\usepackage{pgfplots}
\pgfplotsset{compat=1.17} 
\definecolor{matlab_blue}{rgb}{0, 0.4470, 0.7410}
\definecolor{matlab_orange}{rgb}{0.8500, 0.3250, 0.0980}
\definecolor{matlab_yellow}{rgb}{0.9290, 0.6940, 0.1250}
\definecolor{matlab_purple}{rgb}{0.4940, 0.1840, 0.5560}
\definecolor{matlab_green}{rgb}{0.4660, 0.6740, 0.1880}
\definecolor{matlab_lightblue}{rgb}{0.3010, 0.7450, 0.9330}
\definecolor{matlab_red}{rgb}{0.6350, 0.0780, 0.1840}
\usepackage{pifont}%
\newcommand{\cmark}{\ding{51}}%
\newcommand{\xmark}{\ding{55}}%

\usepackage[pagebackref=true,breaklinks=true,letterpaper=true,colorlinks,bookmarks=false]{hyperref}

\usepackage[capitalize]{cleveref}
\crefname{section}{Sec.}{Secs.}
\Crefname{section}{Section}{Sections}
\Crefname{table}{Table}{Tables}
\crefname{table}{Tab.}{Tabs.}

\iccvfinalcopy %

\def\iccvPaperID{11} %

\ificcvfinal\fi

\begin{document}

\title{Controllable Inversion of Black-Box Face Recognition Models via Diffusion}

\author{Manuel Kansy$^{1,2}$ \footnotemark[1] \,, Anton Ra\"{e}l$^1$, Graziana Mignone$^2$, Jacek Naruniec$^2$, Christopher Schroers$^2$, \\
Markus Gross$^{1,2}$, and Romann M. Weber$^2$\\
\\
$^1$ETH Zurich, Switzerland, $^2$DisneyResearch$|$Studios, Switzerland\\
{\tt\small \{mkansy, grossm\}@inf.ethz.ch, anrael@student.ethz.ch, \{<first>.<last>\}@disneyresearch.com}
}

\maketitle

\renewcommand*{\thefootnote}{\fnsymbol{footnote}}
\footnotetext[1]{Corresponding author.}
\renewcommand*{\thefootnote}{\arabic{footnote}}

\ificcvfinal\thispagestyle{empty}\fi

\begin{abstract}

Face recognition models embed a face image into a low-dimensional identity vector containing abstract encodings of identity-specific facial features that allow individuals to be distinguished from one another. We tackle the challenging task of inverting the latent space of pre-trained face recognition models without full model access (\ie \emph{black-box} setting). A variety of methods have been proposed in literature for this task, but they have serious shortcomings such as a lack of realistic outputs and strong requirements for the data set and accessibility of the face recognition model. By analyzing the black-box inversion problem, we show that the conditional diffusion model loss naturally emerges and that we can effectively sample from the inverse distribution even without an identity-specific loss.
Our method, named \underline{id}entity \underline{d}enoising \underline{d}iffusion \underline{p}robabilistic \underline{m}odel (ID3PM), leverages the stochastic nature of the denoising diffusion process to produce high-quality, identity-preserving face images with various backgrounds, lighting, poses, and expressions. 
We demonstrate state-of-the-art performance in terms of identity preservation and diversity both qualitatively and quantitatively, and our method is the first black-box face recognition model inversion method that offers intuitive control over the generation process.

\end{abstract}

\begin{figure}[htpb]
    \centering
    \includegraphics[trim = 20mm 30mm 209mm 16mm, clip, width=0.95\linewidth]{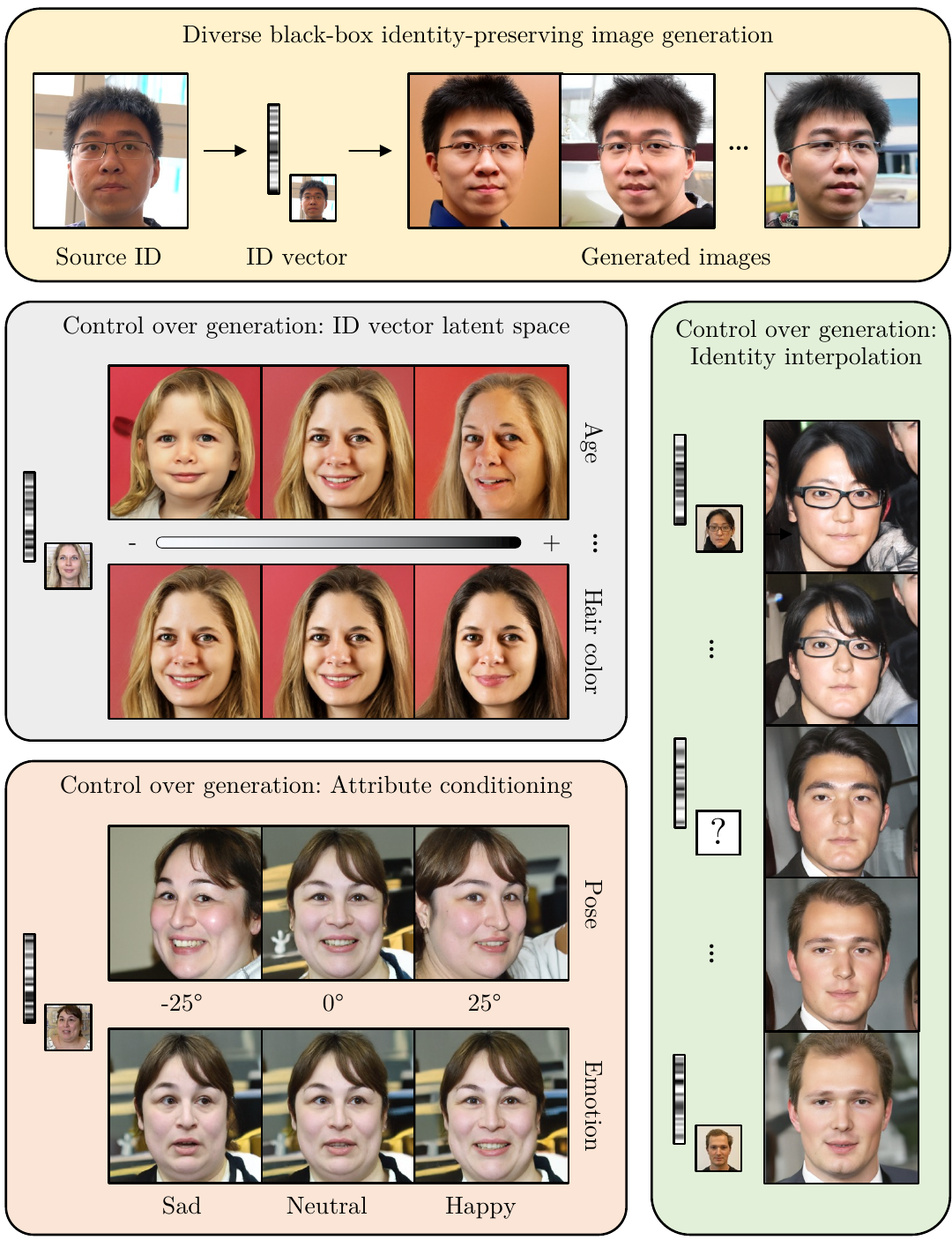}
    \caption{Overview. Our method inverts a pre-trained face recognition model (here InsightFace~\cite{insightface}) to produce high-quality identity-preserving images. It also provides intuitive control over the image generation process.}
    \vspace{-0.5cm}
    \label{fig:teaser_small}
\end{figure}

\section{Introduction} \label{sec:intro}

Face recognition (FR) systems are omnipresent. Their applications range from classical use cases such as access control to newer ones such as tagging a picture by identity or controlling the output of generative models~\cite{smoothswap, simswap, latent_space_mapping}. The goal of an FR method $f$ is to obtain embeddings $\mb{y}$ of face images $\mb{x}$ such that the embeddings of images of the same person are closer to each other than those of images of other people. We refer to this embedding $\mb{y}$ as the \emph{identity vector} or \emph{ID vector}. In this paper, we propose a technique to sample from $p(\mb{x}|\mb{y})$, \ie to produce realistic face images from an ID vector.

By design, the many-to-one mapping of FR methods assigns multiple images of a given identity to the same ID vector. The inverse one-to-many problem, \ie producing a high-dimensional image from a low-dimensional ID vector, is extremely challenging. Previous methods often rely on the gradient of FR models either directly~\cite{DBLP:journals/corr/ZhmoginovS16} or use it during training in the form of a loss function~\cite{cole, latent_space_mapping}. This gradient or information about the model's architecture and weights is often not available, \eg if using an API of a proprietary model. We therefore focus on the more generally applicable \emph{black-box} setting, where only the resulting ID vectors are available. 
In addition to being more general, the black-box setting simplifies the analysis of different FR models as explored in the supplementary material. Another benefit is that we can easily extend our conditioning mechanism to include information from different, even non-differentiable, sources (\eg labels, biological signals).

We propose the \underline{id}entity \underline{d}enoising \underline{d}iffusion \underline{p}robabilistic \underline{m}odel (ID3PM), the first method that uses a diffusion model (DM) to invert the latent space of an FR model, \ie to generate identity-preserving face images conditioned solely on black-box ID vectors as seen in \cref{fig:teaser_small}. We show mathematically that we can effectively invert a model $f$ even without access to its gradients by using a conditional DM. This allows us to train our method with an easy-to-obtain data set of pairs of images and corresponding ID vectors (easily extracting from images) without an identity-specific loss term. 

Our method obtains state-of-the-art performance for the inversion task and is, to the best of our knowledge, the first black-box FR model inversion method \underline{with} control over the generation process as seen in \cref{fig:teaser_small}. 
Specifically, we can control (1) the diversity among samples generated from the same ID vector via the classifier-free guidance scale, (2) identity-specific features (\eg age) via smooth transitions in the ID vector latent space, 
and (3) identity-agnostic features (\eg pose) via explicit attribute conditioning.

To summarize, our main contributions are:
\begin{enumerate}[itemsep=-2pt,topsep=2pt]
    \item Showing that the conditional diffusion model loss naturally emerges from an analysis of the black-box inversion problem.
    \item Applying the resulting framework to invert face recognition models without identity-specific loss functions. 
    \item Demonstrating state-of-the-art performance in generating diverse, identity-preserving face images from black-box ID vectors.
    \item Providing control mechanisms for the face recognition model inversion task.
\end{enumerate}

\section{Related work} \label{sec:related_work}

\begin{table*}[htbp]
\centering
    \footnotesize{
    \begin{tabular}{ llllllll }
    \toprule
    Method & Black-box & \multicolumn{2}{c}{FR model queries (inference)} & Training data set & Realistic $^1$ & Mapping \\
    \midrule
    Zhmoginov and Sandler~\cite{DBLP:journals/corr/ZhmoginovS16} & \cellcolor{red!10}No & \cellcolor{red!10} $\sim$ 1000 $^2$ & \cellcolor{green!10}1  $^2$ & \cellcolor{green!10}Any images & \cellcolor{red!10}No & \cellcolor{red!10}One-to-one\\
    Cole \etal~\cite{cole} & \cellcolor{red!10}No & \cellcolor{green!10}1 & \cellcolor{green!10} & \cellcolor{red!10}Frontalized images & \cellcolor{green!10}Yes & \cellcolor{red!10}One-to-one \\
    Nitzan \etal~\cite{latent_space_mapping} & \cellcolor{red!10}No & \cellcolor{green!10}1 & \cellcolor{green!10} & \cellcolor{green!10}Any images & \cellcolor{green!10}Yes & \cellcolor{green!10}One-to-many \\
    NbNet~\cite{nbnet} & \cellcolor{green!10}Yes & \cellcolor{green!10}1 & \cellcolor{green!10} & \cellcolor{red!10}Huge data set & \cellcolor{red!10}No & \cellcolor{red!10}One-to-one \\
    Gaussian sampling~\cite{gaussian_sampling} & \cellcolor{green!10}Yes & \cellcolor{red!10}240000 & \cellcolor{red!10} & \cellcolor{green!10}Data-set-free & \cellcolor{red!10}No & \cellcolor{green!10}One-to-many \\
    Yang \etal~\cite{background_knowledge_alignment} & \cellcolor{green!10}Yes & \cellcolor{green!10}1 & \cellcolor{green!10} & \cellcolor{green!10}Any images & \cellcolor{red!10}No & \cellcolor{red!10}One-to-one \\
    Vec2Face~\cite{vec2face} & \cellcolor{green!10}Yes & \cellcolor{green!10}1 & \cellcolor{green!10} & \cellcolor{red!10}Multiple images per identity & \cellcolor{green!10}Yes & \cellcolor{green!10}One-to-many \\
    StyleGAN search~\cite{stylegan-search} & \cellcolor{green!10}Yes & \cellcolor{red!10}400 & \cellcolor{red!10} & \cellcolor{green!10}Data-set-free & \cellcolor{green!10}Yes & \cellcolor{green!10}One-to-many \\
    \midrule
    ID3PM (Ours) & \cellcolor{green!10}Yes & \cellcolor{green!10}1 & \cellcolor{green!10} & \cellcolor{green!10}Any images & \cellcolor{green!10}Yes & \cellcolor{green!10}One-to-many\\
    \bottomrule
    \end{tabular}
    }
\caption{Comparison of state-of-the-art face recognition (FR) model inversion methods. Our method does not have any of the common shortcomings, producing diverse, realistic images from black-box ID vectors with few requirements for the training data set or accessibility of the FR model during inference. 
$^1$~By visual inspection of the results of the respective papers.
$^2$~The authors propose two methods: one taking hundreds or thousands of queries and the second one doing it in one shot. 
}
\label{table:comp}
\end{table*}

\subsection{Face recognition}

While early deep learning works such as DeepFace~\cite{deepface} and VGG-Face~\cite{vggface} treated FR as a classification problem (one class per identity), FaceNet~\cite{facenet} introduced the triplet loss, a distance-based loss function. 
The trend then shifted towards margin-based softmax methods~\cite{margin_based_1, margin_based_2, margin_based_3, arcface} that incorporate a margin penalty and perform sample-to-class rather than sample-to-sample comparisons. More recently, some FR methods tackle specific challenges such as robustness to different quality levels~\cite{adaface} and occlusions~\cite{mask_robustness, from}.

\subsection{Inversion of face recognition models}

Similar to gradient-based feature visualization techniques~\cite{feature_vis_1, feature_vis_2, feature_vis_3, deepdream}, Zhmoginov and Sandler~\cite{DBLP:journals/corr/ZhmoginovS16} perform gradient ascent steps using the gradient of a pre-trained FR model to generate images that approach the same ID vector as a target image. To avoid generating adversarial examples, strong image priors such as a total-variation loss and a guiding image are necessary. Cole \etal~\cite{cole} transform the one-to-many task into a one-to-one task by mapping features of an FR model to frontal, neutral-expression images, which requires a difficult-to-obtain data set. Nitzan \etal~\cite{latent_space_mapping} map the identity features and attributes of images into the style space of a pre-trained StyleGAN~\cite{stylegan} to produce compelling results. However, their method struggles to encode real images since it is trained exclusively with images generated by StyleGAN. Furthermore, all of the above methods require white-box access to (the gradient of) an FR model, which is not always available in practice.

Many black-box methods view the problem from a security lens, focusing on generating images that deceive an FR model rather than appearing realistic. Early attempts using linear~\cite{mohanty2007scores} or radial basis function models~\cite{mignon2013reconstructing} lacked generative capacity to produce realistic images. NbNet~\cite{nbnet} introduces a neighborly de-convolutional neural network that can generate images with a reasonable resemblance to a given image, but it has line artifacts and relies on a huge data set augmented with a GAN. On the contrary, Razzhigaev \etal~\cite{gaussian_sampling} propose a data-set-free method using Gaussian blobs (which we call ``Gaussian sampling'' for simplicity), but they need  thousands of FR model queries ($10$-$15$ minutes) per image, and their results lack realism. Yang \etal~\cite{background_knowledge_alignment} rely on background knowledge to invert a model and only produce blurry images in the black-box setting. Vec2Face~\cite{vec2face} uses a bijection metric and knowledge distillation from a black-box FR model to produce realistic identity-preserving faces; however, it requires a large data set with multiple images per identity (Casia-WebFace~\cite{casiawebface}) during training. 
The method by Vendrow and Vendrow~\cite{stylegan-search} (which we call ``StyleGAN search'') searches the latent space of a pre-trained StyleGAN2~\cite{stylegan2} to find images with an ID vector close to the target.
While their search strategy generates highly realistic images, it needs hundreds of FR model queries ($5$-$10$ minutes) per image and often lands in local minima, resulting in completely 
different identities.

\Cref{table:comp} compares attributes of state-of-the-art FR model inversion methods. Ours is the only one that generates diverse, realistic, identity-preserving images in the black-box setting, can be trained with easy-to-obtain data, and only requires one FR model query during inference.

\subsection{Diffusion models for inverse problems}

A number of approaches for solving inverse problems in a more general setting using conditional~\cite{SR3, palette} and unconditional~\cite{kadkhodaie2021stochastic, DDRM, PiGDM, DPS, MCG, plugandplay, BlindDPS, CCDF, ILVR, RED_Diff, song2020score} exist; however, they mostly focus on image-to-image tasks such as inpainting and super-resolution whereas we focus on a vector-to-image task. The method by Graikos \etal~\cite{plugandplay} can generate images from low-dimensional, nonlinear constraints such as attributes, but it requires the gradient of the attribute classifier during inference whereas ours does not. Thus, conditional diffusion models with vectors as additional input~\cite{ddpm3, dalle2, stable_diffusion, imagen}, while not directly geared towards inversion, are conceptually more similar to our approach.

\section{Motivation} \label{sec:motivation}

\subsection{Inverse problems}

In a system under study, we often have a \emph{forward problem} or function $f$ that corresponds to a set of observations $\mb{y} \sim \mathcal{Y}$.  The function $f$ has input arguments $\mb{x}$ and a set of parameters $\bm{\theta}$, such that $f(\mb{x}; \bm{\theta}) = \mb{y}$.  An \emph{inverse problem} seeks to reverse this process and make inferences about the values of $\mb{x}$ or $\bm{\theta}$ given the observations $\mb{y}$. For the application explored in this work, $f$ is a face recognition model that takes an image $\mb{x}$ as input and produces an ID vector $\mb{y}$.

When the function $f$ is not bijective, no inverse exists in the traditional mathematical sense.  However, it is possible to generalize our concept of what an inverse is to accommodate the problem of model inversion, namely by considering an inverse to be the set of pre-images of the function $f$ that map $\epsilon$-close to the target $\mb{y}$.  For bijective $f$, this corresponds to the traditional inverse for $\epsilon = 0$.

\begin{figure*}[htbp]
    \centering
    \includegraphics[trim = 14mm 23mm 69mm 84mm, clip, width=0.98\linewidth]{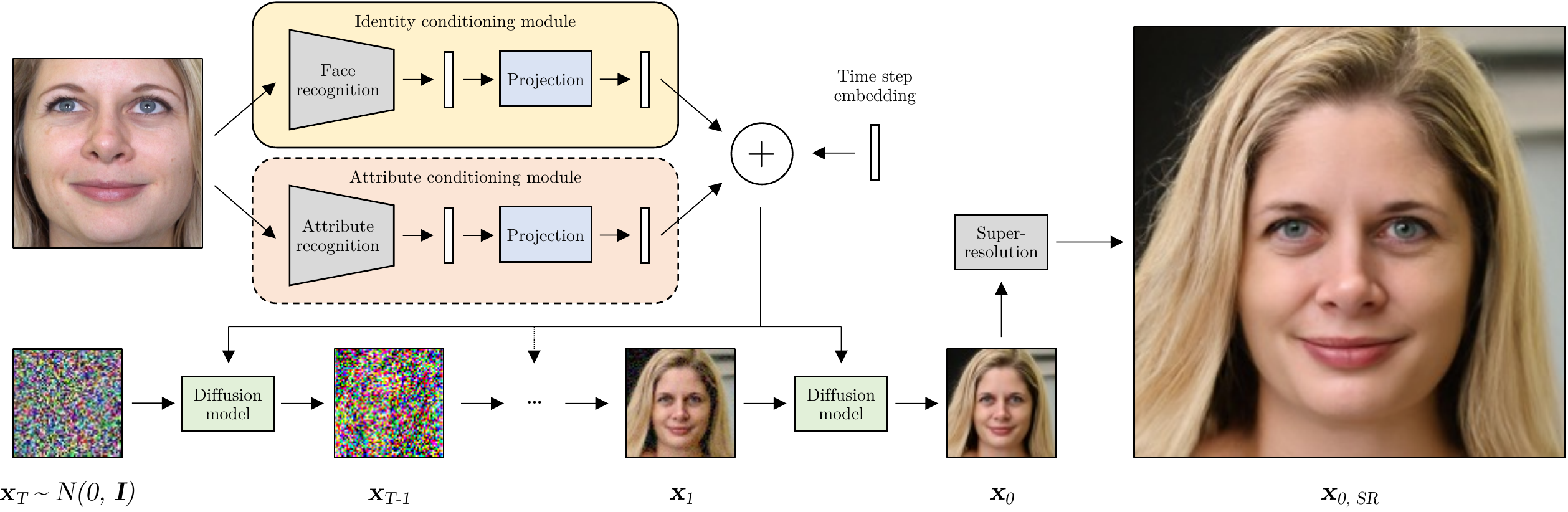}
    \caption{Method architecture. Given an image of a source identity, the identity conditioning module extracts the ID vector with a black-box, pre-trained face recognition network. This is projected with a fully connected layer and added to the time step embedding which is injected into the residual blocks of a diffusion model (DM). Starting with Gaussian noise $\mb{x}_T$, the DM iteratively denoises the image to finally obtain the output image $\mb{x}_0$ in $64 \times 64$ resolution. Lastly, the image is upsampled to a resolution of $256 \times 256$ using an unconditional super-resolution DM. The optional attribute conditioning module helps disentangle features and allows intuitive control over attributes such as the pose. Gray components are frozen during training.}
    \label{fig:architecture}
\end{figure*}

\subsubsection{Model inversion with model access} \label{sec:inv_access}

One way to handle the model-inversion problem when $f$ is not bijective is to treat it pointwise, defining a loss, such as 
\begin{equation}
	\mathcal{L} = \frac{1}{2} \| \mb{y} - f(\mb{x}) \|^2, \label{eq:loss}
\end{equation}
and minimizing it via gradient descent on $\mb{x}$ from some starting point $\mb{x}_0$ 
according to
\begin{equation}
	\Delta \mb{x}_t = - \nabla_{\mb{x}} \mathcal{L} = \left( \frac{\partial f}{\partial \mb{x}} \right)^\top \left( \mb{y} - f(\mb{x}) \right). \label{eq:update}
\end{equation}

In common cases where the inverse problem is one-to-many, we can take a statistical approach.  Here we want to sample from $p(\mb{x}|\mb{y})$, which is equivalent to drawing from the pre-image set that defines the inverse $f^{-1}(\mb{y})$.  

However, if we assume a Gaussian observation model 
\begin{equation}
	p(\mb{y} |\mb{x}) = \mathcal{N}(\mb{y}; f(\mb{x}), \sigma^2 \mb{I}) \propto \exp \left( -\frac{\mathcal{L}}{\sigma^2} \right), \label{eq:observation}
\end{equation}
where  the last term follows from \eqref{eq:loss}, then 
we can rewrite equation \eqref{eq:update} as $\Delta \mb{x}_t \propto \sigma^2 \nabla_{\mb{x}} \log p(\mb{y} |\mb{x}_t)$.

This shows that traditional model inversion via gradient descent performs a type of deterministic sampling from $p(\mb{y}|\mb{x})$---and not the distribution we want, $p(\mb{x}|\mb{y})$---by pushing toward modes of $p(\mb{y}|\mb{x})$ close to the initialization point $\mb{x}_0$, regardless of whether it possesses the desired characteristics of the data $p(\mb{x})$.  This can lead to results, such as adversarial examples~\cite{goodfellow2014explaining}, that, while technically satisfying the mathematical criteria of inversion, do not appear to come from $p(\mb{x})$.  

Various types of regularization exist to attempt to avoid this issue, which are most often \emph{ad hoc} methods geared toward the specific problem at hand~\cite{DBLP:journals/corr/ZhmoginovS16, mahendran2015understanding, creswell2018inverting, xia2022gan}.  A more general approach is to introduce regularization terms proportional to the \emph{(Stein) score}, $\nabla_{\mb{x}} \log p(\mb{x})$, since $$\nabla_{\mb{x}} \log p(\mb{x} |\mb{y}) = \nabla_{\mb{x}} \log p(\mb{y} |\mb{x}) + \nabla_{\mb{x}} \log p(\mb{x})$$ provides the \emph{conditional} score needed to sample from $p(\mb{x}|\mb{y})$, the distribution we are actually interested in.

Previous work has shown that diffusion models (DMs) effectively learn the score $\nabla_{\mb{x}} \log p(\mb{x})$, which allows them to be used alongside model gradients to guide sampling~\cite{song2020score, ddpm1, ddpm2, ddpm3, edm}.  When those models are classifiers, the procedure is known as \emph{classifier guidance}~\cite{ddpm3}.  However, this imposes an additional computational burden on sampling and also requires that the model $f$ be differentiable.

\subsubsection{Model inversion without full model access} \label{sec:inv_no_access}

In the case we focus on in this work, we assume to have access only to the values of the function $f$ via some oracle or a lookup table of $(\mb{x}, \mb{y})$ pairs but not its gradient $\nabla f$.  In this case, also referred to as \emph{black-box} setting, we may wish to train a function $g_{\bm{\psi}}$ to learn the inverse by minimizing  
\begin{equation}
	\mathcal{J} = \| \mb{x} - g_{\bm{\psi}}(\mb{y}) \|^2 \label{eq:J-loss}
\end{equation}
across all observed $\{ (\mb{x}, \mb{y}) \}$.  Recalling that $\mb{y} = f(\mb{x})$, we have essentially described an encoder-decoder setup with the encoder frozen and only the decoder being trained, which requires no gradients from the ``encoder'' $f$.

If we consider perturbed data $\tilde{\mb{x}} = \mb{x} + \bm{\epsilon}$, where $\bm{\epsilon} \sim \mathcal{N}(\bm{0}, \sigma^2_t \mb{I})$.  Then \eqref{eq:J-loss} is equivalent to 
\begin{equation}
\begin{split}
	\tilde{\mathcal{J}} &= \| (\tilde{\mb{x}} - \mb{x}) - (\tilde{\mb{x}} - g_{\bm{\psi}}(\mb{y})) \|^2 \\
	&= \| \bm{\epsilon}- \bm{\epsilon}_{\bm{\theta}}(\tilde{\mb{x}}, \mb{y}, t)  \|^2, \label{eq:err_loss}
\end{split}	
\end{equation}
and we are now training a conditional model $\bm{\epsilon}_{\bm{\theta}}$ to learn the noise added to $\mb{x}$ instead of a model $g$ to reconstruct $\mb{x}$.  This new task is exactly the one facing conditional diffusion models (\Cref{sec:diff_form}).

Although we cannot \emph{force} the model to leverage the conditioning on $\mb{y}$ or $t$, if it is to successfully minimize the loss $\tilde{\mathcal{J}}$, it should learn a function proportional to the conditional score.  That is because, by Tweedie's formula~\cite{efron2011tweedie, kim2021noise2score}, 
\begin{equation}
    \begin{split}
        \mathbb{E}[\mb{x}|\tilde{\mb{x}}, \mb{y}] &= \tilde{\mb{x}} + \sigma^2_t \nabla_{\tilde{\mb{x}}} \log p(\tilde{\mb{x}} | \mb{y}) \\
        &\approx \tilde{\mb{x}} - \bm{\epsilon}_{\bm{\theta}}(\tilde{\mb{x}}, \mb{y}, t). \\ 
    \end{split}
\end{equation}
As a result, we can effectively sample from the ``inverse distribution'' $p(\tilde{\mb{x}}|\mb{y})$ via $\bm{\epsilon}_{\bm{\theta}}(\tilde{\mb{x}}, \mb{y}, t)$ using Langevin dynamics~\cite{bussi2007accurate, welling2011bayesian} without having access to the gradients of the model $f$ or any other model-specific loss terms.  %

Intuitively, during training, especially in early denoising steps, it is difficult for the DM to both denoise an image to get a realistic face \underline{and} match the specific training image. The ID vector contains information (\eg face shape) that the DM is incentivized to use ($\rightarrow$ lower loss) to get closer to the training image. 
During inference, the random seed determines identity-agnostic features ($\rightarrow$ many results), and the ID conditioning pushes the DM to generate an image that resembles the target identity.

\section{Method} \label{sec:method}

Motivated by the results from \cref{sec:motivation}, we adopt a conditional diffusion model (DM) for the task of inverting a face recognition (FR) model.
Since conditional DMs have inherent advantages for one-to-many and inversion tasks, this results in a minimal problem formulation compared to task-specific methods that require complicated supervision~\cite{vec2face} or regularization~\cite{DBLP:journals/corr/ZhmoginovS16} signals.
The overall architecture of our method is visualized in \cref{fig:architecture}.

\subsection{Diffusion model formulation}  \label{sec:diff_form}

We build up on the diffusion model proposed by Dhariwal and Nichol~\cite{ddpm3}. Given a sample $\mb{x}_0$ from the image distribution, a sequence $\mb{x}_1, \mb{x}_2,..., \mb{x}_T$ of noisy images is produced by progressively adding Gaussian noise according to a variance schedule. At the final time step, $\mb{x}_T$ is assumed to be pure Gaussian noise: $\mathcal{N}(0,\mathbf{I})$.
A neural network is then trained to reverse this diffusion process in order to predict $\mb{x}_{t-1}$ from the noisy image $\mb{x}_t$ and the time step $t$. 
To sample a new image, we sample $\mb{x}_T \sim \mathcal{N}(0,\mathbf{I})$ and iteratively denoise it, producing a sequence $\mb{x}_T, \mb{x}_{T-1}, \ldots, \mb{x}_1, \mb{x}_0$. The final image, $\mb{x}_0$, should resemble the training data.

As \cite{ddpm3}, we assume that we can model $p_{\bm{\theta}}(\mb{x}_{t-1} | \mb{x}_t)$ as a Gaussian $\mathcal{N}(\mb{x}_{t-1}; \mb{\mu}_{\bm{\theta}}(\mb{x}_t,t), \bm{\Sigma}_{\bm{\theta}}(\mb{x}_t,t))$ whose mean $\mb{\mu}_{\bm{\theta}}(\mb{x}_t,t)$ can be calculated as a function of $\bm{\epsilon}_{\bm{\theta}}(\mb{x}_t,t)$, the (unscaled) noise component of $\mb{x}_t$. We extend this by conditioning on the ID vector $\mb{y}$ and thus predict $\bm{\epsilon}_{\bm{\theta}}(\mb{x}_t,\mb{y},t)$.
Extending \cite{ddpm2} to the conditional case, we predict the noise $\bm{\epsilon}_{\bm{\theta}}(\mb{x}_t,\mb{y},t)$ and the variance $\bm{\Sigma}_{\bm{\theta}}(\mb{x}_t,\mb{y},t)$ from the image $\mb{x}_t$, the ID vector $\mb{y}$, and the time step $t$, using the objective
\begin{equation}
    \mathcal{L}_{\text{simple}} = \mathbb{E}_{t,\mb{x}_0,\mb{y},\bm{\epsilon}}[||\bm{\epsilon} - \bm{\epsilon}_{\bm{\theta}}(\mb{x}_t,\mb{y},t)||^2].
\end{equation}
For more details, refer to the diffusion model works~\cite{ddpm1, ddpm2, ddpm3}. Note that this objective is identical to the one theoretically derived in~\eqref{eq:err_loss}.  While some recent work has considered the application of diffusion models to inverse problems, they typically assume $p(\mb{y}|\mb{x})$ is known~\cite{kadkhodaie2021stochastic, DDRM, PiGDM, DPS, MCG, plugandplay, BlindDPS, CCDF, ILVR, RED_Diff, song2020score}, while we make no such assumption.

Following Ramesh \etal~\cite{dalle2}, we adapt classifier-free guidance~\cite{classifierfree} by setting the ID vector to the $\mb{0}$-vector with $10\%$ probability during training (unconditional setting). During inference, we sample from both settings, and the model prediction $\hat{\bm{\epsilon}}_\theta$ becomes
\begin{equation}
    \hat{\bm{\epsilon}}_{\bm{\theta}}(\mb{x}_t, \mb{y}, t) = \bm{\epsilon}_{\bm{\theta}}(\mb{x}_t, \mb{0}, t) + s [\bm{\epsilon}_{\bm{\theta}}(\mb{x}_t, \mb{y}, t) - \bm{\epsilon}_{\bm{\theta}}(\mb{x}_t, \mb{0}, t)],
\end{equation}
where $s \geq 1$ is the guidance scale. Higher guidance scales cause the generation process to consider the identity conditioning more.

\subsection{Architecture}

The model is a U-net~\cite{unet} that takes the image $\mb{x}_t$, the ID vector $\mb{y}$, and the time step $t$ as input. The U-net architecture is adapted from \cite{ddpm3} and is described in detail in the supplementary material.
To condition the diffusion model on the identity, we add an identity embedding to the residual connections of the ResNet blocks, as commonly done for class embeddings~\cite{ddpm3} and the CLIP~\cite{clip} embedding in text-to-image generation methods~\cite{dalle2, imagen}. The identity embedding is obtained by projecting the ID vector through a learnable fully connected layer such that it has the same size as the time step embedding and can be added to it.

\subsection{Controllability}

Due to its robustness and ability to pick a mode by setting the random seed during image generation, our method permits smooth interpolations and analyses in the ID vector latent space unlike other works that invert FR models. For example, we can smoothly interpolate between different identities as visualized in \cref{fig:teaser_small}. Furthermore, we can find meaningful directions in the latent spaces. Since the directions extracted automatically using principal component analysis (PCA) are generally difficult to interpret beyond the first dimension (see supplementary material), we calculate custom directions using publicly available metadata~\cite{ffhq_metadata} for the FFHQ data set. For binary features (\eg glasses), we define the custom direction vector as the difference between the mean ID vectors of the two groups. For continuous features (\eg age), we map to the binary case by considering ID vectors with feature values below the $10$th percentile and values above the $90$th percentile for the two groups respectively. 
Examples of traveling along meaningful ID vector directions can be seen in \cref{fig:teaser_small}.

To better disentangle identity-specific and identity-agnostic information and obtain additional interpretable control, we can optionally extend our method by also conditioning the DM on an attribute vector as done for the ID vector. In practice, we recommend using only \underline{identity-agnostic} attributes (referred to as set 1) along with the identity. In the supplementary material, we also show attribute sets that overlap more with identity (sets 2 \& 3) for completeness.

\subsection{Implementation details} \label{sec:implementation_details}

As data set, we use FFHQ~\cite{stylegan}
and split it into $69000$ images for training and $1000$ for testing. As we can only show images of individuals with written consent (see \cref{sec:ethics}), we use a proprietary data set of faces for the qualitative results in this paper. To condition our model, we use ID vectors from a PyTorch FaceNet implementation~\cite{facenet, facenet_pytorch} or the default InsightFace method~\cite{insightface}. To evaluate the generated images and thereby match the verification accuracy on real images shown in Vec2Face~\cite{vec2face} as closely as possible, we use the official PyTorch ArcFace implementation~\cite{arcface, arcface_torch} and a TensorFlow FaceNet implementation~\cite{facenet, facenet_tf}. A detailed description of the remaining implementation details and ID vectors is in the supplementary material.

\section{Experiments and results} \label{sec:experiments}

\subsection{Comparison to state-of-the-art methods}

We mainly compare our model with the three methods that generate faces from black-box features whose code is available online: NbNet~\cite{nbnet} (``vgg-percept-nbnetb'' parameters), Gaussian sampling~\cite{gaussian_sampling}, and StyleGAN search~\cite{stylegan-search}. 

\begin{figure}[htpb]
    \centering
    \addtolength{\tabcolsep}{-5.5pt}
    \footnotesize{
    \begin{tabular}{ccccc}
        \includegraphics[width=0.09\textwidth]{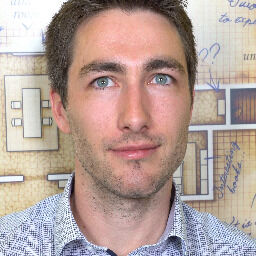} & 
        \includegraphics[width=0.09\textwidth]{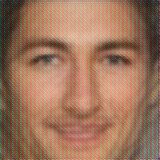} & 
        \includegraphics[width=0.09\textwidth]{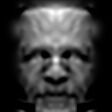} & 
        \includegraphics[width=0.09\textwidth]{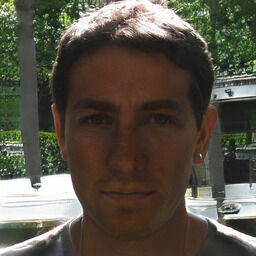} & 
        \includegraphics[width=0.09\textwidth]{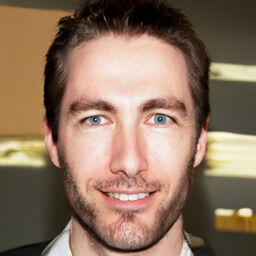} \\
        \\[-0.4cm]
        \includegraphics[width=0.09\textwidth]{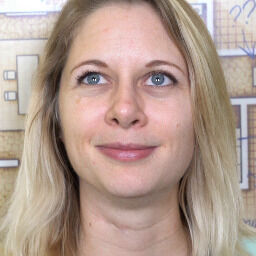} & 
        \includegraphics[width=0.09\textwidth]{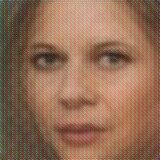} & 
        \includegraphics[width=0.09\textwidth]{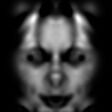} & 
        \includegraphics[width=0.09\textwidth]{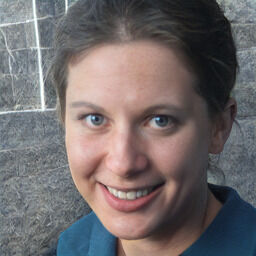} & 
        \includegraphics[width=0.09\textwidth]{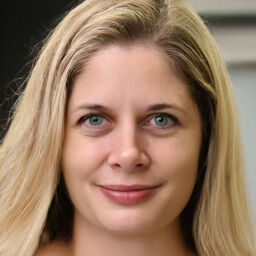} \\
        \\[-0.4cm]
        \includegraphics[width=0.09\textwidth]{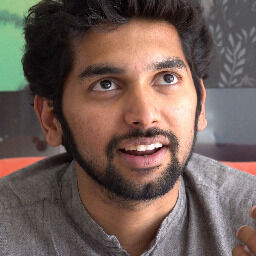} & 
        \includegraphics[width=0.09\textwidth]{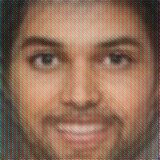} & 
        \includegraphics[width=0.09\textwidth]{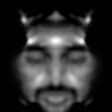} & 
        \includegraphics[width=0.09\textwidth]{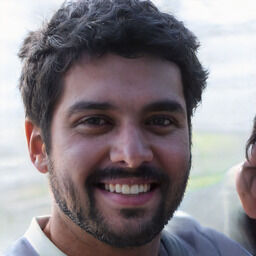} & 
        \includegraphics[width=0.09\textwidth]{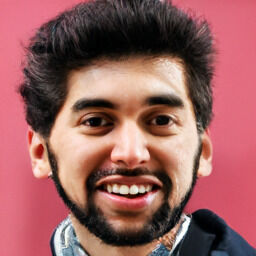} \\
        \\[-0.4cm]
        \includegraphics[width=0.09\textwidth]{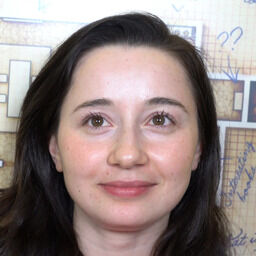} & 
        \includegraphics[width=0.09\textwidth]{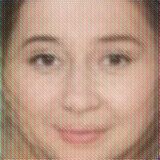} & 
        \includegraphics[width=0.09\textwidth]{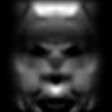} & 
        \includegraphics[width=0.09\textwidth]{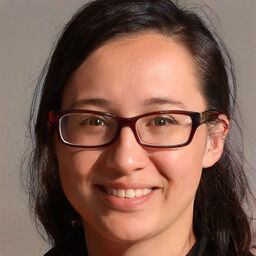} & 
        \includegraphics[width=0.09\textwidth]{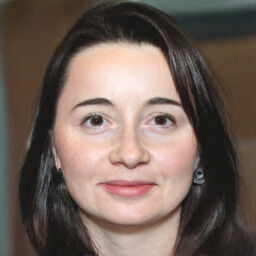} \\
        \\[-0.4cm]
        \includegraphics[width=0.09\textwidth]{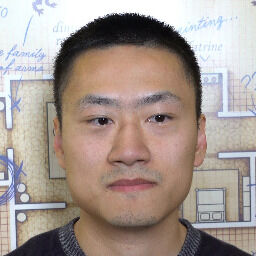} & 
        \includegraphics[width=0.09\textwidth]{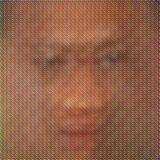} & 
        \includegraphics[width=0.09\textwidth]{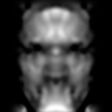} & 
        \includegraphics[width=0.09\textwidth]{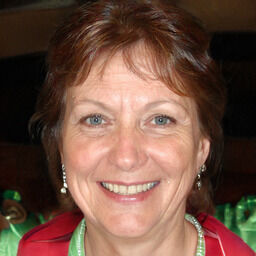} & 
        \includegraphics[width=0.09\textwidth]{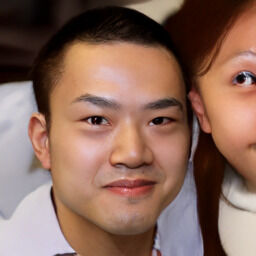} \\

        \multirow{2}{*}{\begin{tabular}[c]{@{}c@{}}Original \\ image \end{tabular}} & \multirow{2}{*}{NbNet~\cite{nbnet}} & \multirow{2}{*}{\begin{tabular}[c]{@{}c@{}}Gaussian \\ sampling~\cite{gaussian_sampling} \end{tabular}} & \multirow{2}{*}{\begin{tabular}[c]{@{}c@{}}StyleGAN \\ search~\cite{stylegan-search} \end{tabular}} & \multirow{2}{*}{\begin{tabular}[c]{@{}c@{}}ID3PM \\ (Ours) \end{tabular}} \\
        \\
    \end{tabular}
    }
    \addtolength{\tabcolsep}{5.5pt}
    \caption{Qualitative evaluation with state-of-the-art methods. The generated images of our method (with InsightFace~\cite{insightface} ID vectors) look realistic and resemble the identity of the original image more closely than other methods. Note that the second-best performing method, StyleGAN search~\cite{stylegan-search}, often fails completely as seen in the last row.}
    \label{fig:qualitative_comp}
\end{figure}

\Cref{fig:qualitative_comp} compares the outputs of our method with those of current state-of-the-art methods. While capturing the identity of the input face well in some cases, NbNet~\cite{nbnet} and Gaussian sampling~\cite{gaussian_sampling} both fail to produce realistic faces. In contrast, StyleGAN search~\cite{stylegan-search} always generates high-quality images, but they are not always faithful to the original identity, sometimes failing completely 
as seen in the last row. Our method is the only method that produces high-quality, realistic images that consistently resemble the original identity. Our observations are supported by the user study in the supplementary material.

For the quantitative evaluation of the identity preservation, we generate one image from each ID vector of all 1000 images of the FFHQ~\cite{stylegan} test set for each method. We then calculate the distances according to the ArcFace~\cite{arcface, arcface_torch} and FaceNet~\cite{facenet, facenet_tf} face recognition methods for the $1000$ respective pairs. The resulting distance distributions are plotted in \cref{fig:plot_ffhq}. Note that StyleGAN search~\cite{stylegan-search} optimizes the FaceNet distance during the image generation and thus performs well when evaluated with FaceNet but poorly when evaluated with ArcFace. The opposite effect can be seen for Gaussian sampling, which optimizes ArcFace during image generation. Despite not optimizing the ID vector distance directly (neither during training nor inference), our method outperforms all other methods, producing images that are closer to the original images' identities.

\pgfplotstableread{plots/ffhq_facenet_plot_hist.dat}{\tableffhqfacenethist}
\pgfplotstableread{plots/ffhq_arcface_plot_hist.dat}{\tableffhqarcfacehist}

\pgfplotsset{
compat=1.11,
legend image code/.code={
\draw[mark repeat=2,mark phase=2]
plot coordinates {
(0cm,0cm)
(0.15cm,0cm)        %
(0.3cm,0cm)         %
};%
}
}
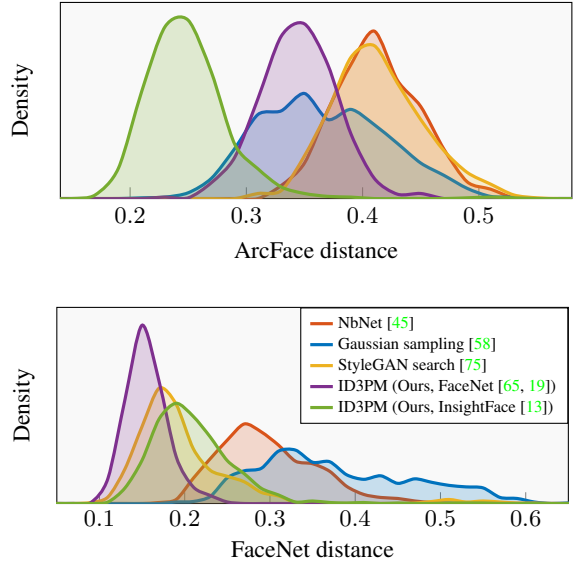
\begin{figure}[htpb]
    \centering
    \begin{subfigure}[b]{0.48\textwidth}
        \centering
        \begin{tikzpicture}
        \tikzstyle{every node}=[font=\small]
        \begin{axis}[
            axis background/.style={fill=gray!5},
            every axis plot/.append style={very thick},
            xmin = 0.14, xmax = 0.58,
            ymin = 0, 
            xtick pos=bottom,
            ytick style={draw=none},
            yticklabels={,,},
            width = \textwidth,
            height = 0.5\textwidth,
            legend cell align = {left},
            legend pos = north east,
            legend style={nodes={scale=0.7, transform shape}, at={(1, 1)},anchor=north east},
            xlabel=ArcFace distance,
            ylabel=Density,
        ]
        \addplot[matlab_orange, smooth, fill=matlab_orange, fill opacity=0.2] table [x = {x}, y = {y0}] {\tableffhqarcfacehist};
        \addplot[matlab_blue, smooth, fill=matlab_blue, fill opacity=0.2] table [x = {x}, y = {y1}] {\tableffhqarcfacehist};
        \addplot[matlab_yellow, smooth, fill=matlab_yellow, fill opacity=0.2] table [x = {x}, y = {y2}] {\tableffhqarcfacehist};
        \addplot[matlab_purple, smooth, fill=matlab_purple, fill opacity=0.2] table [x = {x}, y = {y3}] {\tableffhqarcfacehist};
        \addplot[matlab_green, smooth, fill=matlab_green, fill opacity=0.2] table [x = {x}, y = {y4}] {\tableffhqarcfacehist};
        \end{axis}
        \end{tikzpicture}
        \label{fig:plot_ffhq_arcface_hist}
        \vspace{0.5cm}
    \end{subfigure}
    \begin{subfigure}[b]{0.48\textwidth}
        \centering
        \begin{tikzpicture}
        \tikzstyle{every node}=[font=\small]
        \begin{axis}[
            axis background/.style={fill=gray!5},
            every axis plot/.append style={very thick},
            xmin = 0.05, xmax = 0.65,
            ymin = 0, 
            xtick pos=bottom,
            ytick style={draw=none},
            yticklabels={,,},
            width = \textwidth,
            height = 0.5\textwidth,
            legend cell align = {left},
            legend pos = north east,
            legend style={nodes={scale=0.7, transform shape}, at={(1, 1)},anchor=north east},
            xlabel=FaceNet distance,
            ylabel=Density,
        ]
        \addplot[matlab_orange, smooth, fill=matlab_orange, fill opacity=0.2] table [x = {x}, y = {y0}] {\tableffhqfacenethist};
        \addplot[matlab_blue, smooth, fill=matlab_blue, fill opacity=0.2] table [x = {x}, y = {y1}] {\tableffhqfacenethist};
        \addplot[matlab_yellow, smooth, fill=matlab_yellow, fill opacity=0.2] table [x = {x}, y = {y2}] {\tableffhqfacenethist};
        \addplot[matlab_purple, smooth, fill=matlab_purple, fill opacity=0.2] table [x = {x}, y = {y3}] {\tableffhqfacenethist};
        \addplot[matlab_green, smooth, fill=matlab_green, fill opacity=0.2] table [x = {x}, y = {y4}] {\tableffhqfacenethist};
        \footnotesize{
        \legend{
            NbNet~\cite{nbnet},
            Gaussian sampling~\cite{gaussian_sampling},
            StyleGAN search~\cite{stylegan-search},
            {ID3PM (Ours, FaceNet~\cite{facenet, facenet_pytorch})},
            {ID3PM (Ours, InsightFace~\cite{insightface})}
        }
        }
        \end{axis}
        \end{tikzpicture}
        \label{fig:plot_ffhq_facenet_hist}
    \end{subfigure}
    \caption{Probability density functions of the ArcFace~\cite{arcface, arcface_torch} and FaceNet~\cite{facenet, facenet_tf} distances (lower is better) of $1000$ FFHQ~\cite{stylegan} test images and their respective reconstructions.}
    \label{fig:plot_ffhq}
\end{figure}

\begin{table*}[htbp]
\centering
    \small{
    \begin{tabular}{llcclcclcc}
    \toprule
    \multirow{2}{*}{Method} &  & \multicolumn{2}{c}{LFW} &  & \multicolumn{2}{c}{AgeDB-30} &  & \multicolumn{2}{c}{CFP-FP} \\
    \cmidrule{3-4} \cmidrule{6-7} \cmidrule{9-10}
    & & {ArcFace $\uparrow$} & {FaceNet $\uparrow$} & & {ArcFace $\uparrow$} & {FaceNet $\uparrow$} & & {ArcFace $\uparrow$} & {FaceNet $\uparrow$} \\
    \midrule
    Real images & & 99.83\% & 99.65\% & & 98.23\% & 91.33\% & & 98.86\% & 96.43\% \\
    \midrule
    NbNet~\cite{nbnet} & & 87.32\% & 92.48\% & & 81.83\% & 82.25\% & & 87.36\% & 89.89\% \\
    Gaussian sampling~\cite{gaussian_sampling} & & 89.10\% & 75.07\% & & 80.43\% & 63.42\% & & 61.39\% & 55.26\% \\
    StyleGAN search~\cite{stylegan-search} & & 82.43\% & 95.45\% & & 72.70\% & 85.22\% & & 80.83\% & 92.54\% \\
    Vec2Face~\cite{vec2face} $^1$ & & 99.13\% & 98.05\% & & 93.53\% & \textbf{89.80\%} & & 89.03\% & 87.19\% \\
    \midrule
    ID3PM (Ours, FaceNet~\cite{facenet, facenet_pytorch}) & & 97.65\% & \textbf{98.98\%} & & 88.22\% & 88.00\% & & 94.47\% & \textbf{95.23\%} \\
    ID3PM (Ours, InsightFace~\cite{insightface}) & & \textbf{99.20\%} & 96.02\% & & \textbf{94.53\%} & 79.15\% & & \textbf{96.13\%} & 87.43\% \\
    \bottomrule
    \end{tabular}
    }
\caption{Quantitative evaluation of the identity preservation with state-of-the-art methods. The scores depict the matching accuracy when replacing one image of each positive pair with the image generated from its ID vector for the protocols of the LFW~\cite{lfw}, AgeDB-30~\cite{agedb}, and CFP-FP~\cite{cfpfp} data sets. The best performing method per column is marked in bold. 
$^1$ Values taken from their paper.}
\label{table:face_recog_scores}
\end{table*}

To further evaluate the identity preservation and to compare to Vec2Face~\cite{vec2face} despite their code not being available online, we follow the procedure used in Vec2Face~\cite{vec2face}. Specifically, we use the official validation protocols of the LFW~\cite{lfw}, AgeDB-30~\cite{agedb}, and CFP-FP~\cite{cfpfp} data sets and replace the first image in each positive pair with the image reconstructed from its ID vector, while keeping the second image as the real reference face. The face matching accuracies for ArcFace~\cite{arcface, arcface_torch} and FaceNet~\cite{facenet, facenet_tf} are reported in \cref{table:face_recog_scores}. Our method outperforms NbNet~\cite{nbnet}, Gaussian sampling~\cite{gaussian_sampling}, and StyleGAN search~\cite{stylegan-search} in almost all tested configurations and performs on-par with or better than Vec2Face~\cite{vec2face}. 
Note that our method has fewer requirements for the training data set ($70000$ images vs. $490000$ images grouped into $10000$ classes) and produces visually superior results compared to Vec2Face~\cite{vec2face}, as confirmed in the user study in the supplementary material.

To evaluate the diversity of the generated results, we generate $100$ images for the first $50$ identities of the FFHQ~\cite{stylegan} test set. Motivated by the diversity evaluation common in unpaired image-to-image translation literature~\cite{starganv2, drit++}, we calculate the mean pairwise LPIPS~\cite{lpips} distances among all images of the same identity. We further calculate the mean pairwise pose and expressions extracted using 3DDFA\_V2~\cite{3ddfa}. We additionally calculate the mean identity vector distances according to ArcFace~\cite{arcface, arcface_torch} and FaceNet~\cite{facenet, facenet_tf} to measure the identity preservation. We report these values in \cref{table:diversity_eval}.

Since NbNet~\cite{nbnet} is a one-to-one method and Gaussian sampling~\cite{gaussian_sampling} produces faces that often fail to be detected by 3DDFA\_V2~\cite{3ddfa}, we only compare with StyleGAN search~\cite{stylegan-search}. In our default configuration (marked with $^*$ in \cref{table:diversity_eval}), we obtain similar diversity scores as StyleGAN search~\cite{stylegan-search}, while preserving the identity much better. 
Note that the diversity scores are slightly skewed in favor of methods whose generated images do not match the identity closely since higher variations in the identity also lead to more diversity in the LPIPS~\cite{lpips} features. 

\begin{table*}[htbp]
\centering
    \addtolength{\tabcolsep}{-2pt}
    \small{
    \begin{tabular}{lllS[table-format=2.2]S[table-format=1.2]S[table-format=1.3]lS[table-format=1.3]S[table-format=1.3]}
    \toprule
    \multirow{2}{*}{Method} & \multirow{2}{*}{Setting} & & \multicolumn{3}{c}{Diversity} &  & \multicolumn{2}{c}{Identity distance} \\ 
    \cmidrule{4-6} \cmidrule{8-9}
    & & & {Pose $\uparrow$} & {Expression $\uparrow$} & {LPIPS $\uparrow$} & & {ArcFace $\downarrow$} & {FaceNet $\downarrow$} \\
    \midrule
    StyleGAN search~\cite{stylegan-search} & {-} & & 12.57 & \textbf{1.57} & \textbf{0.317} & & 0.417 & 0.215 \\
    \midrule
    \multirow{5}{*}{\begin{tabular}[l]{@{}l@{}}ID3PM (Ours) \end{tabular}} & Guidance scale = 1.0 & & \textbf{17.36} & 1.35 & 0.315 & & 0.291 & 0.234 \\
    & \phantom{Guidance scale =} 1.5 & & 16.69 & 1.18 & 0.301 & & 0.260 & 0.211 \\
    & \phantom{Guidance scale =} 2.0 $^*$ & & 16.24 & 1.10 & 0.290 & & 0.247 & 0.203 \\
    & \phantom{Guidance scale =} 2.5 & & 15.88 & 1.05 & 0.282 & & 0.242 & 0.201 \\
    & \phantom{Guidance scale =} 3.0 & & 15.55 & 1.01 & 0.274 & & \textbf{0.239} & \textbf{0.200} \\
    \midrule
    \multirow{1}{*}{\begin{tabular}[l]{@{}l@{}}ID3PM (Ours) \end{tabular}} & Attribute conditioning & & 16.93 & 1.45 & 0.306 & & 0.302 & 0.252\\
    \bottomrule
    \end{tabular}
    }
    \addtolength{\tabcolsep}{2pt}
\caption{Quantitative evaluation of the diversity and identity distances of $100$ generated images for $50$ identities with StyleGAN search~\cite{stylegan-search}, different guidance scales, and attribute conditioning (set 1). InsightFace~\cite{insightface} ID vectors are used for our methods in this experiment. The best performing method per column is marked in bold.
$^*$ Indicates the default setting used in this paper and also for the run with attribute conditioning.}
\label{table:diversity_eval}
\end{table*}

\subsection{Controllability}

\subsubsection{Guidance scale}

\begin{figure}[htbp]
\centering
    \addtolength{\tabcolsep}{-5pt}
    \small{
    \begin{tabular}{cc}
    \raisebox{1.1\height}{\includegraphics[width=0.075\textwidth]{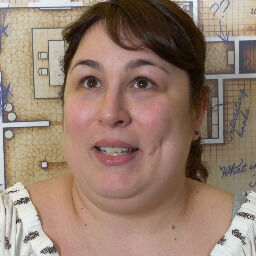}} & 
    \adjincludegraphics[width=0.4\textwidth, trim={0 0 0 {0.25\height}}, clip]{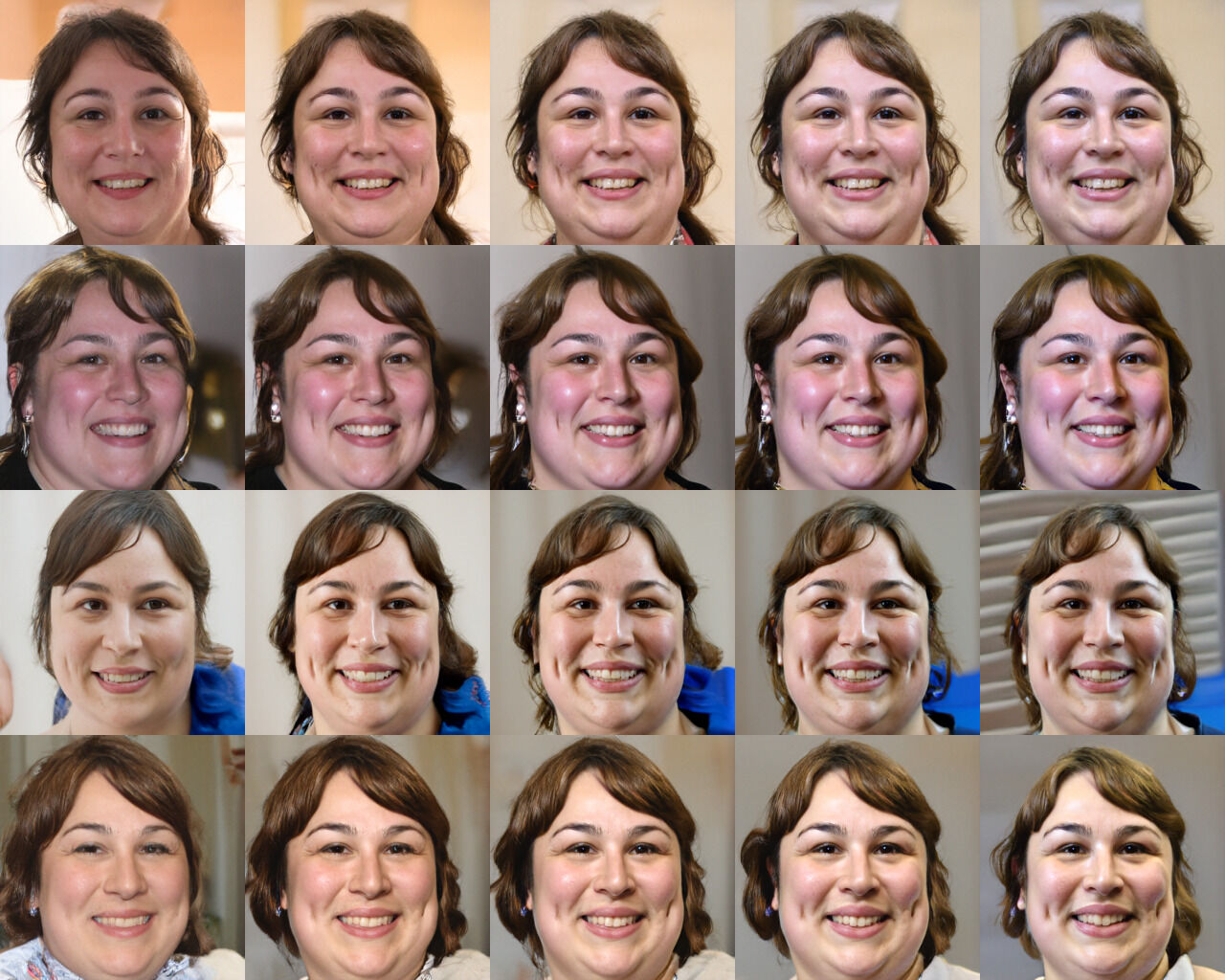} \\
    $s$ & \begin{tabularx}{0.375\textwidth}{ *{5}{Y} } $1.0$ & $2.0$ & $3.0$ & $4.0$ & $5.0$ \end{tabularx} \\
    \end{tabular}
    }
    \addtolength{\tabcolsep}{5pt}
\caption{Effect of the guidance scale on the generated images. For the (InsightFace~\cite{insightface}) ID vector extracted from the image on the left, we generate images for four seeds at guidance scales $s$ ranging from $1.0$ to $5.0$.}
\label{fig:qualitative_guidance}
\end{figure}

The classifier-free guidance scale offers control over the trade-off between the fidelity and diversity of the generated results. As seen in \cref{fig:qualitative_guidance}, by increasing the guidance, the generated faces converge to the same identity, resemble the original face more closely, and contain fewer artifacts. At the same time, higher guidance values reduce the diversity of identity-agnostic features such as the background and expressions and also increase contrast and saturation. 

To measure this effect quantitatively, we perform the same evaluation as in the previous section and report the results in \cref{table:diversity_eval}. As the guidance scale increases, the identity preservation improves as indicated by the decreasing identity distances, but the diversity in terms of poses, expressions, and LPIPS~\cite{lpips} features decreases. In practice, we choose a guidance scale of $2.0$ for all experiments unless stated otherwise because that appears to be the best compromise between image quality and diversity. In the supplementary material, we further show FID~\cite{fid} as well as precision and recall~\cite{recall_precision} values that measure how well the image distribution is preserved as the guidance scale varies.

\subsubsection{Identity vector latent space}

As described in \cref{sec:method}, we can find custom directions in the ID vector latent space that enable us to smoothly interpolate identities as well as change features such as the age or hair color as seen in \cref{fig:teaser_small} and in the supplementary material. Note that we refer to these features as \emph{identity-specific} because they exist in the ID vector latent space. In theory, this space should not contain any identity-agnostic information such as the pose. In practice, however, some FR methods inadvertently do extract this information. This is shown in great detail in the supplementary material, where we show an interesting application of our method to analyze pre-trained face recognition methods.

\subsubsection{Attribute conditioning}

\begin{figure}[htpb]
    \centering
    \addtolength{\tabcolsep}{-4pt}
    \small{
    \begin{tabular}{lc}
    ID & \raisebox{-.45\height}{\adjincludegraphics[width=.38\textwidth, trim={0 0 0 0}, clip]{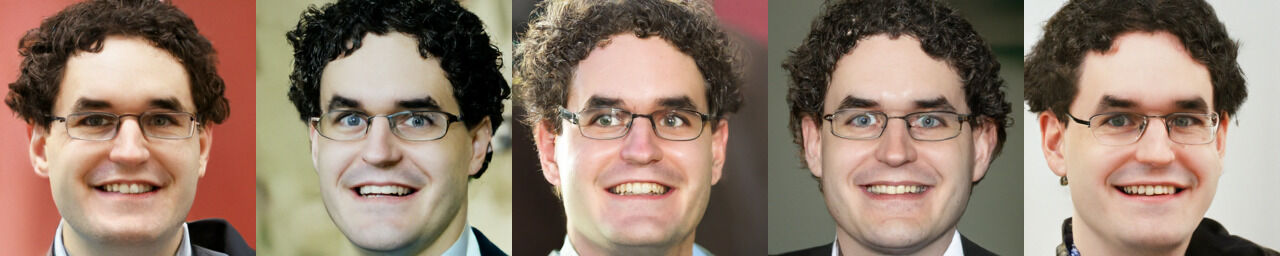}} \\
    \\[-0.35cm]
    ID + Set 1 & \raisebox{-.45\height}{\adjincludegraphics[width=.38\textwidth, trim={0 0 0 0}, clip]{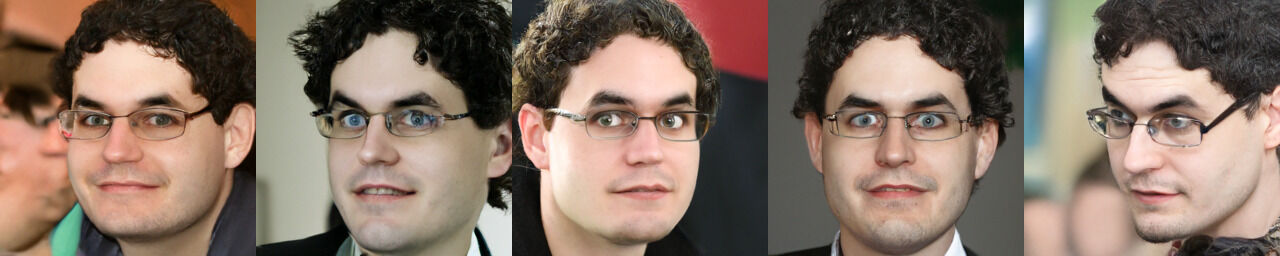}} \\
    \end{tabular}
    }
    \addtolength{\tabcolsep}{4pt}
    \caption{Attribute conditioning diversity. Through additional attribute conditioning, we can disentangle identity-specific and identity-agnostic features. As a result, we obtain more diverse results when using both (InsightFace~\cite{insightface}) ID vector and attribute vector conditioning (set 1) compared to when only using ID vector conditioning.}
    \label{fig:attribute_cond_diversity}
\end{figure}

By additionally conditioning our method on attributes, we can disentangle identity-specific and identity-agnostic features. As seen in \cref{fig:attribute_cond_diversity}, the additional attribute conditioning allows us to recover more of the original data distribution in terms of head poses and expressions whereas a model conditioned only on the ID vector is more likely to overfit and learn biases from the training data set. This is also shown in \cref{table:diversity_eval}, where the diversity increases with attribute conditioning at the expense of worse identity preservation compared to the base configuration. The attribute conditioning also enables intuitive control over the generated images by simply selecting the desired attribute values as shown in \cref{fig:teaser_small} and in the supplementary material.

\section{Limitations} \label{sec:limitations}

Our method outputs images at a relatively low resolution of $64 \times 64$. While this can be upsampled using super-resolution models, some fine identity-specific details such as moles cannot be modeled currently (but this information might not even be stored in the ID vector). Our method also has relatively long inference times ($15$ seconds per image when using batches of $16$ images on one NVIDIA RTX 3090 GPU) in the default setting, but this can be reduced to less than one second per image when using $10$ respacing steps at a slight decrease in quality as shown in the supplementary material. Our method also occasionally has small image generation artifacts, but the above aspects are expected to improve with future advancements in diffusion models. Lastly, our model inherits the biases of both the face recognition model and the training data set. This can manifest as either accessorizing images corresponding to certain demographic factors (\eg via make-up, clothing) or losing identity fidelity for underrepresented groups as shown in the supplementary material. This suggests an additional application of our work to the study of systematic biases in otherwise black-box systems.

\section{Ethical concerns} \label{sec:ethics}

All individuals portrayed in this paper provided informed consent to use their images as test images. This was not possible for the images from the FFHQ~\cite{stylegan}, LFW~\cite{lfw}, AgeDB-30~\cite{agedb}, and CFP-FP~\cite{cfpfp} data sets. Therefore, we do not show them in the paper and cannot provide qualitative comparisons to Vec2Face~\cite{vec2face} (code not available).

We recognize the potential for misuse of any method that creates realistic imagery of human beings, especially when the images are made to correspond to specific individuals.  We condemn such misuse and support ongoing research into the identification of artificially manipulated data.

\section{Conclusion} \label{sec:conclusion}

We propose a method to generate high-quality identity-preserving face images by injecting black-box, low-dimensional embeddings of a face into the residual blocks of a diffusion model. We mathematically reason and empirically show that our method produces images close to the target identity despite the absence of any identity-specific loss terms. Our method obtains state-of-the-art performance on identity preservation and output diversity, as demonstrated qualitatively and quantitatively. We further showcase advantages of our approach in providing control over the generation process. 
We thus provide a useful tool to create data sets with user-defined variations in identities and attributes as well as to analyze the latent spaces of face recognition methods, motivating more research in this direction.

\clearpage

\renewcommand{\refname}{References\footnote{Note: The numbering of the references in the arXiv version differs from that of the official versions.}}

{\small
\bibliographystyle{ieee_fullname}
\bibliography{egbib}
}

\clearpage
\setcounter{footnote}{0}
\onecolumn
\appendix

\section{Additional implementation details} \label{sec:add_implementation_details}

\paragraph{ID vectors} \Cref{table:id_vectors} lists the face recognition methods used in this work. Note that we use two implementations for ArcFace~\cite{arcface} and FaceNet~\cite{facenet}, one for training and the other one for evaluation in each case. The methods used for training were chosen to match those used in Gaussian sampling~\cite{gaussian_sampling} and StyleGAN search~\cite{stylegan-search} respectively. The methods used for evaluation were chosen to match the verification accuracy on real images as closely as possible to the values shown in Vec2Face~\cite{vec2face} to enable a fair comparison. For both evaluation methods, we extract the identity embeddings for each image as well as its horizontally flipped version and then calculate the angular distance of the concatenated identity embeddings after subtracting the mean embedding, similar to \cite{facenet_tf}. In order to avoid having the face detector stage of different face recognition vectors influence the qualitative results, we manually confirmed that all shown test images were properly aligned. 

\begin{table}[h!]
\centering
    \small{
    \begin{tabular}{lllll} 
    \toprule
    Method & Usage & Alignment & Implementation & Checkpoint \\ 
    \midrule
    AdaFace~\cite{adaface} & Training $^1$ & Provided MTCNN~\cite{mtcnn} & Official GitHub repository & ``adaface\_ir50\_ms1mv2.ckpt'' \\
    ArcFace~\cite{arcface, arcface_torch} & Evaluation & RetinaFace~\cite{retinaface} from~\cite{insightface} & Official GitHub repository & ``ms1mv3\_arcface\_r100\_fp16'' \\
    ArcFace~\cite{arcface, gaussian_sampling} & Training $^1$ & MTCNN~\cite{mtcnn} from \cite{facenet_pytorch} & From Razzhigaev \etal~\cite{gaussian_sampling} & ``torchtest.pt'' \\
    FaceNet~\cite{facenet, facenet_tf} & Evaluation & Provided MTCNN~\cite{mtcnn} & From David Sandberg~\cite{facenet_tf} & ``20180402-114759'' \\
    FaceNet~\cite{facenet, facenet_pytorch} & Training & Provided MTCNN~\cite{mtcnn} & From Tim Esler~\cite{facenet_pytorch} & ``20180402-114759'' \\
    FROM~\cite{from} & Training $^1$ & MTCNN~\cite{mtcnn} from 
    \cite{facenet_pytorch} & Official GitHub repository & ``model\_p5\_w1\_9938\_9470\_6503.pth.tar'' \\
    InsightFace~\cite{insightface} & Training & Provided RetinaFace~\cite{retinaface} & InsightFace repository~\cite{insightface} & ``buffalo\_l'' \\
    \bottomrule
    \end{tabular}
}
\caption{Overview over the considered face recognition methods. $^1$ Only used in supplementary material.}
\label{table:id_vectors}
\end{table}

\paragraph{Model} We use the official U-net~\cite{unet} implementation by Dhariwal and Nichol~\cite{ddpm3, ddpm3_repo} and their recommended hyperparameters, whenever applicable, for the main $64 \times 64$ ID-conditioned face generation model and the $64 \rightarrow 256$ super-resolution model as listed in \cref{table:hyperparameters}. The U-net architecture is divided into several levels, with each level composed of ResNet~\cite{resnet} blocks and down- or upsampling layers. The U-net also contains global attention layers at $32 \times 32$, $16 \times 16$, and $8 \times 8$ resolutions. The time step $t$ is passed through a sinusoidal position embedding layer, known from transformers~\cite{transformer}, and is then added to the residual connection of the ResNet blocks. The most important additions to the baseline model are the identity conditioning module (identity\_cond) and introducing classifier-free guidance (classifier\_free) by setting the conditioning vector to the $0$-vector\footnote{For attribute conditioning, the $-1$-vector is used since the $0$-vector is a valid attribute vector (\eg age $0$) and the $-1$-vector performed better empirically.} for $10\%$ of the training samples to obtain an unconditional and conditional setting with just one trained model.

\paragraph{Training} The training set is composed of images along with their corresponding (pre-computed) ID vectors. We train the $64 \times 64$ ID-conditioned face generation model for $100000$ batches and the $64 \times 64$ unconditional upsampling model for $50000$ batches, both with a batch size of $64$, learning rate of $10^{-4}$, and from scratch. We use the weights with an exponential moving average rate of $0.9999$ because it generally leads to better results. Training takes around two days on one NVIDIA RTX 3090 GPU.

\begin{figure*}[htbp]
\centering
\begin{subfigure}{\textwidth}
\centering
    \addtolength{\tabcolsep}{-5pt}
    \small{
    \begin{tabular}{cc}
    \raisebox{1.5\height}{\includegraphics[width=0.095\textwidth]{images/original/Martina.jpg}} & 
    \adjincludegraphics[width=0.855\textwidth, trim={0 0 0 0}, clip]{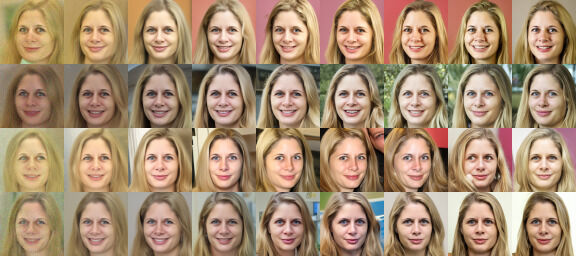} \\ 
    Time (s) & \begin{tabularx}{0.855\textwidth}{ *{9}{Y} } $0.2$ & $0.2$ & $0.3$ & $0.6$ & $0.9$ & $1.8$ & $4.1$ & $8.3$ & $17.0$ \end{tabularx} \\
    \\
    Steps & \begin{tabularx}{0.855\textwidth}{ *{9}{Y} } $3$ & $5$ & $10$ & $25$ & $50$ & $100$ & $250$ & $500$ & $1000$ \end{tabularx} \\
    \end{tabular}
    }
    \addtolength{\tabcolsep}{5pt}
    \caption{$64 \times 64$ main model}
    \vspace{5mm}
\end{subfigure}
\begin{subfigure}{\textwidth}
\centering
    \addtolength{\tabcolsep}{-5pt}
    \small{
    \begin{tabular}{cc}
    \raisebox{1.5\height}{\includegraphics[width=0.095\textwidth]{images/original/Martina.jpg}} & 
    \adjincludegraphics[width=0.855\textwidth, trim={0 0 0 0}, clip]{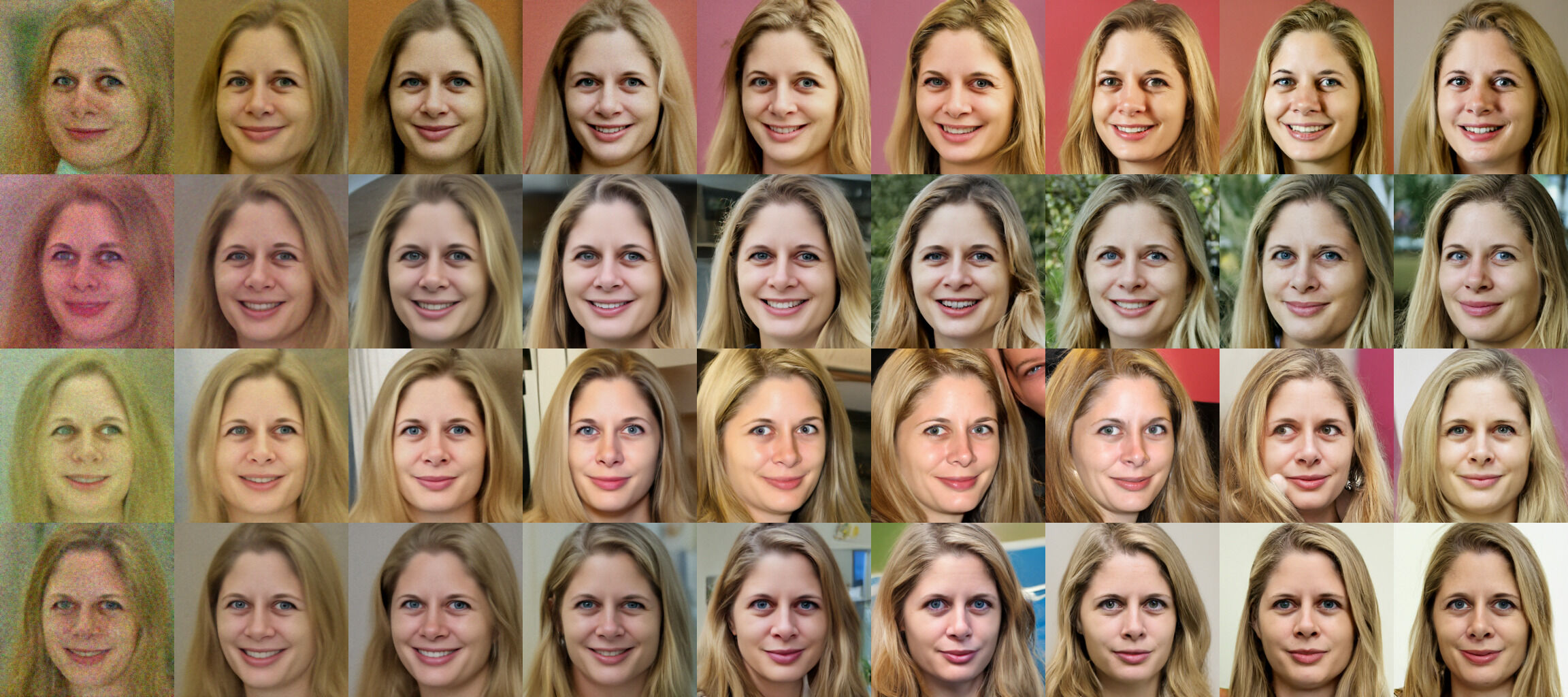} \\ 
    Time (s) & \begin{tabularx}{0.855\textwidth}{ *{9}{Y} } $0.6$ & $0.6$ & $0.9$ & $1.8$ & $3.3$ & $6.3$ & $15.1$ & $30.1$ & $60.4$ \end{tabularx} \\
    \\
    Steps & \begin{tabularx}{0.855\textwidth}{ *{9}{Y} } $3$ & $5$ & $10$ & $25$ & $50$ & $100$ & $250$ & $500$ & $1000$ \end{tabularx} \\
    \end{tabular}
    }
    \addtolength{\tabcolsep}{5pt}
    \caption{$64 \times 64$ main model + $64 \times 64 \rightarrow 256 \times 256$ super-resolution model}
\end{subfigure}
\caption{Qualitative evaluation of the effect of the number of respacing steps. For each ID vector extracted from the image on the left, we generate images for four seeds with different number of respacing steps, with and without super-resolution. We also report the inference time per image (when using batches of 16 images) in seconds on one NVIDIA RTX 3090 GPU. Note that the time listed for the images with super-resolution also includes the time to generate the $64 \times 64$ images.}
\label{fig:ablation_time_steps}
\end{figure*}

\paragraph{Inference} All models are trained with $T=1000$ but respaced to $250$ time steps during inference for computational reasons with a negligible decrease in quality. We use a classifier-free guidance scale of $2$ unless otherwise stated. Furthermore, we fix the randomness seeds whenever comparing different methods to ensure a fair comparison. Inference (main model + super-resolution to $256 \times 256$ resolution) takes around $15$ seconds per image when using batches of $16$ images on one NVIDIA RTX 3090 GPU. The inference time can be drastically reduced by using fewer respacing steps at a slight decrease in quality, as shown in \cref{fig:ablation_time_steps}. For example, when using $10$ respacing steps, the inference time decreases to around $1$ second per image with a comparable identity fidelity and only slightly fewer details (especially in the background).

\begin{table}[h!]
\centering
    \small{
    \begin{tabular}{lrr} 
    \toprule
    & $64 \times 64$ main model & $64 \times 64 \rightarrow 256 \times 256$ super-resolution model \\ 
    \midrule
    \underline{Diffusion parameters} \\
    diffusion\_steps & $1000$ & $1000$ \\ 
    noise\_schedule & cosine & linear \\
    \\
    \underline{Model parameters} \\
    attention\_resolutions & $32, 16, 8$ & $32, 16, 8$ \\ 
    classifier\_free & True & False \\
    dropout & $0.1$ & $0$ \\ 
    identity\_cond & True & False \\
    learn\_sigma & True & True \\
    num\_channels & $192$ & $192$ \\ 
    num\_heads & $3$ & $4$ \\ 
    num\_res\_blocks & $3$ & $2$ \\ 
    resblock\_updown & True & True \\
    use\_fp16 & True & True \\
    use\_new\_attention\_order & True & False \\
    use\_scale\_shift\_norm & True & True \\
    \\
    \underline{Training parameters} \\
    batch\_size & $64$ & $64$ \\
    ema\_rate & $0.9999$ & $0.9999$ \\
    lr (learning rate) & $10^{-4}$ & $10^{-4}$ \\
    total\_steps (batches) & $100000$ & $50000$ \\
    \bottomrule
    \end{tabular}
}
\caption{Hyperparameters of our diffusion models. We use one diffusion model to generate $64 \times 64$ resolution images and one super-resolution diffusion model to increase the resolution to $256 \times 256$. All other parameters are named as in the baseline implementation (where applicable).}
\label{table:hyperparameters}
\end{table}

\clearpage

\clearpage

\section{Additional comparisons to state-of-the-art methods}

\subsection{Qualitative results}

\begin{figure*}[htpb]
    \centering
    \addtolength{\tabcolsep}{-5pt}
    \small{
    \begin{tabular}{cccccc}
        \includegraphics[width=0.13\textwidth]{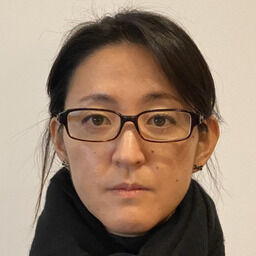} & 
        \includegraphics[width=0.13\textwidth]{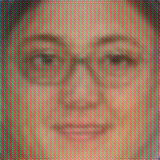} & 
        \includegraphics[width=0.13\textwidth]{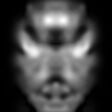} & 
        \includegraphics[width=0.13\textwidth]{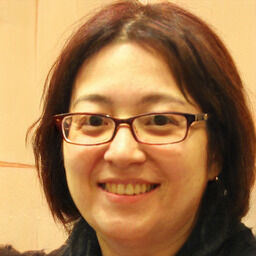} & 
        \includegraphics[width=0.13\textwidth]{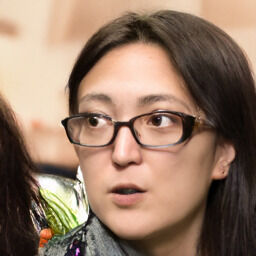} & 
        \includegraphics[width=0.13\textwidth]{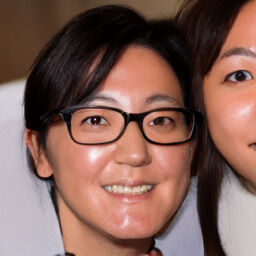} \\
        \\[-0.46cm]   
        \includegraphics[width=0.13\textwidth]{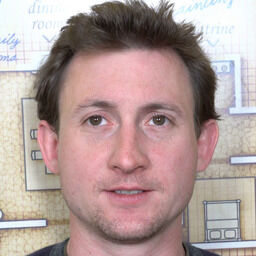} & 
        \includegraphics[width=0.13\textwidth]{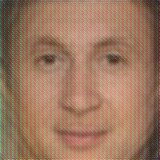} & 
        \includegraphics[width=0.13\textwidth]{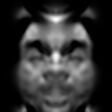} & 
        \includegraphics[width=0.13\textwidth]{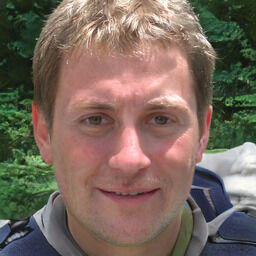} & 
        \includegraphics[width=0.13\textwidth]{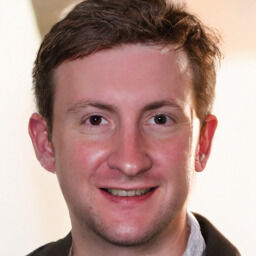} & 
        \includegraphics[width=0.13\textwidth]{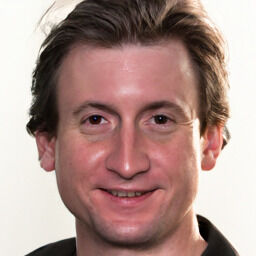} \\
        \\[-0.46cm]   
        \includegraphics[width=0.13\textwidth]{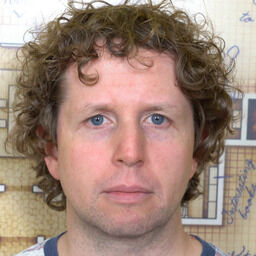} & 
        \includegraphics[width=0.13\textwidth]{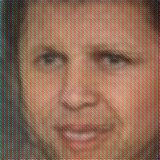} & 
        \includegraphics[width=0.13\textwidth]{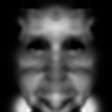} & 
        \includegraphics[width=0.13\textwidth]{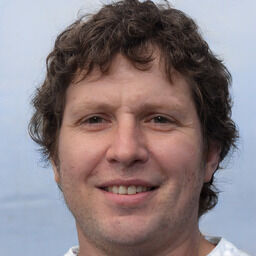} & 
        \includegraphics[width=0.13\textwidth]{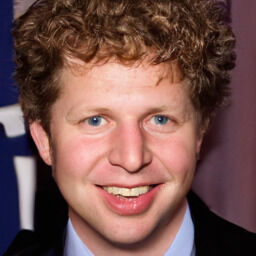} & 
        \includegraphics[width=0.13\textwidth]{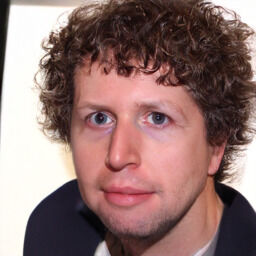} \\
        \\[-0.46cm]
        \includegraphics[width=0.13\textwidth]{images/original/Sandra.jpg} & 
        \includegraphics[width=0.13\textwidth]{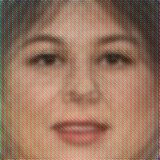} & 
        \includegraphics[width=0.13\textwidth]{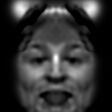} & 
        \includegraphics[width=0.13\textwidth]{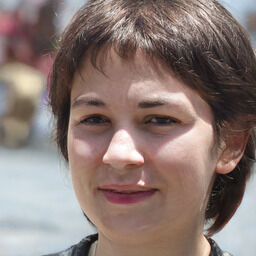} & 
        \includegraphics[width=0.13\textwidth]{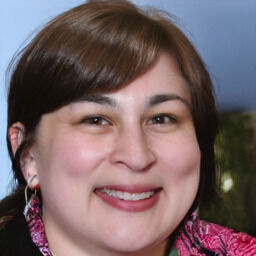} & 
        \includegraphics[width=0.13\textwidth]{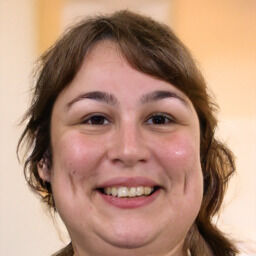} \\
        \\[-0.46cm]
        \includegraphics[width=0.13\textwidth]{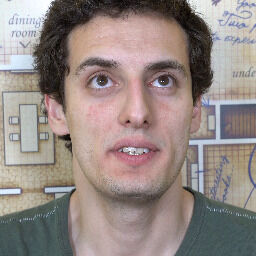} & 
        \includegraphics[width=0.13\textwidth]{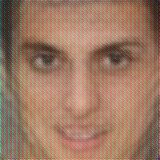} & 
        \includegraphics[width=0.13\textwidth]{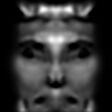} & 
        \includegraphics[width=0.13\textwidth]{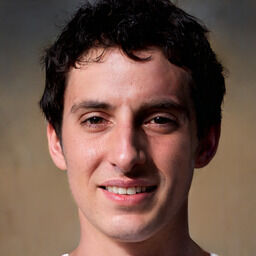} & 
        \includegraphics[width=0.13\textwidth]{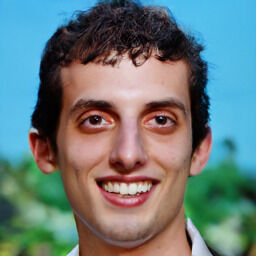} & 
        \includegraphics[width=0.13\textwidth]{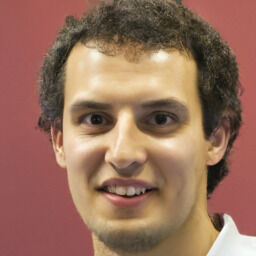} \\
        \\[-0.46cm]  
        \includegraphics[width=0.13\textwidth]{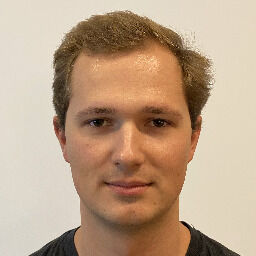} & 
        \includegraphics[width=0.13\textwidth]{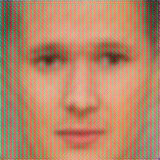} & 
        \includegraphics[width=0.13\textwidth]{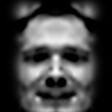} & 
        \includegraphics[width=0.13\textwidth]{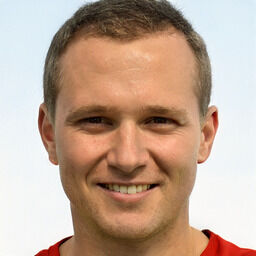} & 
        \includegraphics[width=0.13\textwidth]{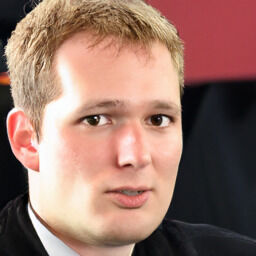} & 
        \includegraphics[width=0.13\textwidth]{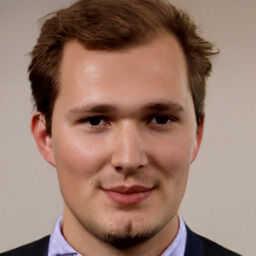} \\
        \\[-0.46cm]  
        \includegraphics[width=0.13\textwidth]{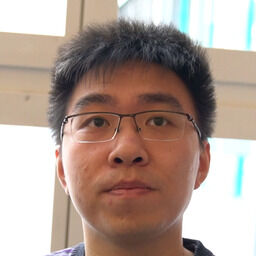} & 
        \includegraphics[width=0.13\textwidth]{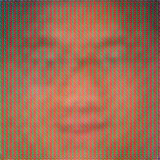} & 
        \includegraphics[width=0.13\textwidth]{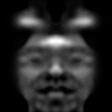} & 
        \includegraphics[width=0.13\textwidth]{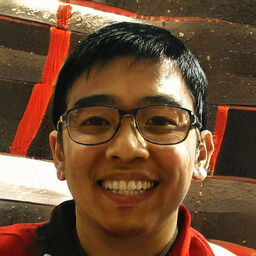} & 
        \includegraphics[width=0.13\textwidth]{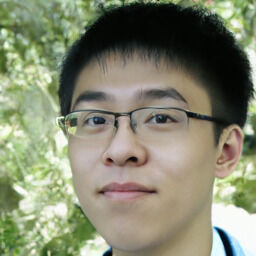} & 
        \includegraphics[width=0.13\textwidth]{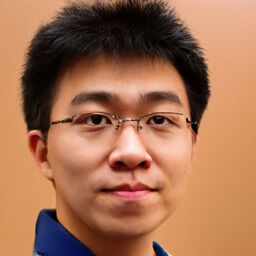} \\
        
        \multirow{2}{*}{\begin{tabular}[c]{@{}c@{}}Original \\ image \end{tabular}} & \multirow{2}{*}{NbNet~\cite{nbnet}} & \multirow{2}{*}{\begin{tabular}[c]{@{}c@{}}Gaussian \\ sampling~\cite{gaussian_sampling} \end{tabular}} & \multirow{2}{*}{\begin{tabular}[c]{@{}c@{}}StyleGAN \\ search~\cite{stylegan-search} \end{tabular}} & \multirow{2}{*}{\begin{tabular}[c]{@{}c@{}}ID3PM (Ours, \\ FaceNet~\cite{facenet, facenet_pytorch}) \end{tabular}} & \multirow{2}{*}{\begin{tabular}[c]{@{}c@{}}ID3PM (Ours, \\ InsightFace~\cite{insightface}) \end{tabular}} \\
        \\
    \end{tabular}
    }
    \addtolength{\tabcolsep}{5pt}
    \caption{Qualitative evaluation with state-of-the-art methods (additional results). The generated images of our method look realistic and resemble the identity of the original image more closely than any of the other methods. }
    \label{fig:comp}
\end{figure*}

\Cref{fig:comp} shows additional results of the qualitative comparison with the state-of-the-art black-box methods, whose code is available, and demonstrates the superiority of our method both in terms of image quality and identity preservation.

\clearpage

\begin{figure*}[htpb]
    \centering
    \addtolength{\tabcolsep}{-5pt}
    \small{
    \begin{tabular}{ccc}
        \raisebox{1.5\height}{\includegraphics[width=0.075\textwidth]{images/original/Derek.jpg}} & 
        \includegraphics[width=0.3\textwidth]{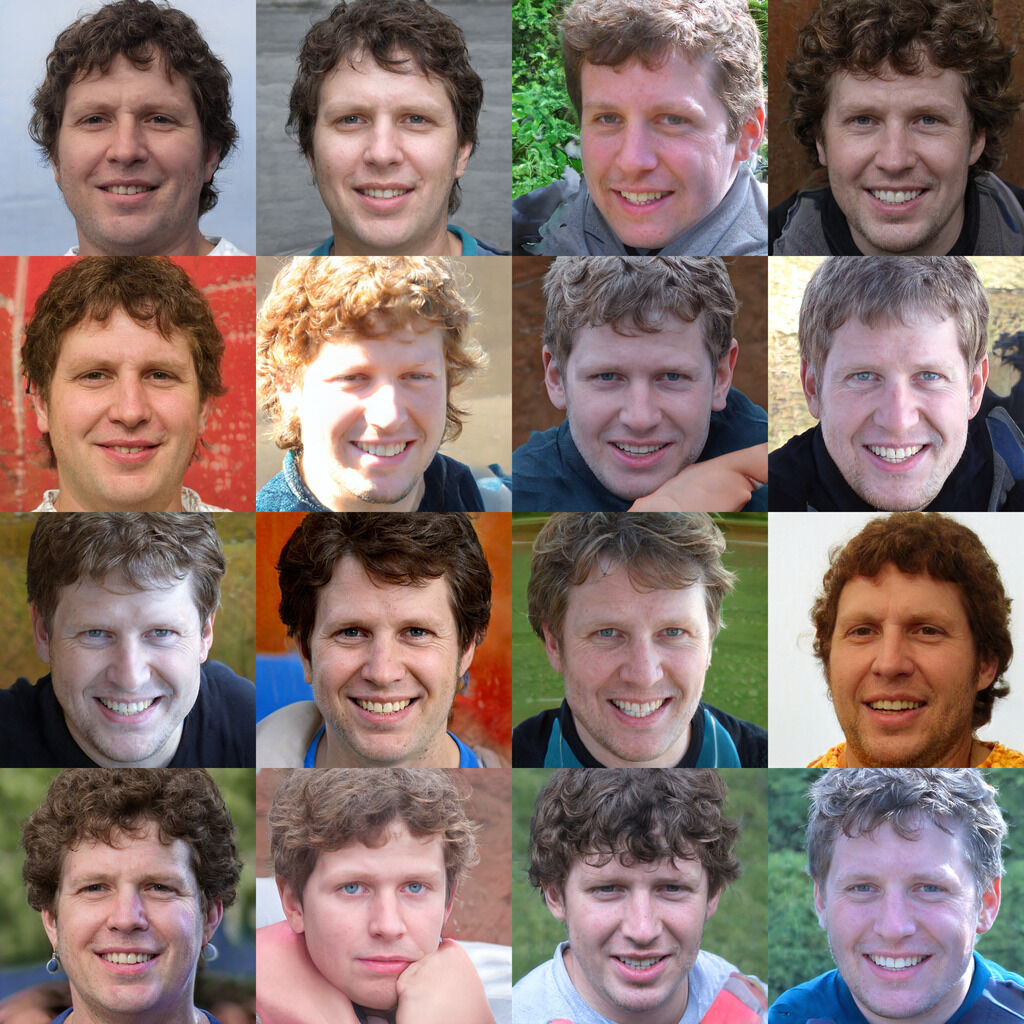} & 
        \includegraphics[width=0.3\textwidth]{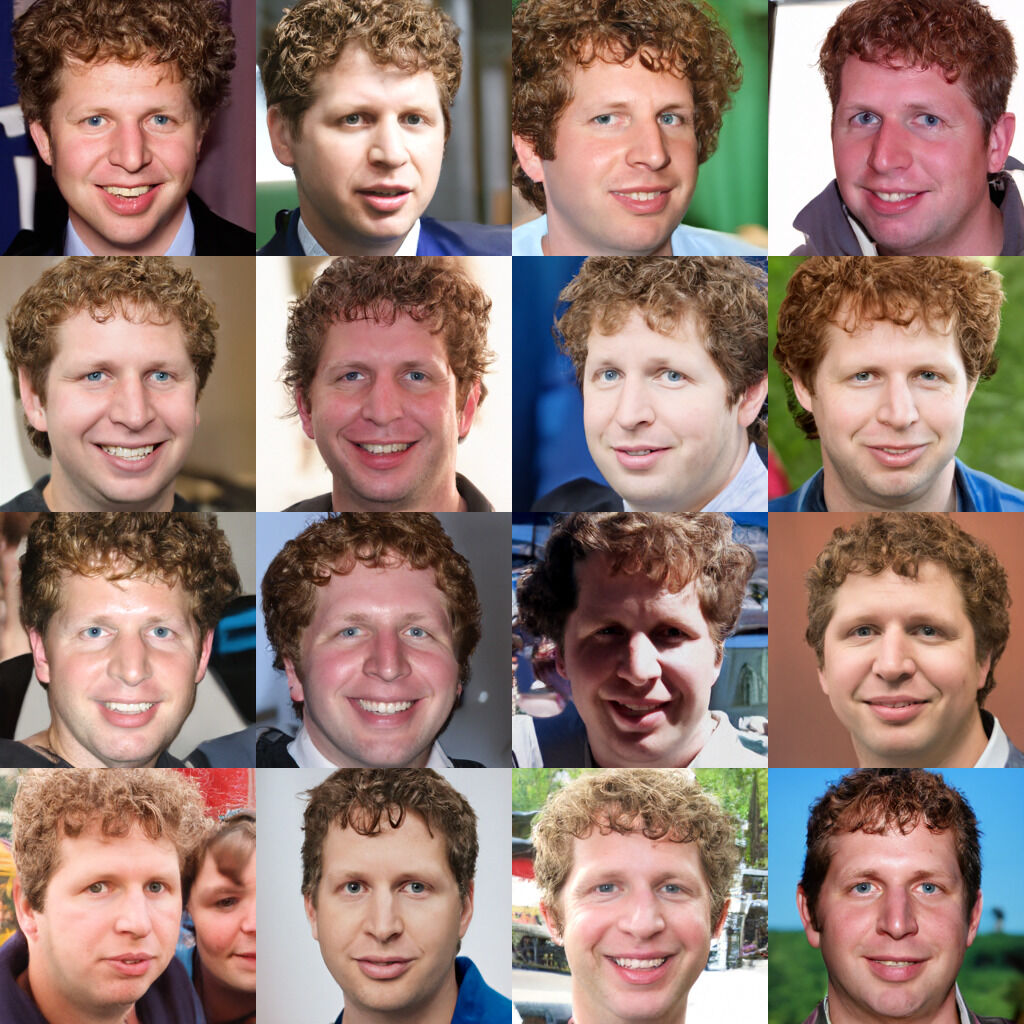} \\
        \\[-0.46cm]
        \raisebox{1.5\height}{\includegraphics[width=0.075\textwidth]{images/original/Sandra.jpg}} & 
        \includegraphics[width=0.3\textwidth]{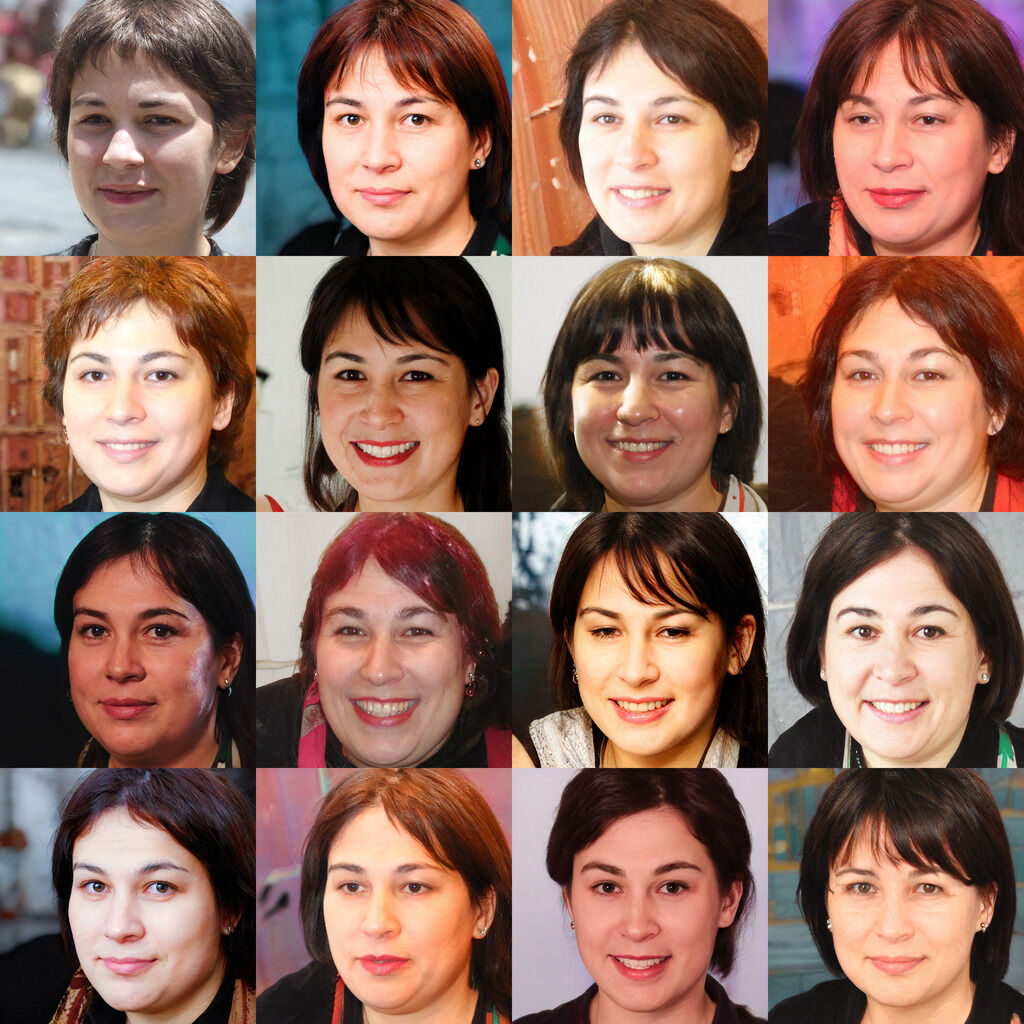} & 
        \includegraphics[width=0.3\textwidth]{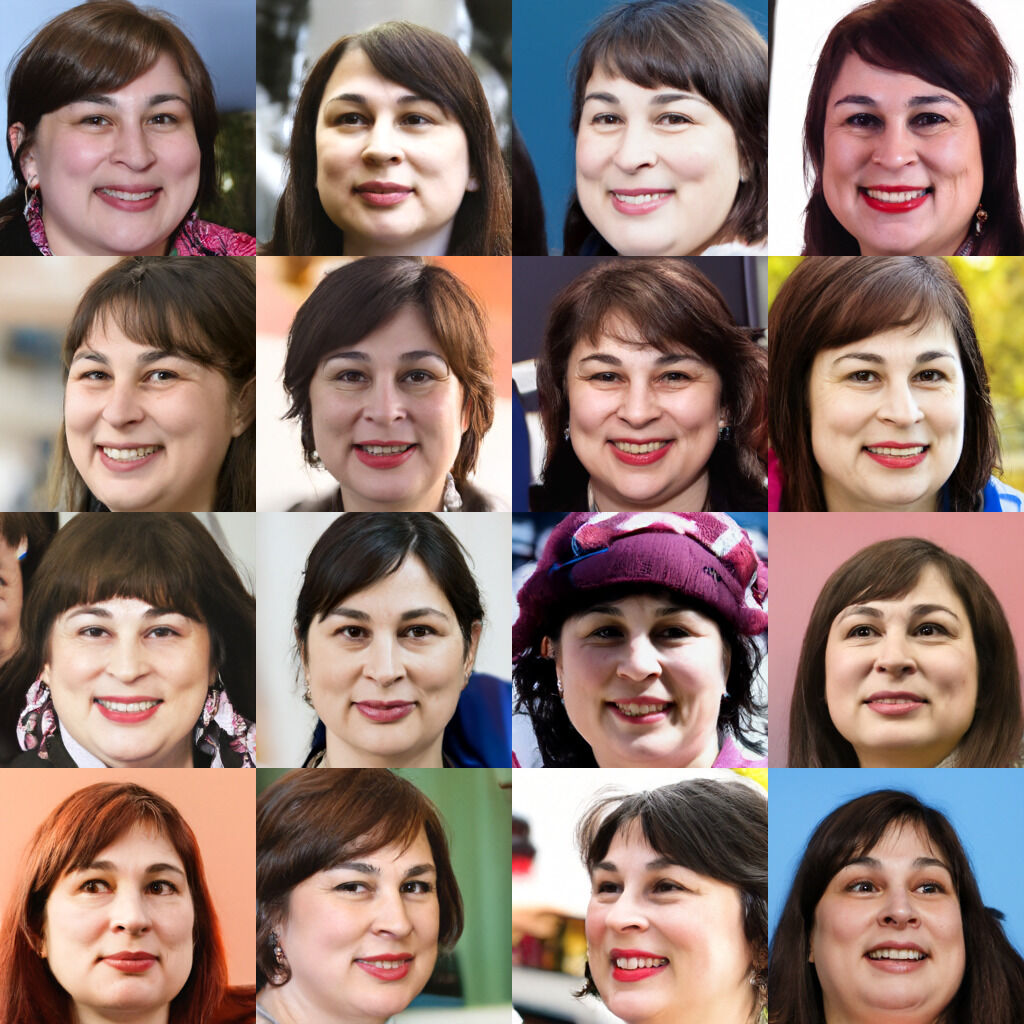} \\
        \\[-0.46cm]
        \raisebox{1.5\height}{\includegraphics[width=0.075\textwidth]{images/original/Yang.jpg}} & 
        \includegraphics[width=0.3\textwidth]{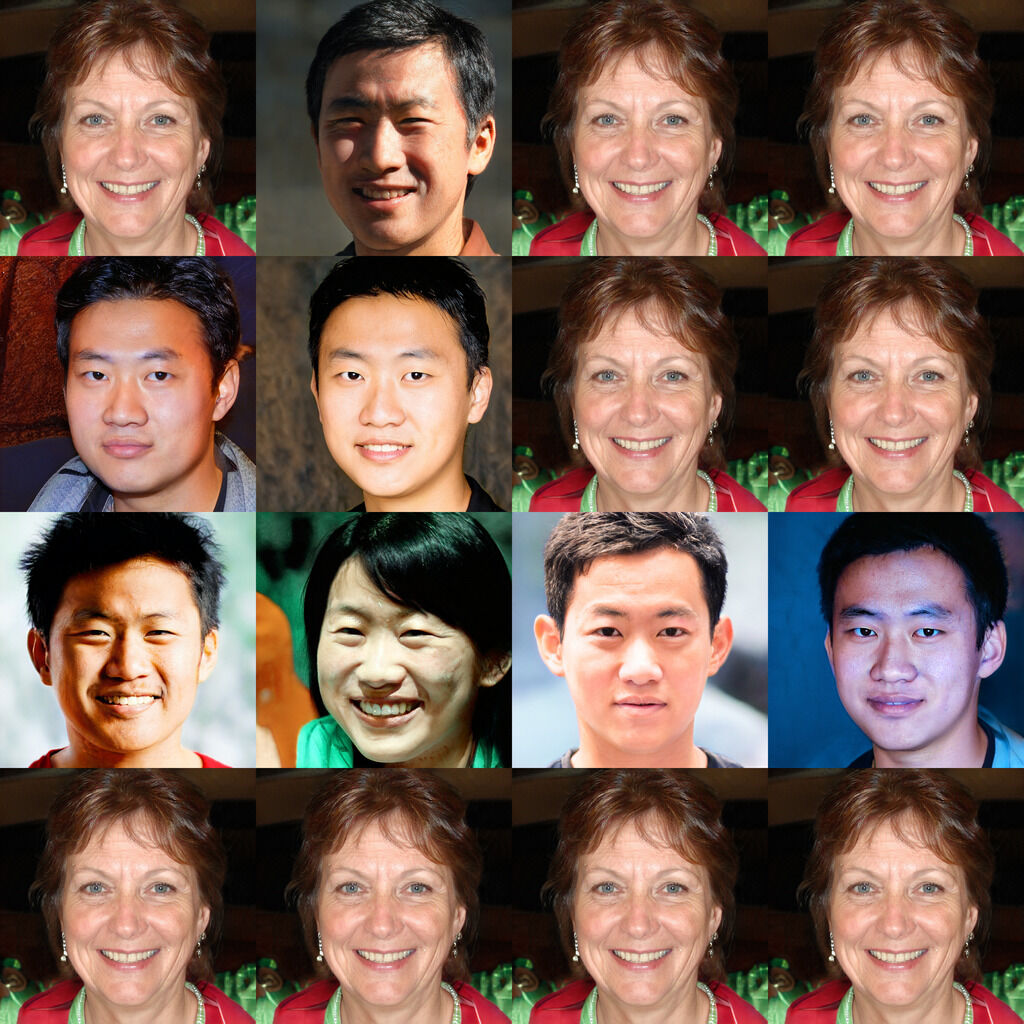} & 
        \includegraphics[width=0.3\textwidth]{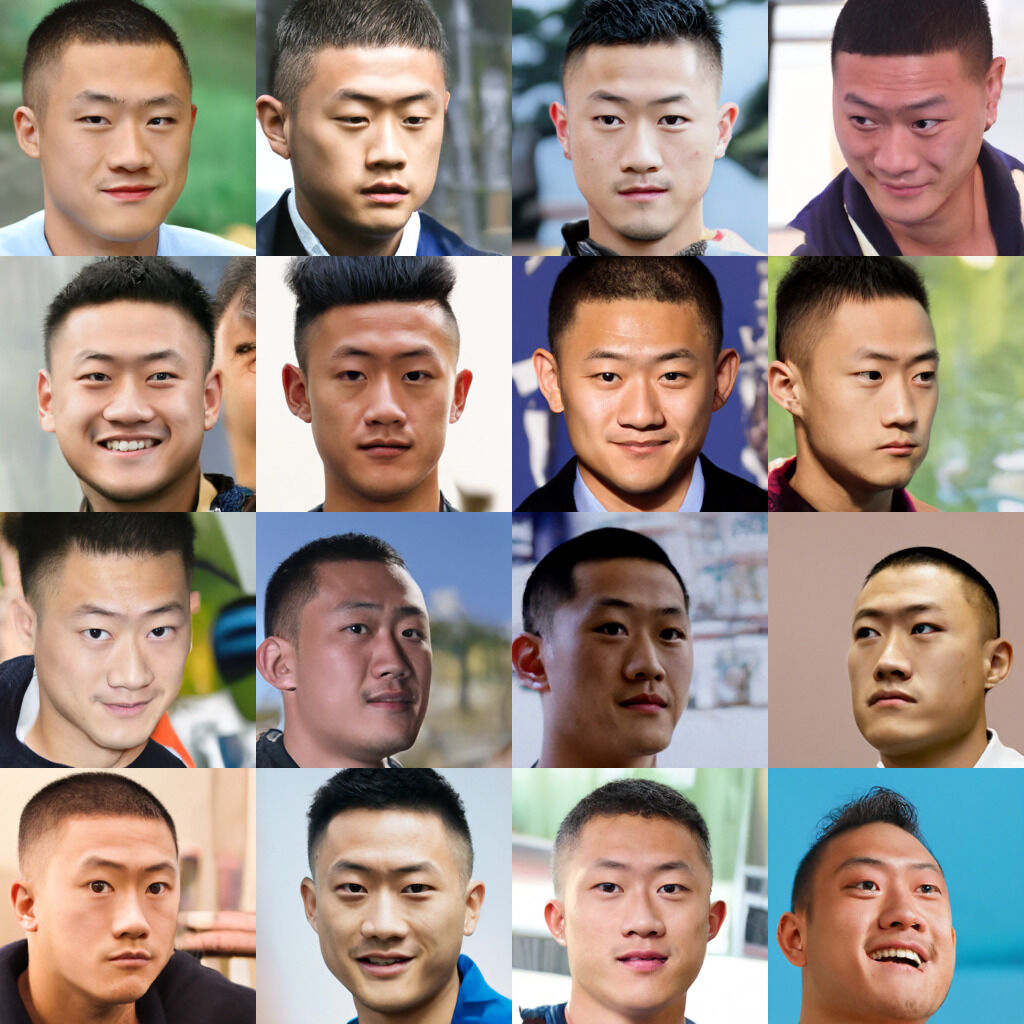} \\
        
        Original image & StyleGAN search~\cite{stylegan-search} & ID3PM (Ours, FaceNet~\cite{facenet, facenet_pytorch}) \\
        \\
    \end{tabular}
    }
    \addtolength{\tabcolsep}{5pt}
    \caption{Qualitative evaluation with StyleGAN search~\cite{stylegan-search}. The generated images of our method resemble the identity of the original image closely and more consistently. Note that StyleGAN search~\cite{stylegan-search} often fails completely for the third identity, whereas our method trained with the same face recognition method (FaceNet~\cite{facenet, facenet_pytorch}) reproduces the identity well.}
    \label{fig:comp_stylegan}
\end{figure*}

In the main paper, we quantitatively compare the diversity of our method with that of StyleGAN search~\cite{stylegan-search}, which is the only competing method that can produce realistic images at a high resolution. 
\Cref{fig:comp_stylegan} qualitatively confirms that our method produces similarly diverse images but with better identity preservation. For fairness reasons, we use our model trained with FaceNet~\cite{facenet, facenet_pytorch} ID vectors since StyleGAN search~\cite{stylegan-search} uses the same FaceNet implementation~\cite{facenet_pytorch}. For the first two identities, the  StyleGAN search~\cite{stylegan-search} algorithm finds images that share facial features with the original face; however, the identity does not resemble the original face very closely. For the third identity, the search strategy often fails completely by landing in local minima. 

\clearpage

\subsection{User study}

To accompany the qualitative results and since we cannot show images from public data sets without the individuals' written consents (as explained in the ethics section of the main paper), we performed a user study with two parts. For the first part, we took the first 10 positive pairs with unique identities according to the LFW~\cite{lfw} protocol (same protocol as used for the quantitative evaluation of the identity preservation in the main paper) to compare the following methods: NbNet~\cite{nbnet}, Gaussian sampling~\cite{gaussian_sampling}, StyleGAN search~\cite{stylegan-search}, ours with FaceNet~\cite{facenet, facenet_pytorch}, and ours with InsightFace~\cite{insightface}. 

Specifically, we instructed the users to:
\begin{itemize}[itemsep=-2pt,topsep=2pt]
    \item Rank the generated images from most \textbf{similar to the person of the input image} to least.
    \item Rank the generated images from most \textbf{realistic} to least.
\end{itemize}

The results in \Cref{table:user_study} (left) provide further evidence of our method outperforming competitor methods, with the exception of StyleGAN search, which achieves better average realism at the expense of identity preservation. The user study also supports our realism labels in the related work section of the main paper.

\begin{table}[h!]
\centering
    \small{
    \subfloat{
    \begin{tabular}{lcc}
    \toprule
    \multirow{2}{*}{Method} & \multicolumn{2}{c}{LFW} \\
    \cmidrule{2-3}
    & {ID $\downarrow$} & {Real $\downarrow$} \\
    \midrule
    NbNet~\cite{nbnet} & 3.52 & 4.06 \\
    Gaussian sampling~\cite{gaussian_sampling} & 4.83 & 4.90 \\
    StyleGAN search~\cite{stylegan-search} & 2.53 & \textbf{1.39} \\
    \midrule
    ID3PM (Ours, FaceNet~\cite{facenet, facenet_pytorch}) & \textbf{1.90} & 2.05 \\
    ID3PM (Ours, InsightFace~\cite{insightface}) & 2.22 & 2.61 \\
    \bottomrule
    \end{tabular}}
    \quad
    \subfloat{
    \begin{tabular}{lcc}
    \toprule
    \multirow{2}{*}{Method} & \multicolumn{2}{c}{Vec2Face images} \\
    \cmidrule{2-3}
    & {ID $\downarrow$} & {Real $\downarrow$} \\
    \midrule
    Vec2Face~\cite{vec2face} & 3.51 & 3.80 \\
    \midrule
    ID3PM (Ours, FaceNet~\cite{facenet, facenet_pytorch}) & 2.665& \textbf{1.52} \\
    ID3PM (Ours, InsightFace~\cite{insightface}) & 3.50 & 3.23 \\
    ID3PM (Ours, FaceNet~\cite{facenet, facenet_pytorch}, CASIA-WebFace~\cite{casiawebface}) & 2.87 & 3.10 \\
    ID3PM (Ours, InsightFace~\cite{insightface}, CASIA-WebFace~\cite{casiawebface}) & \textbf{2.46} & 3.36 \\
    \bottomrule
    \end{tabular}}}
\caption{User study. The listed scores are the mean ranks (1 - 5) for realism (Real) and identity preservation (ID) of the different methods on LFW~\cite{lfw} images (left) and Vec2Face~\cite{vec2face} images (right).}
\label{table:user_study}
\end{table}

For the second part, we took screenshots of the 14 input and result images from Fig.\ 4 of the Vec2Face~\cite{vec2face} paper to compare our method to Vec2Face despite their code not being available. We then computed ID vectors for these faces and generated one image per ID vector with each variation of our method with a fixed random seed and ran a similar user study as above, again with 25 users. We also trained versions of our method on CASIA-WebFace~\cite{casiawebface}\footnote{Not upscaled from $64 \times 64$ to $256 \times 256$ for time reasons.} to have the same training data as Vec2Face. The results in \Cref{table:user_study} show that all variations of our method beat Vec2Face despite the experimental setup favoring Vec2Face (\eg low-quality screenshots as input for our method and using Vec2Face authors' chosen examples).

\subsection{Fairness of comparisons}

Our comparisons are fair or to the benefit of competing methods, and no retraining of competing methods was necessary.
Gaussian sampling~\cite{gaussian_sampling} does not have a training data set. 
StyleGAN search~\cite{stylegan-search} uses a StyleGAN2~\cite{stylegan2} trained on all 70000 images\footnote{See \url{https://tinyurl.com/ffhq70k}.} of FFHQ~\cite{stylegan}, so it saw the 1k test images used in the main paper during training (unlike our method that was trained only on the first 69000 images). 
NbNet~\cite{nbnet} and Vec2Face~\cite{vec2face} will likely not work well when trained only on FFHQ. NbNet reports significantly worse results in their Tab. 4 and section 4.2.1 when not augmenting their data set (which is already orders of magnitudes larger than FFHQ) with millions of images. Vec2Face uses CASIA-WebFace~\cite{casiawebface}, which is $\sim 7$ times bigger than FFHQ, and needs class information during training. One can thus consider Vec2Face as white-box with a slightly worse face recognition model (trained with knowledge distillation). When training our method with CASIA-WebFace instead of FFHQ, we obtain similar results and also match or outperform Vec2Face as seen in the above user study. Also, for fairness, we used Vec2Face's protocol for the quantitative comparison of the identity preservation. 
Lastly, note that no images visualized in the paper were included in any method's training data.

\clearpage

\section{Controllability}

To the best of our knowledge, our method is the first black-box face recognition model inversion method that offers intuitive control over the generation process. The mechanisms described in the following section enable the generation of data sets with control over variation and diversity of the identities as well as their attributes.

\subsection{Guidance scale}

\begin{figure*}[htbp]
\centering
\begin{subfigure}{\textwidth}
\centering
    \addtolength{\tabcolsep}{-5pt}
    \small{
    \begin{tabular}{cc}
    \raisebox{1.5\height}{\includegraphics[width=0.095\textwidth]{images/original/Sandra.jpg}} & 
    \adjincludegraphics[width=0.855\textwidth, trim={0 {.5\height} 0 {0.25\height}}, clip]{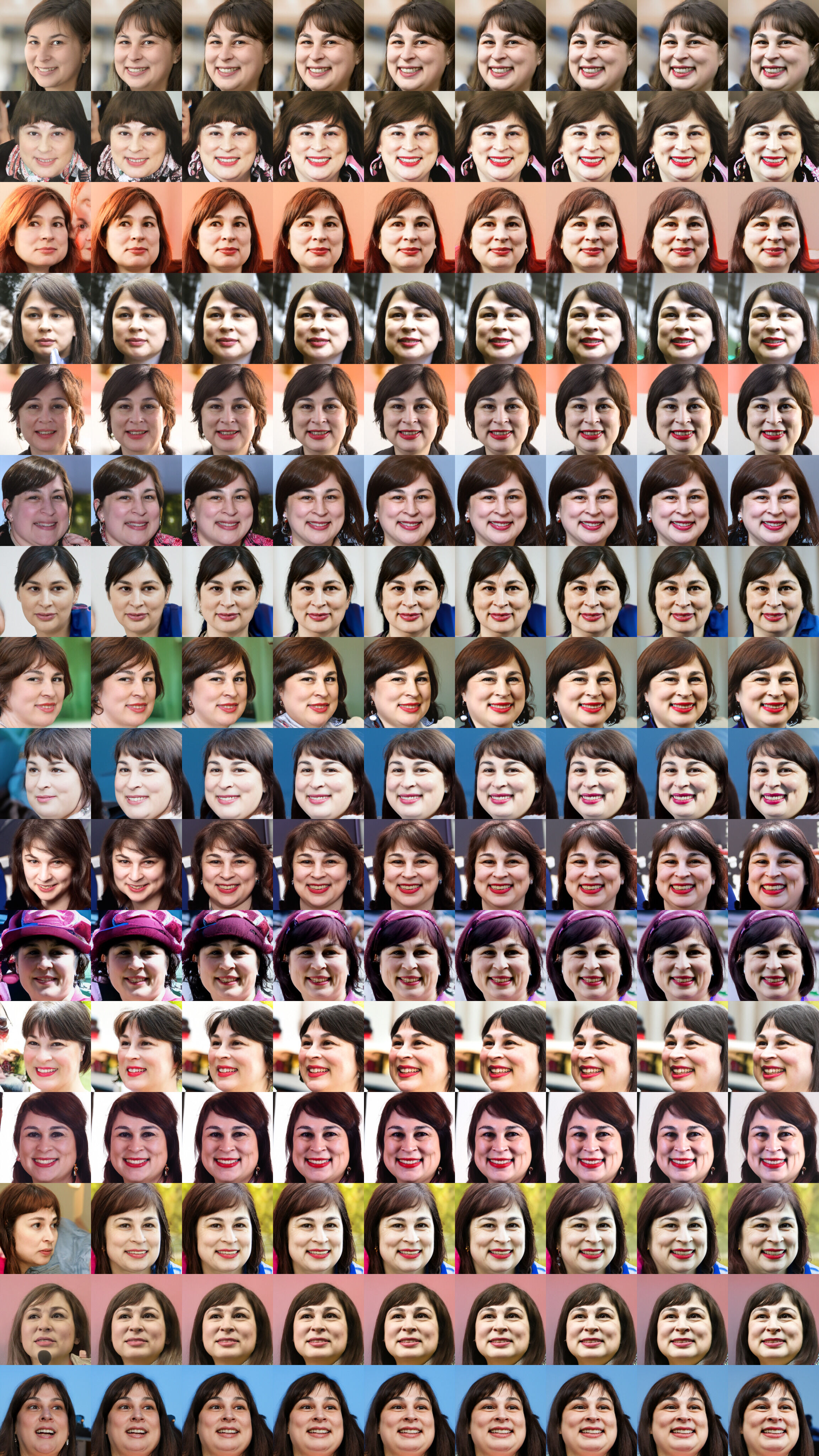} \\ 
    $s$ & \begin{tabularx}{0.855\textwidth}{ *{9}{Y} } $1.0$ & $1.5$ & $2.0$ & $2.5$ & $3.0$ & $3.5$ & $4.0$ & $4.5$ & $5.0$ \end{tabularx} \\
    \end{tabular}
    }
    \addtolength{\tabcolsep}{5pt}
    \caption{FaceNet~\cite{facenet, facenet_pytorch}}
    \vspace{5mm}
\end{subfigure}
\begin{subfigure}{\textwidth}
\centering
    \addtolength{\tabcolsep}{-5pt}
    \small{
    \begin{tabular}{cc}
    \raisebox{1.5\height}{\includegraphics[width=0.095\textwidth]{images/original/Sandra.jpg}} & 
    \adjincludegraphics[width=0.855\textwidth, trim={0 {.5\height} 0 {0.25\height}}, clip]{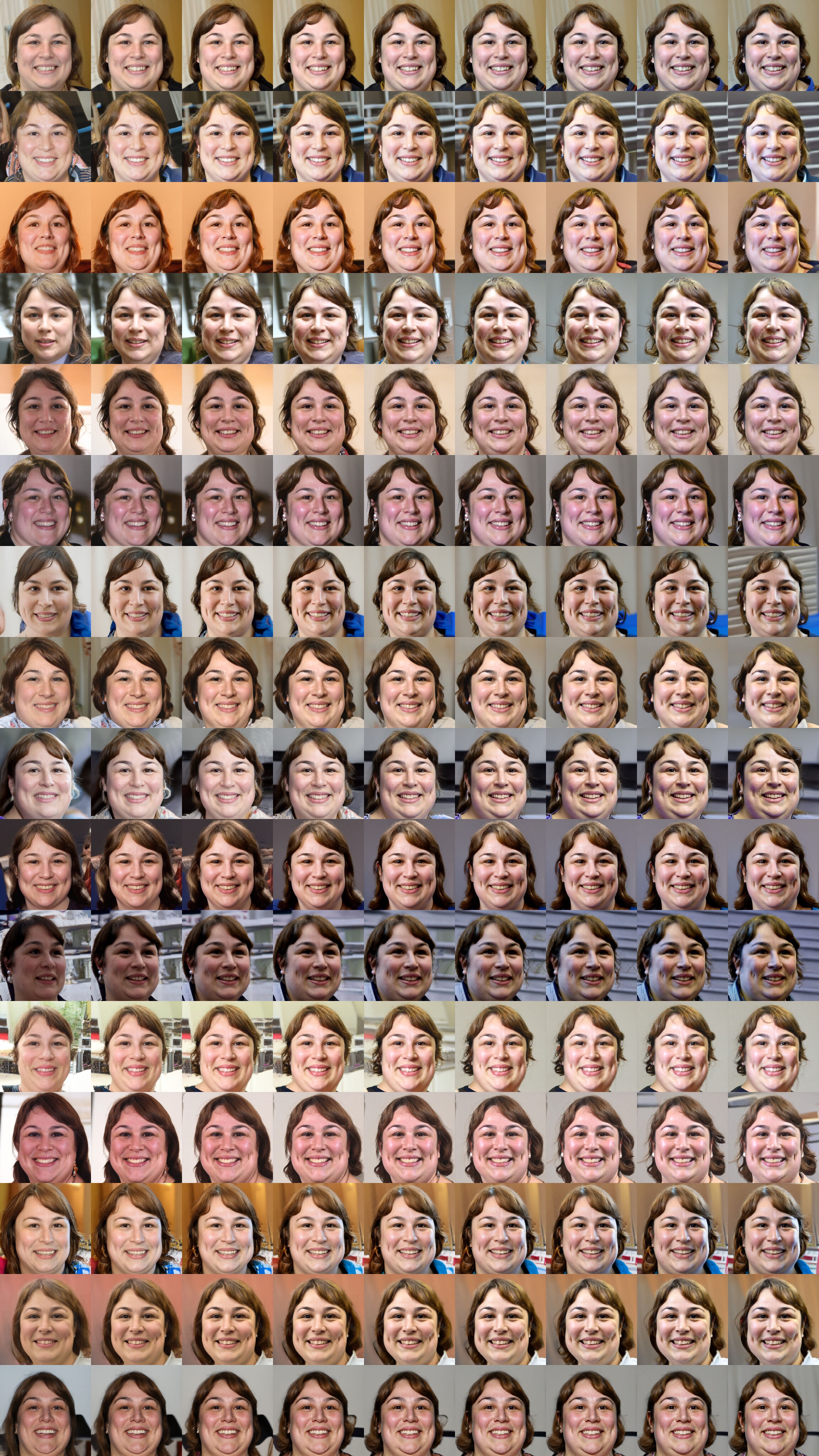} \\
    $s$ & \begin{tabularx}{0.855\textwidth}{ *{9}{Y} } $1.0$ & $1.5$ & $2.0$ & $2.5$ & $3.0$ & $3.5$ & $4.0$ & $4.5$ & $5.0$ \end{tabularx} \\
    \end{tabular}
    }
    \addtolength{\tabcolsep}{5pt}
    \caption{InsightFace~\cite{insightface}}
\end{subfigure}
\caption{Qualitative evaluation of the effect of the guidance scale. For each ID vector extracted from the image on the left, we generate images for four seeds at guidance scales $s$ ranging from $1.0$ to $5.0$.}
\label{fig:qualitative_guidance_full}
\end{figure*}

As described in the main paper, the guidance scale of the classifier-free guidance controls the trade-off between the fidelity in terms of identity preservation (higher guidance) and the diversity of the generated faces (lower guidance). \Cref{fig:qualitative_guidance_full} shows examples of the generated images for different guidance scales $s$ ranging from $1.0$ to $5.0$. To improve the performance for high guidance scales, we adopt dynamic thresholding from Imagen~\cite{imagen} with a threshold of $0.99$.

\clearpage

In the main paper, we evaluate the diversity in terms of the pairwise pose, expression, and LPIPS~\cite{lpips} feature distances among generated images as well as their identity embedding distances. To further quantify the effect of the guidance, we select the first $10000$ images of the FFHQ data set~\cite{stylegan}, extract their ID vectors, and generate one image for each ID vector\footnote{Note that we use the generated images of size $64 \times 64$ (rather than the upsampled images) for computational reasons.}. These $10000$ images are then compared to the corresponding $10000$ original images in terms of their FID scores~\cite{fid} as well as precision and recall~\cite{recall_precision}. The results are shown in \cref{table:fid}. The precision score assesses to which extent the generated samples fall into the distribution of the the real images. Guidance scales in the range $s=1.5$ to $s=2.0$ raise the precision score, implying a higher image quality of the generated images. Even larger guidance scales lead to lower precision scores, which could be explained by the saturated colors observed in \cref{fig:qualitative_guidance_full}. The recall score measures how much of the original distribution is covered by the generated samples and corresponds to the diversity. As the guidance scale increases, recall decreases. Similarly, the FID score gets worse with higher guidance scales, demonstrating the decrease in diversity among the generated images.

\begin{table}[htbp]
\centering
\small{
\begin{tabular}{lcS[table-format=2.3]S[table-format=1.3]S[table-format=1.3]}
    \toprule
    Method ID vector & Guidance scale $s$ & {FID ($\downarrow$)} & {Precision ($\uparrow$)} & {Recall ($\uparrow$)} \\
    \midrule
    \multirow{5}{*}{ID3PM (Ours, FaceNet~\cite{facenet, facenet_pytorch})} & $1.0$ &  \textbf{8.014} & 0.768 & \textbf{0.498} \\
    & $1.5$ &  9.141 & 0.782 & 0.490 \\
    & $2.0$ & 10.434 & \textbf{0.783} & 0.476 \\
    & $2.5$ & 11.659 & 0.775 & 0.452 \\
    & $3.0$ & 12.806 & 0.771 & 0.441 \\
    \midrule
    \multirow{5}{*}{ID3PM (Ours, InsightFace~\cite{insightface})} & $1.0$ &  \textbf{6.786} & 0.774 & \textbf{0.517} \\
    & $1.5$ &  7.442 & \textbf{0.782} & 0.516 \\
    & $2.0$ &  8.497 & 0.771 & 0.508 \\
    & $2.5$ &  9.286 & 0.763 & 0.506 \\
    & $3.0$ & 10.119 & 0.746 & 0.488 \\
    \bottomrule
\end{tabular}
}
\caption{Quantitative evaluation of the effect of the guidance scale on image quality and diversity. The best performing setting for each ID vector is marked in bold.}
\label{table:fid}
\end{table}

Additionally, we perform the face verification experiment from the main paper on LFW~\cite{lfw}, AgeDB-30~\cite{agedb}, and CFP-FP~\cite{cfpfp} with guidance scales between $1.0$ and $3.0$ for our models trained using FaceNet~\cite{facenet, facenet_pytorch} and InsightFace~\cite{insightface} ID vectors. As seen in \cref{table:face_recog_scores_guidance}, the face verification accuracy generally increases with higher guidance scales but saturates eventually, confirming our qualitative findings that the guidance aids in the identity preservation. 

\begin{table*}[htbp]
\centering
    \small{
    \begin{tabular}{lclcclcclcc}
    \toprule
    \multirow{2}{*}{Method} & \multirow{2}{*}{\begin{tabular}[c]{@{}c@{}}Guidance\\ scale $s$ \end{tabular}} &  & \multicolumn{2}{c}{LFW} &  & \multicolumn{2}{c}{AgeDB-30} &  & \multicolumn{2}{c}{CFP-FP} \\
    \cmidrule{4-5} \cmidrule{7-8} \cmidrule{10-11}
    & & & {ArcFace $\uparrow$} & {FaceNet $\uparrow$} & & {ArcFace $\uparrow$} & {FaceNet $\uparrow$} & & {ArcFace $\uparrow$} & {FaceNet $\uparrow$} \\
    \midrule
    Real images & {-} & & 99.83\% & 99.65\% & & 98.23\% & 91.33\% & & 98.86\% & 96.43\% \\
    \midrule
    \multirow{5}{*}{ID3PM (Ours, FaceNet~\cite{facenet, facenet_pytorch})} & 1.0 & & 95.60\% & 98.62\% & & 84.07\% & 86.20\% & & 91.83\% & 94.30\% \\
    & 1.5 & & 97.08\% & 99.00\% & & 87.55\% & 87.90\% & & 94.20\% & 95.04\% \\
    & 2.0 & & 97.65\% & 98.98\% & & 88.22\% & 88.00\% & & 94.47\% & \textbf{95.23\%} \\
    & 2.5 & & 97.92\% & 98.92\% & & \textbf{88.75\%} & \textbf{88.47\%} & & \textbf{94.61\%} & 95.19\% \\
    & 3.0 & & \textbf{98.03\%} & \textbf{99.07\%} & & 88.45\% & \textbf{88.47\%} & & 94.47\% & 95.03\% \\
    \midrule
    \multirow{5}{*}{ID3PM (Ours, InsightFace~\cite{insightface})} & 1.0 & & 98.38\% & 94.37\% & & 91.88\% & 75.60\% & & 93.50\% & 85.26\% \\
    & 1.5 & & 98.95\% & 95.62\% & & 93.88\% & 77.57\% & & 95.59\% & 86.81\% \\
    & 2.0 & & \textbf{99.20\%} & 96.02\% & & 94.53\% & 79.15\% & & 96.13\% & 87.43\% \\
    & 2.5 & & 98.97\% & \textbf{96.37\%} & & \textbf{94.88\%} & 79.20\% & & 96.03\% & 87.83\% \\
    & 3.0 & & 99.15\% & 96.30\% & & 94.78\% & \textbf{79.25\%} & & \textbf{96.16\%} & \textbf{87.97\%} \\
    \bottomrule
    \end{tabular}
    }
\caption{Quantitative evaluation similar to the main paper but with different values for the classifier-free guidance $s$ for our models trained using ID vectors from FaceNet~\cite{facenet, facenet_pytorch} and InsightFace~\cite{insightface}. The best performing setting for each ID vector is marked in bold.}
\label{table:face_recog_scores_guidance}
\end{table*}

\clearpage

\subsection{Identity vector latent space}

\begin{figure*}[htpb]
\centering
\begin{subfigure}{0.95\textwidth}
\centering
    \addtolength{\tabcolsep}{-2pt}
    \small{
    \begin{tabular}{lc}
    NbNet~\cite{nbnet} & \raisebox{-.5\height}{\adjincludegraphics[width=.85\textwidth, trim={0 0 0 0}, clip]{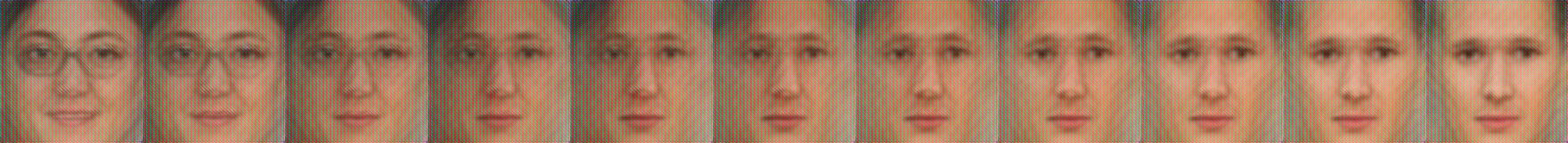}} \\
    \\[-0.35cm]
    {\begin{tabular}[l]{@{}l@{}}Gaussian \\sampling~\cite{gaussian_sampling} \end{tabular}} & \raisebox{-.5\height}{\adjincludegraphics[width=.85\textwidth, trim={0 0 0 0}, clip]{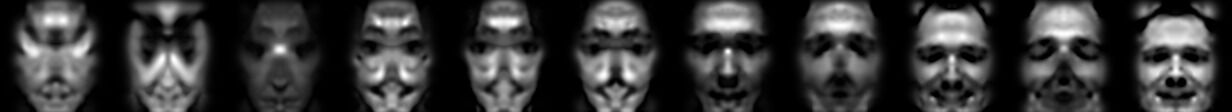}} \\
    \\[-0.35cm]
    {\begin{tabular}[l]{@{}l@{}}StyleGAN \\search~\cite{stylegan-search} \end{tabular}} & \raisebox{-.5\height}{\adjincludegraphics[width=.85\textwidth, trim={0 0 0 0}, clip]{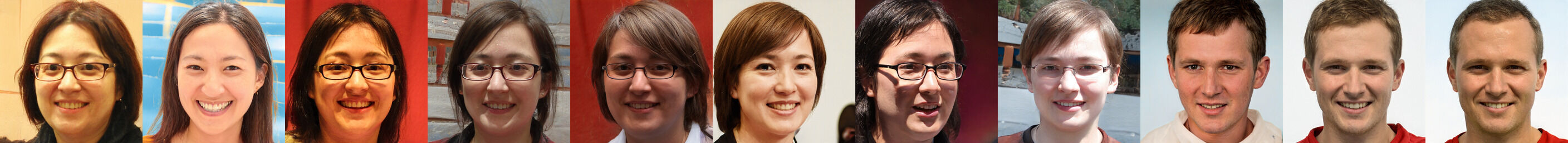}} \\
    \\[-0.35cm]
    {\begin{tabular}[l]{@{}l@{}}ID3PM (Ours, \\ FaceNet~\cite{facenet, facenet_pytorch}) \end{tabular}} & \raisebox{-.5\height}{\adjincludegraphics[width=.85\textwidth, trim={0 {.75\height} 0 0}, clip]{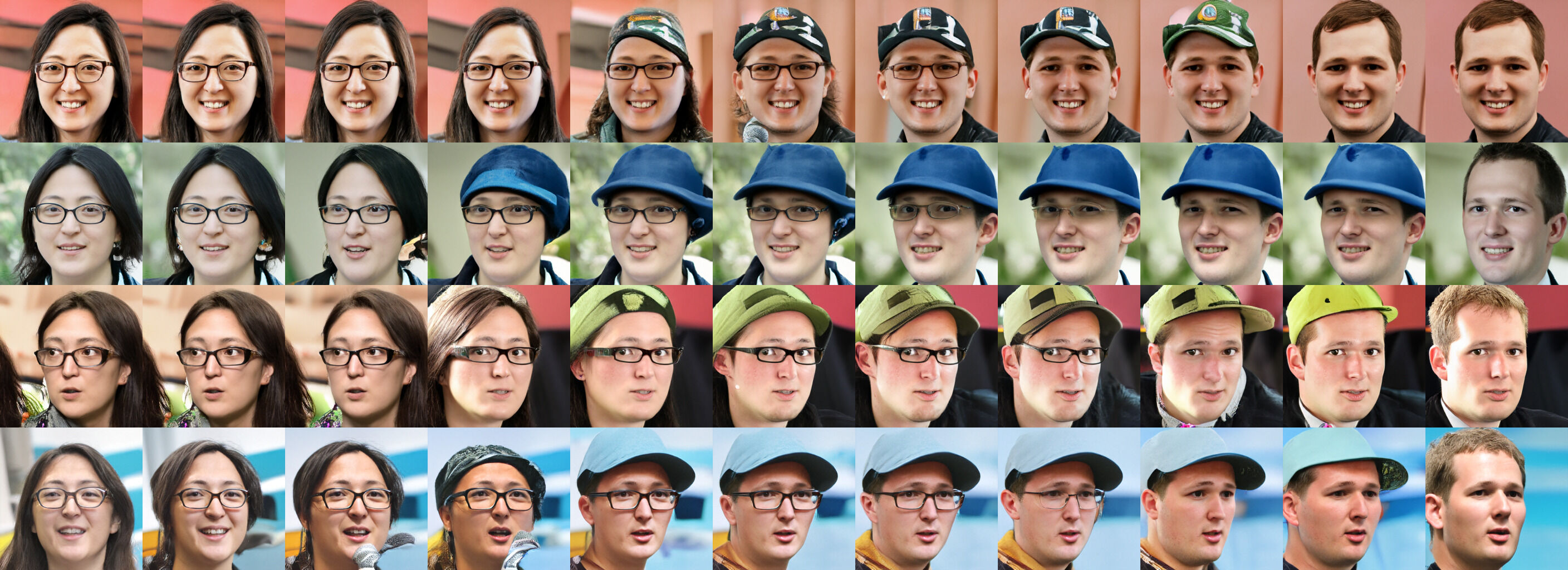}} \\
    \\[-0.35cm]
    {\begin{tabular}[l]{@{}l@{}}ID3PM (Ours, \\ InsightFace~\cite{insightface}) \end{tabular}} & \raisebox{-.5\height}{\adjincludegraphics[width=.85\textwidth, trim={0 {.75\height} 0 0}, clip]{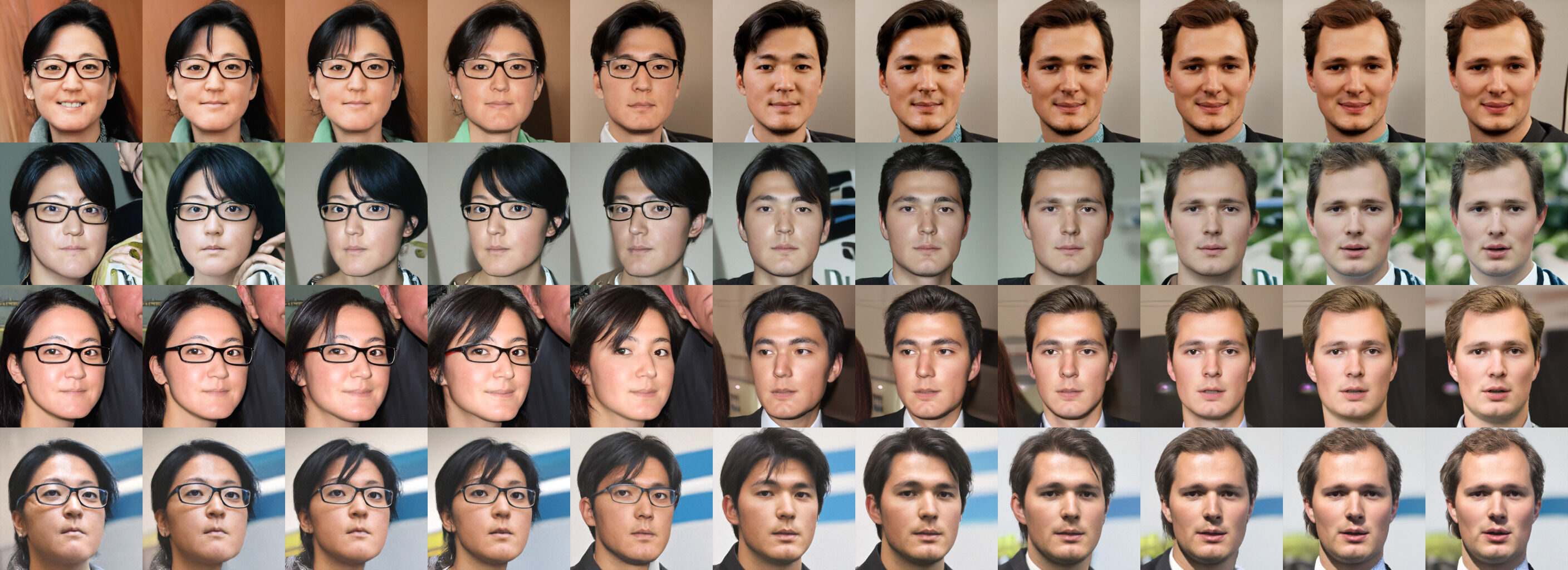}} \\
    \end{tabular}
    }
    \addtolength{\tabcolsep}{2pt}
    \caption{Identity 1 $\longleftrightarrow$ Identity 2}
    \vspace{5mm}
\end{subfigure}
\begin{subfigure}{0.95\textwidth}
\centering
    \addtolength{\tabcolsep}{-2pt}
    \small{
    \begin{tabular}{lc}
    NbNet~\cite{nbnet} & \raisebox{-.5\height}{\adjincludegraphics[width=.85\textwidth, trim={0 0 0 0}, clip]{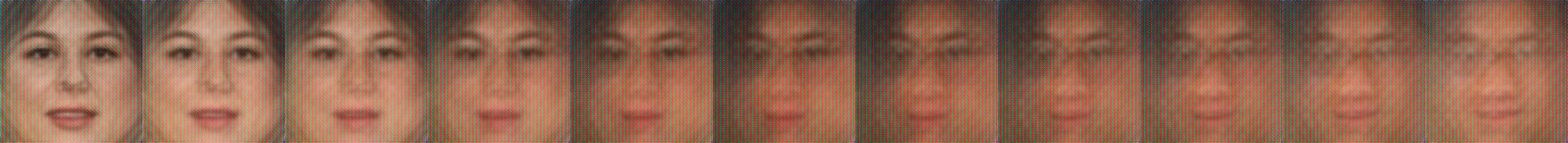}} \\
    \\[-0.35cm]
    {\begin{tabular}[l]{@{}l@{}}Gaussian \\sampling~\cite{gaussian_sampling} \end{tabular}} & \raisebox{-.5\height}{\adjincludegraphics[width=.85\textwidth, trim={0 0 0 0}, clip]{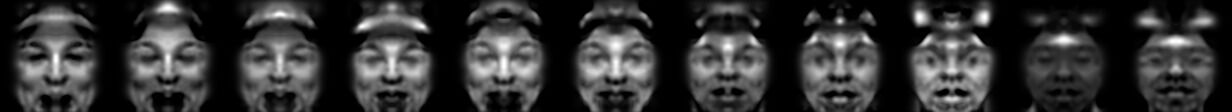}} \\
    \\[-0.35cm]
    {\begin{tabular}[l]{@{}l@{}}StyleGAN \\search~\cite{stylegan-search} \end{tabular}} & \raisebox{-.5\height}{\adjincludegraphics[width=.85\textwidth, trim={0 0 0 0}, clip]{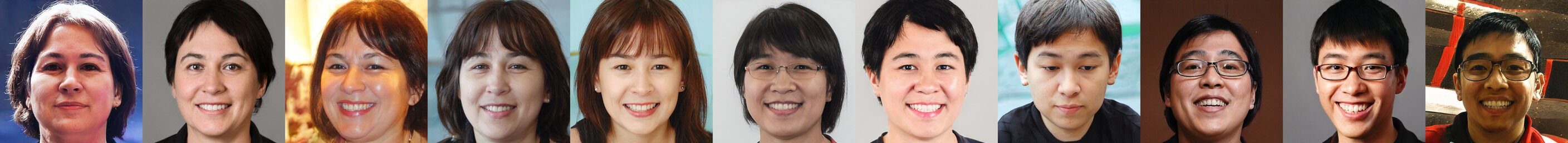}} \\
    \\[-0.35cm]
    {\begin{tabular}[l]{@{}l@{}}ID3PM (Ours, \\ FaceNet~\cite{facenet, facenet_pytorch}) \end{tabular}} & \raisebox{-.5\height}{\adjincludegraphics[width=.85\textwidth, trim={0 {.75\height} 0 0}, clip]{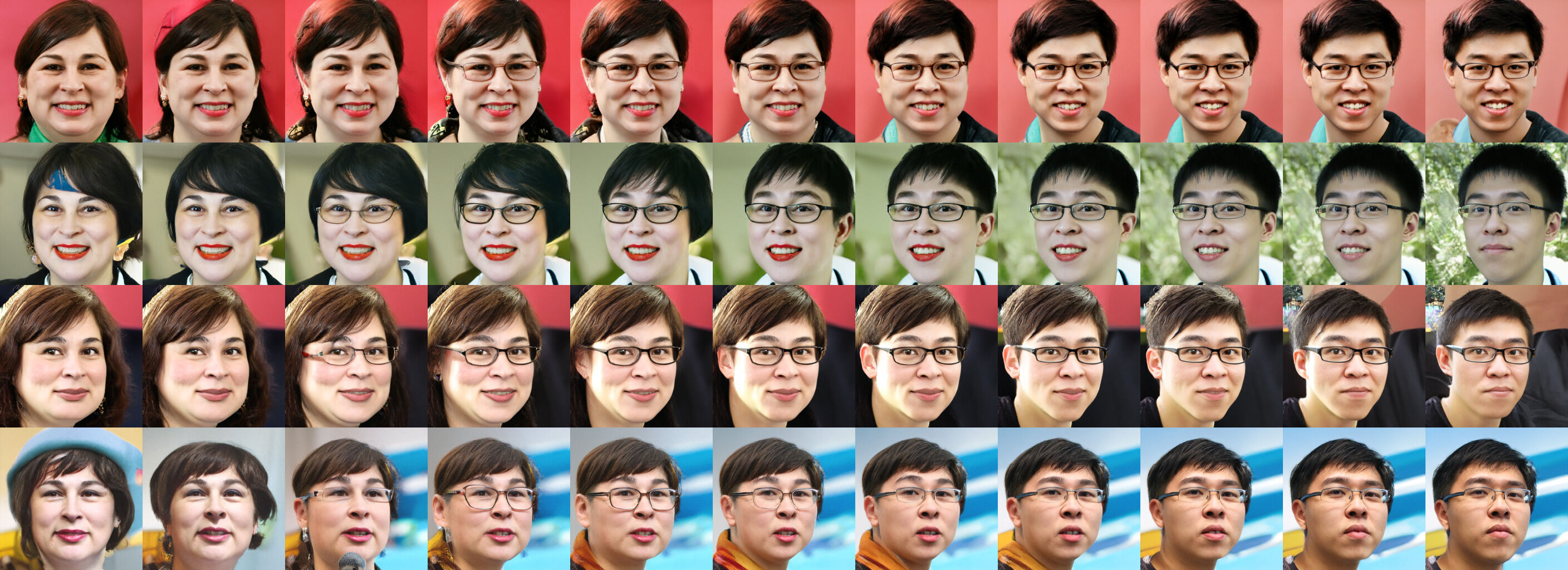}} \\
    \\[-0.35cm]
    {\begin{tabular}[l]{@{}l@{}}ID3PM (Ours, \\ InsightFace~\cite{insightface}) \end{tabular}} & \raisebox{-.5\height}{\adjincludegraphics[width=.85\textwidth, trim={0 {.75\height} 0 0}, clip]{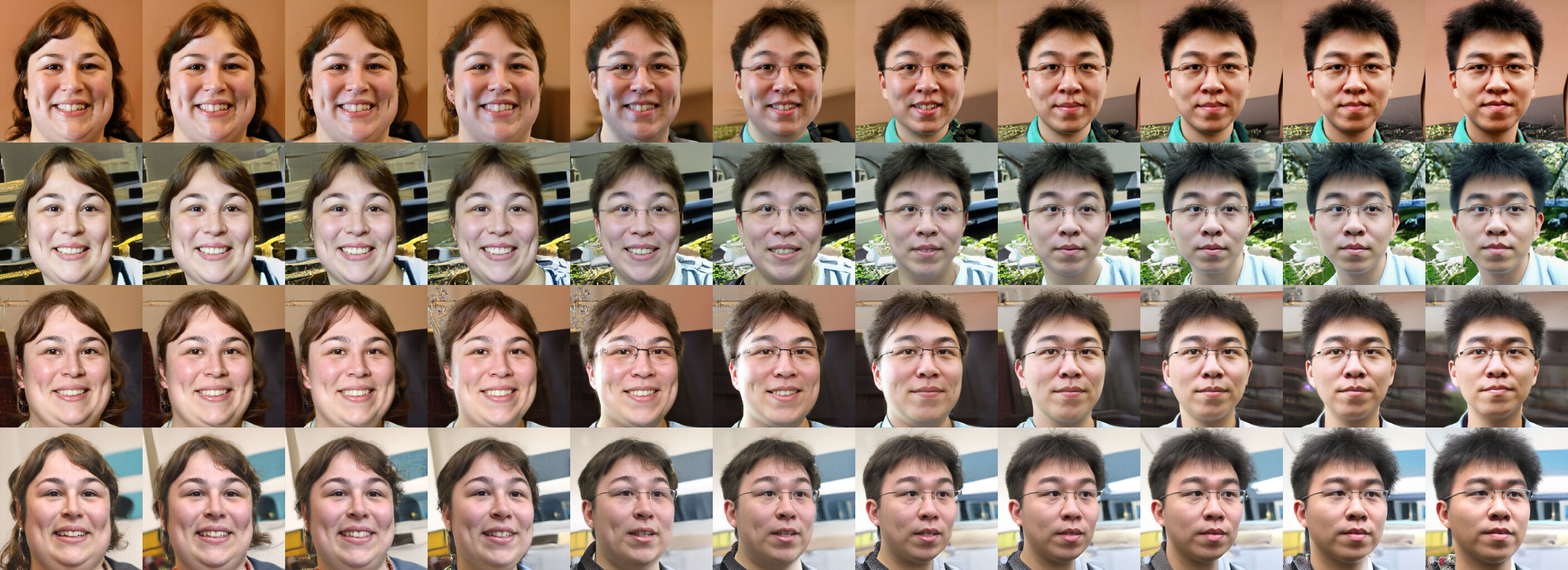}} \\
    \end{tabular}
    }
    \addtolength{\tabcolsep}{2pt}
    \caption{Identity 3 $\longleftrightarrow$ Identity 4}
\end{subfigure}
    \caption{Identity interpolations for two pairs of identities using state-of-the-art methods. Our method is the only method that provides realistic, smooth interpolations.}
    \label{fig:interpolation_competitors}
\end{figure*}

Our method is the first to our knowledge to enable smooth interpolations in the ID vector latent space. While we can condition other methods on an interpolated or adapted ID vector as well, their results lack realism and/or do not transition smoothly between images. This is demonstrated in the identity interpolations in \cref{fig:interpolation_competitors}. Note that spherical linear interpolation was used for all methods, but linear interpolation leads to a similar performance. The other one-to-many approaches, Gaussian sampling~\cite{gaussian_sampling} and StyleGAN search~\cite{stylegan-search}, were extended such that the seed of all random number generators is set before each image is generated to eliminate discontinuities due to the randomness of the generation process. Nevertheless, certain identity-agnostic characteristics, such as the the expression, pose, and background for StyleGAN search~\cite{stylegan-search}, change from one image to the next.

\clearpage

As described in the main paper, we can find custom directions in the ID vector latent space. This allows us to change certain identity-specific features such as the age, glasses, beard, gender, and baldness during image generation by traversing along a given direction as visualized in \cref{{fig:id_vec_custom_dir}}. 

\begin{figure*}[htpb]
    \centering
    \addtolength{\tabcolsep}{-5pt}
    \begin{subfigure}{0.45\textwidth}
        \small{
        \begin{tabular}{lcccc}
            Older & & 
            \raisebox{-.5\height}{\includegraphics[width=0.3\textwidth]{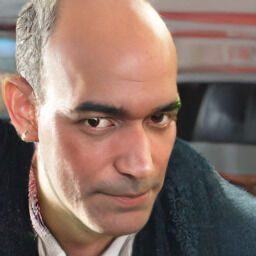}} & \raisebox{-.5\height}{\includegraphics[width=0.3\textwidth]{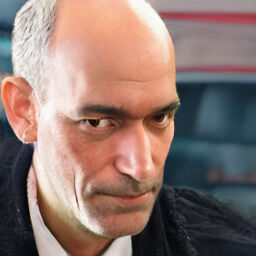}} & \raisebox{-.5\height}{\includegraphics[width=0.3\textwidth]{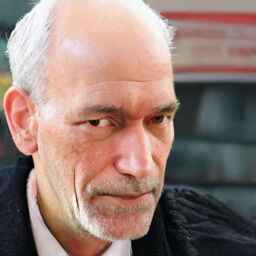}} \\
            \\[-0.32cm]
            Glasses & & 
            \raisebox{-.5\height}{\includegraphics[width=0.3\textwidth]{images/custom_dir/rajesh/0.jpg}} & \raisebox{-.5\height}{\includegraphics[width=0.3\textwidth]{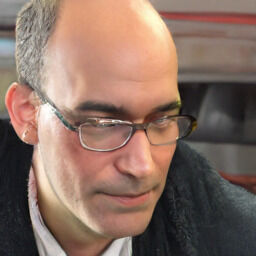}} & \raisebox{-.5\height}{\includegraphics[width=0.3\textwidth]{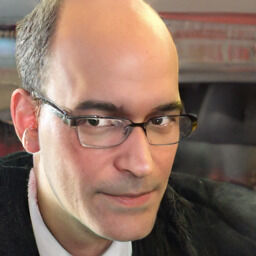}} \\
            \\[-0.32cm]
            Beard & & 
            \raisebox{-.5\height}{\includegraphics[width=0.3\textwidth]{images/custom_dir/rajesh/0.jpg}} & \raisebox{-.5\height}{\includegraphics[width=0.3\textwidth]{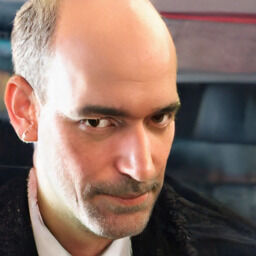}} & \raisebox{-.5\height}{\includegraphics[width=0.3\textwidth]{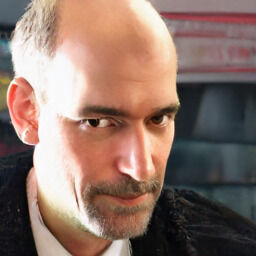}} \\
        \end{tabular}
        }
        \caption{Identity 1}
    \end{subfigure}
    \hspace{1.0cm}
    \begin{subfigure}{0.45\textwidth}
        \small{
        \begin{tabular}{lcccc}
            Younger & & 
            \raisebox{-.5\height}{\includegraphics[width=0.3\textwidth]{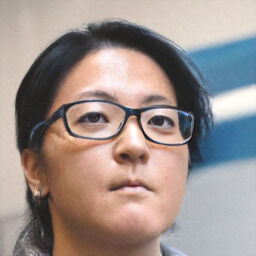}} & \raisebox{-.5\height}{\includegraphics[width=0.3\textwidth]{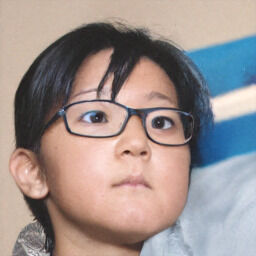}} & \raisebox{-.5\height}{\includegraphics[width=0.3\textwidth]{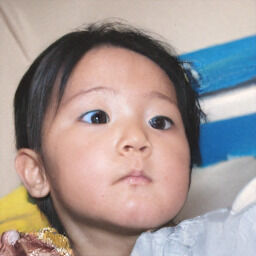}} \\
            \\[-0.32cm]
            Gender & & 
            \raisebox{-.5\height}{\includegraphics[width=0.3\textwidth]{images/custom_dir/sally/0.jpg}} & \raisebox{-.5\height}{\includegraphics[width=0.3\textwidth]{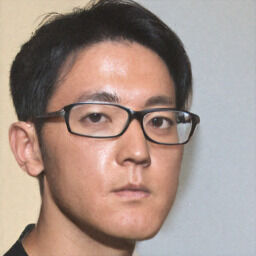}} & \raisebox{-.5\height}{\includegraphics[width=0.3\textwidth]{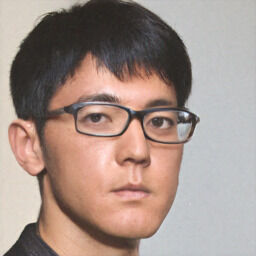}} \\
            \\[-0.32cm]
            Bald & & 
            \raisebox{-.5\height}{\includegraphics[width=0.3\textwidth]{images/custom_dir/sally/0.jpg}} & \raisebox{-.5\height}{\includegraphics[width=0.3\textwidth]{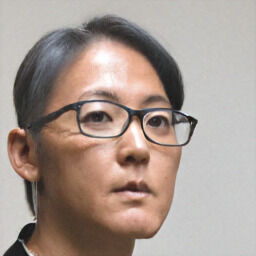}} & \raisebox{-.5\height}{\includegraphics[width=0.3\textwidth]{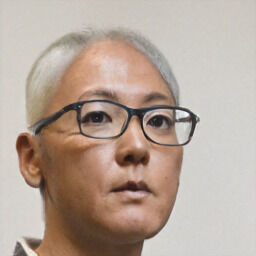}} \\
        \end{tabular}
        }
        \caption{Identity 2}
    \end{subfigure}
    \addtolength{\tabcolsep}{5pt}
    \caption{Controllable image generation through custom directions in the (InsightFace~\cite{insightface}) ID vector latent space.}
    \label{fig:id_vec_custom_dir}
\end{figure*}

\clearpage

\subsection{Attribute conditioning}

To help disentangle identity-specific from identity-agnostic features as well as to obtain additional intuitive control, we propose attribute conditioning in the main paper. We consider three sets of attributes from the FFHQ metadata~\cite{ffhq_metadata}\footnote{We ignore the following attributes from the metadata: smile because it correlates with emotion; and blur, exposure, and noise because we are not interested in them for the purposes of this experiment.} as shown in \cref{table:attributes} by grouping them by how much they contribute to a person's identity. For example, set 1 only contains attributes that are identity-agnostic whereas set 3 also contains attributes that are strongly correlated with identity. In practice, we recommend using set 1 (and thus use that in the main paper) but show sets 2 and 3 for completeness. Note that the attributes from the FFHQ metadata~\cite{ffhq_metadata} are in the form of JSON files. Since the neural networks used to extract the attributes are not publicly available, only black-box approaches such as ours that do not require the neural networks' gradients can be used. The attribute conditioning is thus an example of how our proposed conditioning mechanism can be extended to include information from non-differentiable sources.

\begin{table}[htbp]
\centering
\small{
\begin{tabular}{lcccc}
    \toprule
    \multirow{2}{*}{Attribute} & \multirow{2}{*}{Number of categories} & \multicolumn{3}{c}{Set} \\
    \cmidrule{3-5}
    & & 1 & 2 & 3 \\
    \midrule
    Age & 1 & \xmark & \xmark & \cmark \\
    Emotion & 8 & \cmark & \cmark & \cmark \\
    Facial hair & 3 & \xmark & \xmark & \cmark \\
    Hair & 8 & \xmark & \xmark & \cmark \\
    Head pose & 3 & \cmark & \cmark & \cmark \\
    Gender $^1$ & 1 & \xmark & \xmark & \cmark \\
    Glasses & 1 & \xmark & \cmark & \cmark \\
    Makeup & 2 & \xmark & \cmark & \cmark \\
    Occlusions & 3 & \xmark & \cmark & \cmark \\
    \bottomrule
\end{tabular}
}
\caption{Different attribute sets. The number of (potentially identity-correlated) features increases from left to right. $^1$ While gender arguably falls on a continuous, nonlinear spectrum, we treat it as a binary variable since only this information is available in the data set.}
\label{table:attributes}
\end{table}

\begin{figure*}[htpb]
\centering
\begin{subfigure}{0.9\textwidth}
\centering
    \addtolength{\tabcolsep}{-2pt}
    \small{
    \begin{tabular}{lc}
    ID + Set 1 & \raisebox{-.5\height}{\adjincludegraphics[width=.9\textwidth, trim={0 0 0 0}, clip]{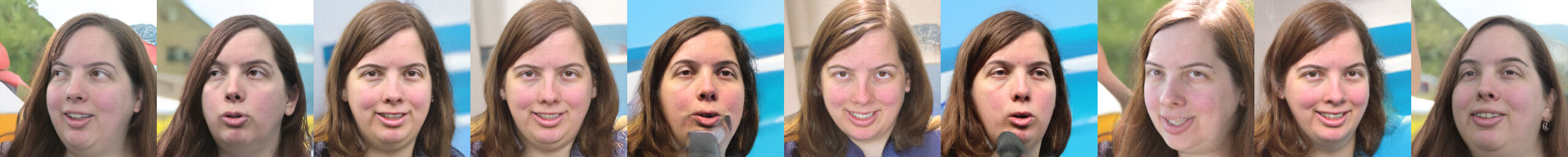}} \\
    \\[-0.35cm]
    ID + Set 2 & \raisebox{-.5\height}{\adjincludegraphics[width=.9\textwidth, trim={0 0 0 0}, clip]{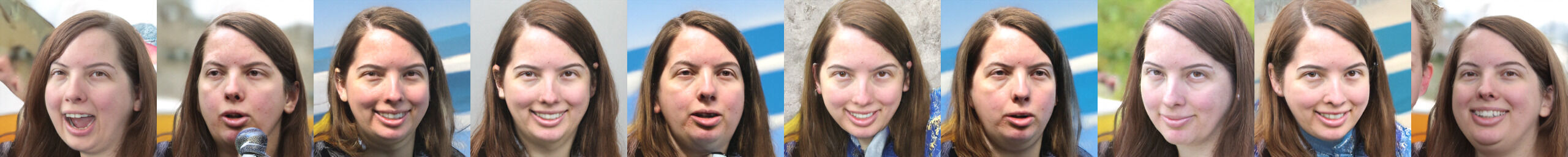}} \\
    \\[-0.35cm]
    ID + Set 3 & \raisebox{-.5\height}{\adjincludegraphics[width=.9\textwidth, trim={0 0 0 0}, clip]{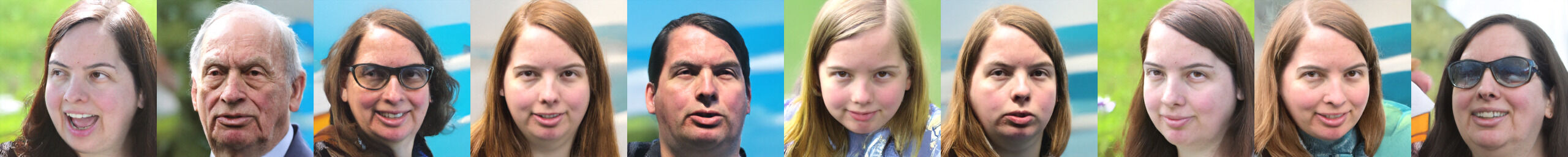}} \\
    Attributes \hspace{1mm} $\rightarrow$ & \begin{tabularx}{0.9\textwidth}{ *{10}{Y} } $69000$ & $69001$ & $69002$ & $69003$ & $69004$ & $69005$ & $69006$ & $69007$ & $69008$ & $69009$ \end{tabularx} \\
    \end{tabular}
    }
    \addtolength{\tabcolsep}{2pt}
    \caption{Identity 1}
    \vspace{5mm}
\end{subfigure}
\begin{subfigure}{0.9\textwidth}
\centering
    \addtolength{\tabcolsep}{-2pt}
    \small{
    \begin{tabular}{lc}
    ID + Set 1 & \raisebox{-.5\height}{\adjincludegraphics[width=.9\textwidth, trim={0 0 0 0}, clip]{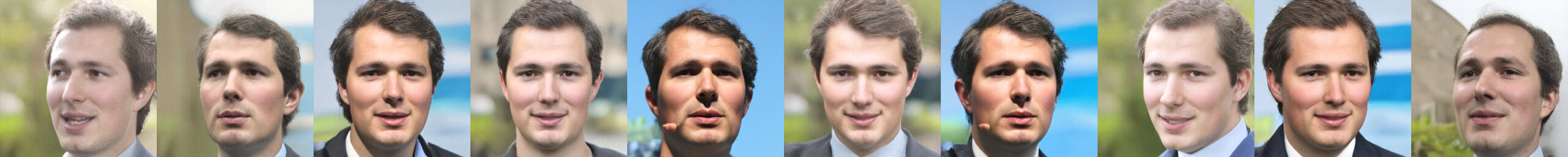}} \\
    \\[-0.35cm]
    ID + Set 2 & \raisebox{-.5\height}{\adjincludegraphics[width=.9\textwidth, trim={0 0 0 0}, clip]{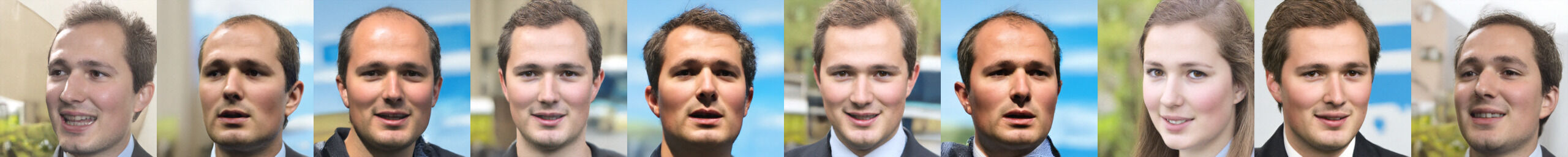}} \\
    \\[-0.35cm]
    ID + Set 3 & \raisebox{-.5\height}{\adjincludegraphics[width=.9\textwidth, trim={0 0 0 0}, clip]{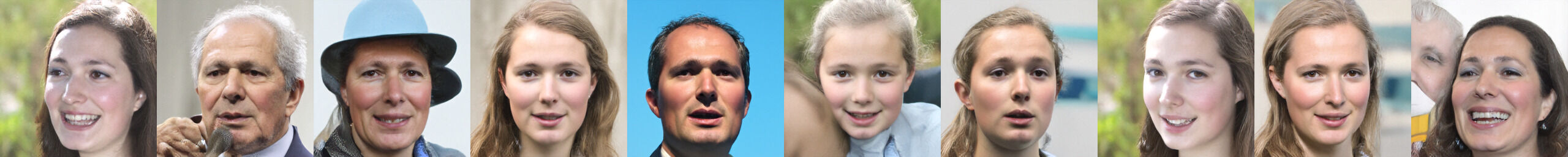}} \\
    Attributes \hspace{1mm} $\rightarrow$ & \begin{tabularx}{0.9\textwidth}{ *{10}{Y} } $69000$ & $69001$ & $69002$ & $69003$ & $69004$ & $69005$ & $69006$ & $69007$ & $69008$ & $69009$ \end{tabularx} \\
    \end{tabular}
    }
    \addtolength{\tabcolsep}{2pt}
    \caption{Identity 2}
    \vspace{5mm}
\end{subfigure}
\begin{subfigure}{0.9\textwidth}
    \centering
    \small{
    \begin{tabular}{llS[table-format=2.0]llS[table-format=2.1]l}
    \toprule
    Image number & Gender & {Age} & Hair & Emotion & {Yaw angle} & Other \\
    \midrule
    69000 & Female & 27 & Brown/black & Happy & -21.9 & Makeup \\
    69001 & Male & 68 & Gray & Neutral & -13.9 & {-} \\
    69002 & Female & 50 & Not visible & Happy & 5.5 & Headwear + glasses \\
    69003 & Female & 20 & Brown/blond & Happy & -0.4 & {-} \\
    69004 & Male & 36 & Black & Neutral & 5.3 & {-} \\
    69005 & Female & 7 & Blond & Happy & -2.6 & {-} \\
    69006 & Female & 18 & Blond & Neutral & 9.0 & {-} \\
    69007 & Female & 21 & Brown & Happy & -24.3 & Makeup \\
    69008 & Female & 28 & Blond & Happy & 10.3 & Makeup \\
    69009 & Female & 43 & Black/brown & Happy & -16.6 & Makeup + glasses \\
    \bottomrule
    \end{tabular}
    }
    \caption{Attribute descriptions}
\end{subfigure}
    \caption{Attribute conditioning for two identities using different attribute sets. Images in each group have the same (InsightFace~\cite{insightface}) ID vector, but the attributes are chosen from images $\{69000, 69001, ..., 69009\}$ of the FFHQ data set.}
    \label{fig:attribute_style}
\end{figure*}

By training our models with attribute conditioning, we can obtain images that recover more of the original data distribution when we sample conditioned on the identity but using the unconditional setting for the attributes (\ie $-1$-vector for the attribute vector), as shown in the main paper. 
However, the attributes can overpower the identity information if the attribute set contains attributes that are heavily correlated with the identity. This is visualized in \cref{fig:attribute_style} where we condition our model trained with InsightFace~\cite{insightface} ID vectors on the same ID vector but different attribute vectors, specifically the ones of the first $10$ images of the FFHQ test set ($\{69000, 69001, ..., 69009\}$). As we cannot show images from the FFHQ~\cite{stylegan} data set as mentioned in the ethics section of the main paper, we instead supply a table with their main attributes. For example, image $69000$ is of a happy 27-year old woman with brown/black hair and makeup whose head is turned slightly to the left. For attribute set 1, only the pose and emotion is copied. For attribute set 2, the makeup is also copied. For attribute 3, the gender is also copied, thus leading to an image of a woman for identity 2. In general, as we add more attributes, the original identity is changed increasingly more. Since attribute sets 2 and 3 can alter the identity significantly, we opt for attribute set 1 in most cases.

Interestingly, even when conditioning on attribute set 1 (only pose and emotion), the average identity distance (seen in the quantitative evaluation of the diversity and identity distances in the main paper) increases despite the visual results appearing similar in terms of identity preservation. We hypothesize that this is because most face recognition vectors (inadvertently) encode the pose and expression (see \cref{sec:analysis}) and are less robust to extreme poses and expressions. Therefore, for the identity distance, it is better to reconstruct a face with a similar pose and expression as the original image. Nevertheless, we argue for the attribute conditioning (using attribute set 1) for most use cases because it leads to more diverse results and allows for an intuitive control over attributes. 

\clearpage

\begin{figure*}[htpb]
\centering
\begin{subfigure}{0.9\textwidth}
\centering
    \addtolength{\tabcolsep}{-4pt}
    \small{
    \begin{tabular}{cc}
    1 & \raisebox{-.5\height}{\adjincludegraphics[width=.95\textwidth, trim={0 0 0 0}, clip]{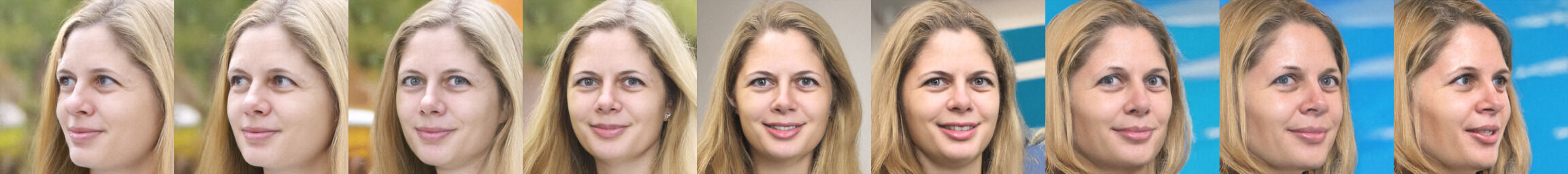}} \\
    \\[-0.3cm]
    2 & \raisebox{-.5\height}{\adjincludegraphics[width=.95\textwidth, trim={0 0 0 0}, clip]{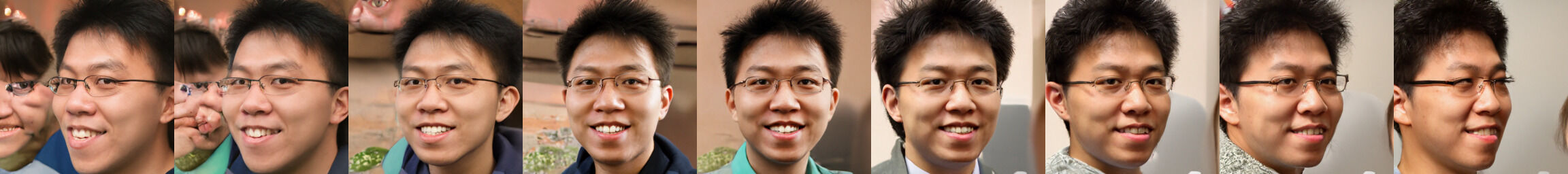}} \\
    \\[-0.3cm]
    3 & \raisebox{-.5\height}{\adjincludegraphics[width=.95\textwidth, trim={0 0 0 0}, clip]{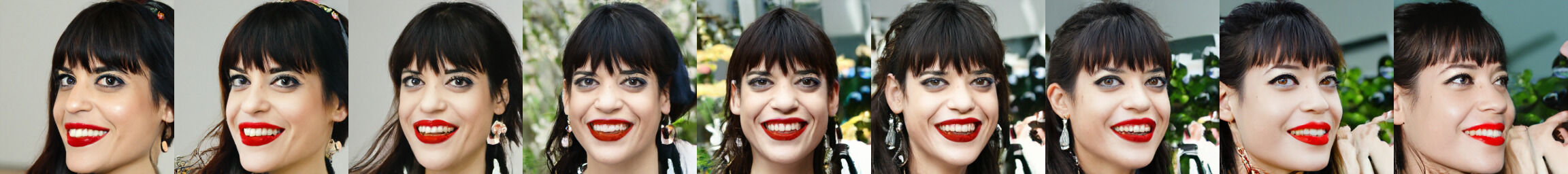}} \\
    \end{tabular}
    }
    \addtolength{\tabcolsep}{4pt}
    \caption{Yaw angle}
    \vspace{5mm}
\end{subfigure}
\begin{subfigure}{0.8\textwidth}
\centering
    \addtolength{\tabcolsep}{-4pt}
    \small{
    \begin{tabular}{cc}
    1 & \raisebox{-.5\height}{\adjincludegraphics[width=.95\textwidth, trim={0 0 0 0}, clip]{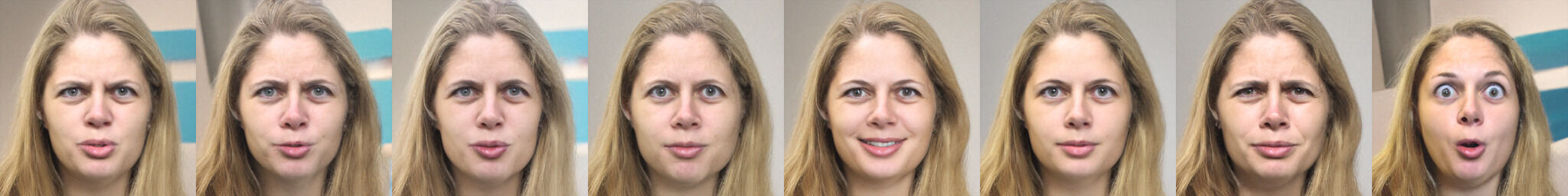}} \\
    \\[-0.3cm]
    2 & \raisebox{-.5\height}{\adjincludegraphics[width=.95\textwidth, trim={0 0 0 0}, clip]{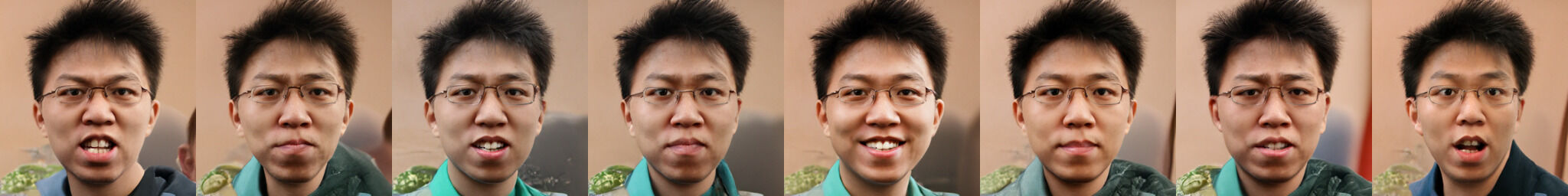}} \\
    \\[-0.3cm]
    3 & \raisebox{-.5\height}{\adjincludegraphics[width=.95\textwidth, trim={0 0 0 0}, clip]{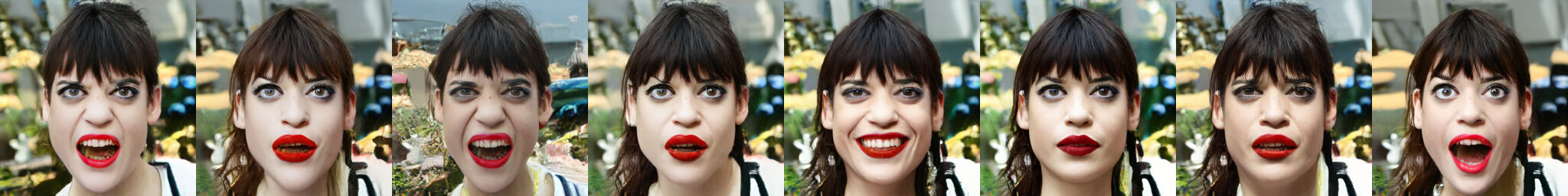}} \\
    & \begin{tabularx}{0.95\textwidth}{ *{8}{Y} } Anger & Contempt & Disgust & Fear & Happiness & Neutral & Sadness & Surprise \end{tabularx} \\
    \end{tabular}
    }
    \addtolength{\tabcolsep}{4pt}
    \caption{Eight emotions}
    \vspace{5mm}
\end{subfigure}
\begin{subfigure}{0.9\textwidth}
\centering
    \addtolength{\tabcolsep}{-4pt}
    \small{
    \begin{tabular}{cc}
    1 & \raisebox{-.5\height}{\adjincludegraphics[width=.95\textwidth, trim={0 0 0 0}, clip]{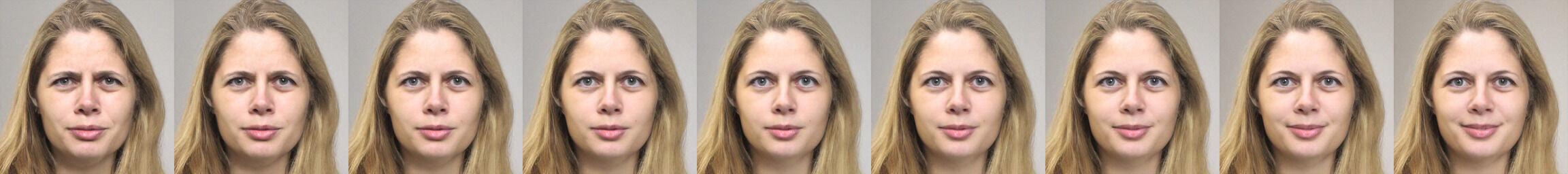}} \\
    \\[-0.3cm]
    2 & \raisebox{-.5\height}{\adjincludegraphics[width=.95\textwidth, trim={0 0 0 0}, clip]{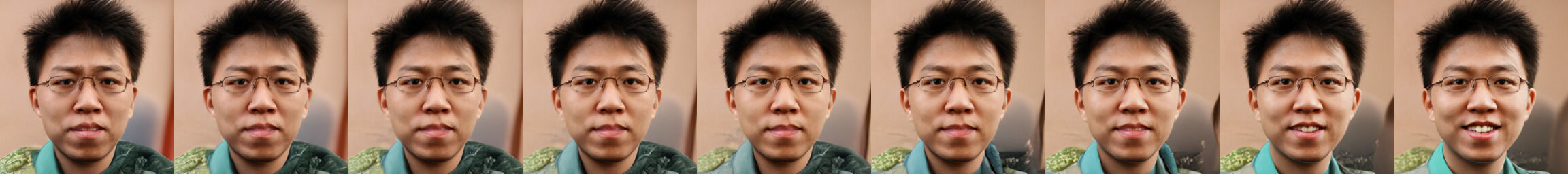}} \\
    \\[-0.3cm]
    3 & \raisebox{-.5\height}{\adjincludegraphics[width=.95\textwidth, trim={0 0 0 0}, clip]{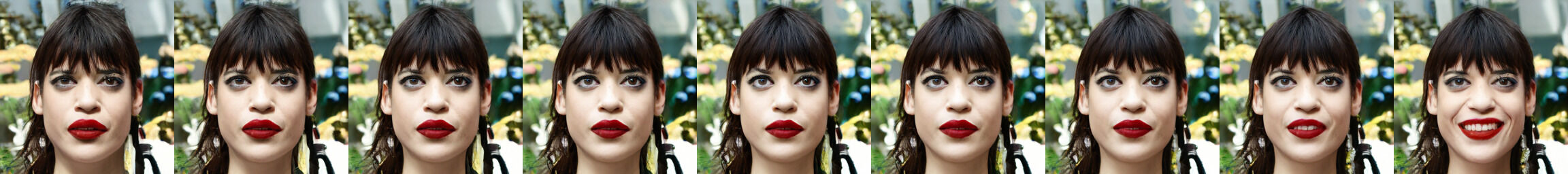}} \\
    \end{tabular}
    }
    \addtolength{\tabcolsep}{4pt}
    \caption{Sad $\longleftrightarrow$ Happy}
\end{subfigure}
    \caption{Controllable image generation through attribute conditioning. We smoothly change three different attributes of three identities. All models were trained using InsightFace~\cite{insightface} ID vectors and attribute set 1.}
    \label{fig:attribute_cond_control_2}
\end{figure*}

Through attribute conditioning, we can simply set the values of desired attributes, such as the emotion and head pose, during inference time to control them, as seen in \cref{fig:attribute_cond_control_2}. Note that our method has no internal structure to enforce 3D consistency. The attribute conditioning alone suffices in generating images that preserve the identity and 3D geometry surprisingly well as we traverse the attribute latent space. This intuitive attribute control paves the way towards using our method to create and augment data sets.

\section{Failure case}

\begin{figure*}[htpb]
    \centering
    \addtolength{\tabcolsep}{-2pt}
    \small{
    \begin{tabular}{cc}
        \raisebox{1.5\height}{\includegraphics[width=0.125\textwidth]{images/original/Prashanth.jpg}} & \includegraphics[width=0.5\textwidth]{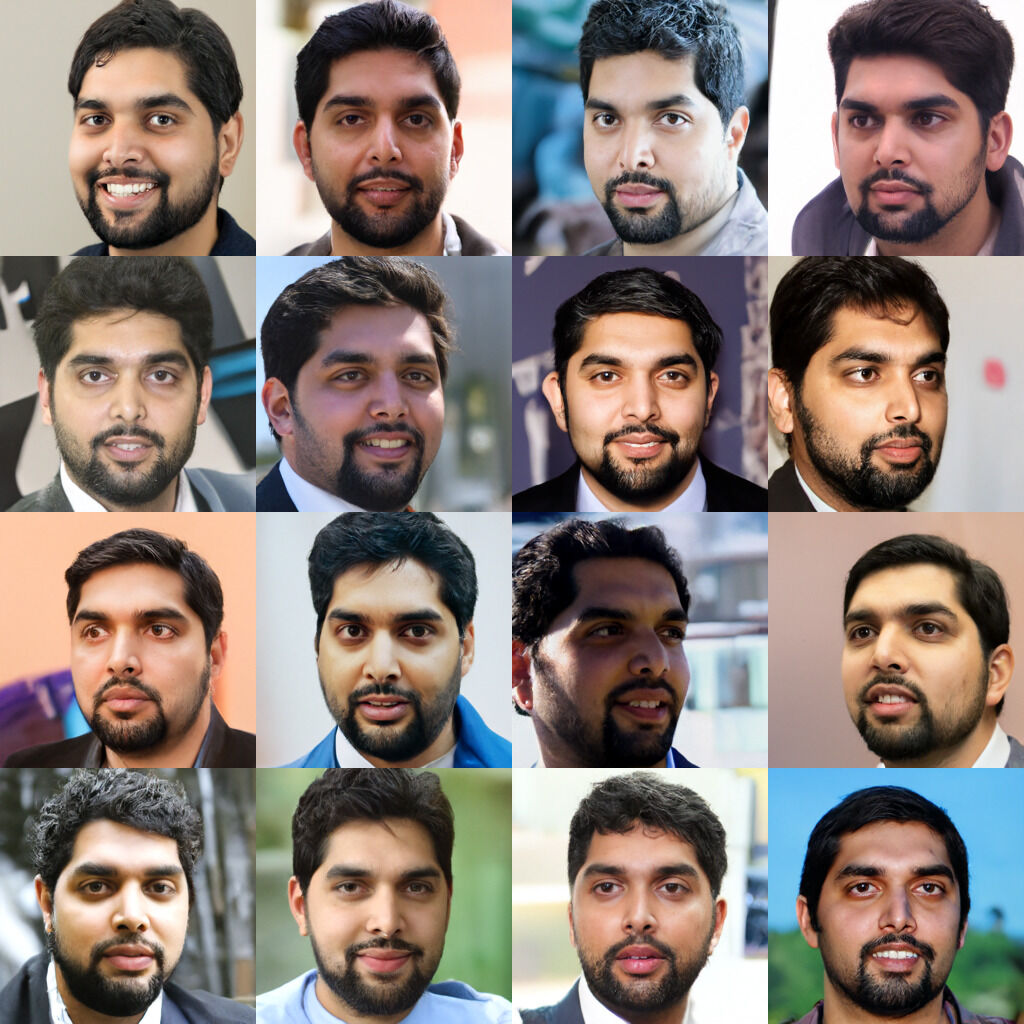} \\
        Original image & ID3PM (Ours, FaceNet~\cite{facenet, facenet_pytorch}) \\
        \\
    \end{tabular}
    }
    \addtolength{\tabcolsep}{2pt}
    \caption{Failure case. Some underrepresented groups in the data sets might have lower identity fidelity for some ID vectors (here FaceNet~\cite{facenet, facenet_pytorch}).}
    \label{fig:failure_case}
\end{figure*}

As described in the main paper, one limitation of our approach is that it inherits the biases of both the face recognition model and the data set used to train the diffusion model. This causes our method to sometimes lose identity fidelity of underrepresented groups in the data set, as seen in the example in \Cref{fig:failure_case}, where the method produces images quite different from the original identity.

\clearpage

\section{Application: Analysis of face recognition methods} \label{sec:analysis}

With the rise of deep learning and large data sets with millions of images, face recognition methods have reached or even surpassed human-level performance~\cite{deepface, deepid, facenet, vggface}. Nevertheless, face recognition systems have known issues in their robustness to different degradations and attacks~\cite{facerec_robustness1, facerec_robustness2} and their biases (\eg in terms of ethnic origin)~\cite{facerec_bias1, facerec_bias2, facerec_bias3}.

Due to its general nature with very low requirements for the data set and few assumptions compared to other methods, our method is very well suited for analyzing and visualizing the latent spaces of different face recognition (FR) models. Since our method does not require access to the internals of the pre-trained face recognition model (\emph{black-box} setting), we can analyze different face recognition methods by simply replacing the input ID vectors without worrying about different deep learning frameworks and the memory burden of adding more models. 

For the analysis in this section, we train multiple versions of the model with ID vectors extracted with the pre-trained face recognition models listed in \cref{table:id_vectors}. Note that since we specifically want to analyze what identity-agnostic features are contained in common face recognition models, we do not use attribute conditioning here since it would disentangle identity-specific and identity-agnostic features.

\subsection{Qualitative evaluation}

\begin{figure*}[htpb]
    \centering
    \addtolength{\tabcolsep}{-5pt}
    \footnotesize{
    \begin{tabular}{cccccc}
        \includegraphics[width=0.09\textwidth]{images/original/Aziz.jpg} & 
        \includegraphics[width=0.09\textwidth]{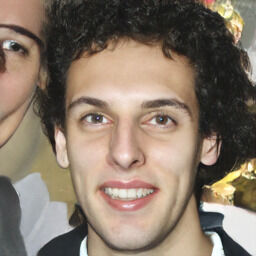} & 
        \includegraphics[width=0.09\textwidth]{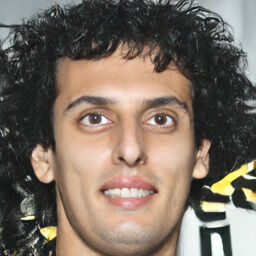} & 
        \includegraphics[width=0.09\textwidth]{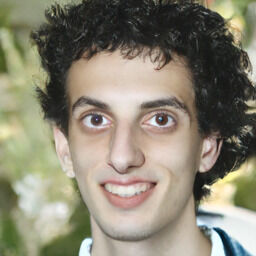} & 
        \includegraphics[width=0.09\textwidth]{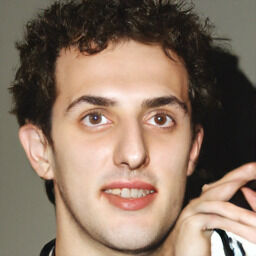} & 
        \includegraphics[width=0.09\textwidth]{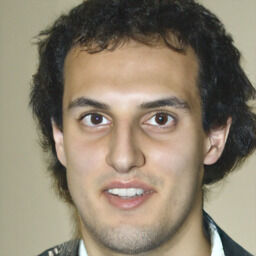} \\
        \\[-0.46cm]
        \includegraphics[width=0.09\textwidth]{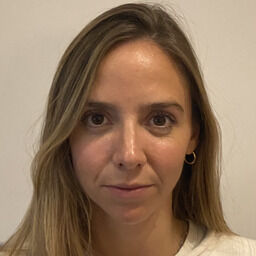} & 
        \includegraphics[width=0.09\textwidth]{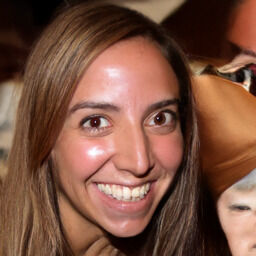} & 
        \includegraphics[width=0.09\textwidth]{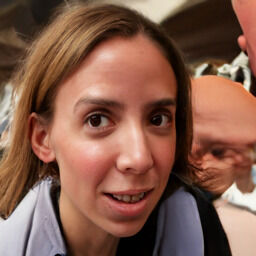} & 
        \includegraphics[width=0.09\textwidth]{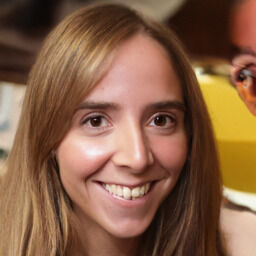} & 
        \includegraphics[width=0.09\textwidth]{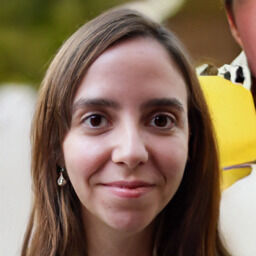} & 
        \includegraphics[width=0.09\textwidth]{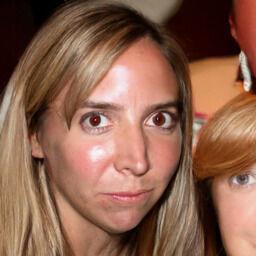} \\
        \\[-0.46cm]
        \includegraphics[width=0.09\textwidth]{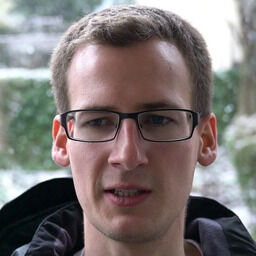} & 
        \includegraphics[width=0.09\textwidth]{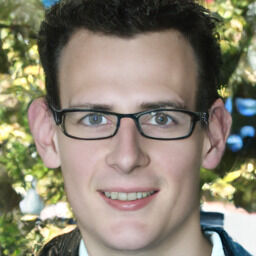} & 
        \includegraphics[width=0.09\textwidth]{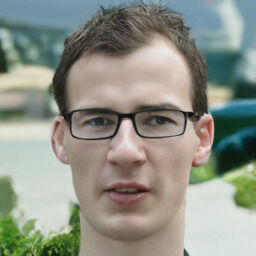} & 
        \includegraphics[width=0.09\textwidth]{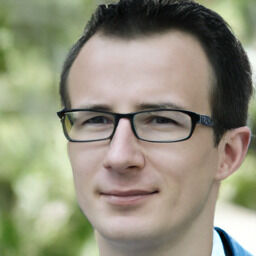} & 
        \includegraphics[width=0.09\textwidth]{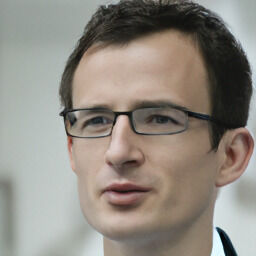} & 
        \includegraphics[width=0.09\textwidth]{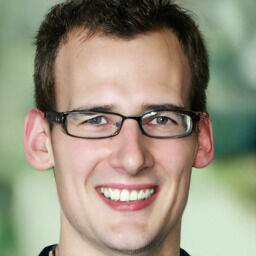} \\
        \\[-0.46cm]
        \includegraphics[width=0.09\textwidth]{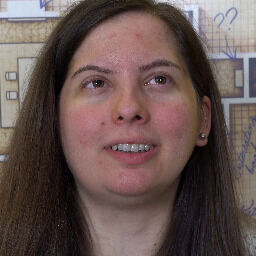} & 
        \includegraphics[width=0.09\textwidth]{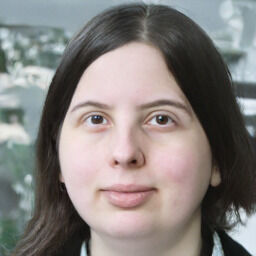} & 
        \includegraphics[width=0.09\textwidth]{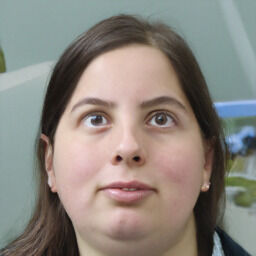} & 
        \includegraphics[width=0.09\textwidth]{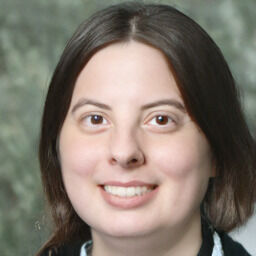} & 
        \includegraphics[width=0.09\textwidth]{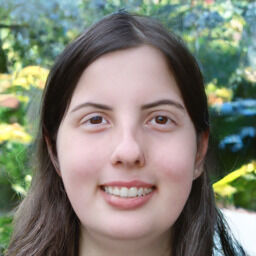} & 
        \includegraphics[width=0.09\textwidth]{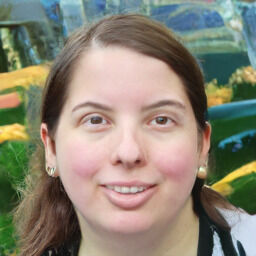} \\
        \\[-0.46cm]
        \includegraphics[width=0.09\textwidth]{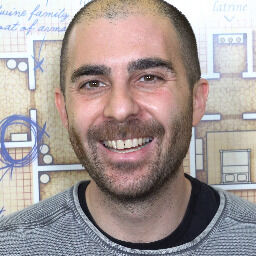} & 
        \includegraphics[width=0.09\textwidth]{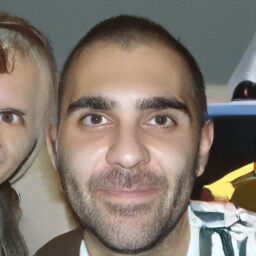} & 
        \includegraphics[width=0.09\textwidth]{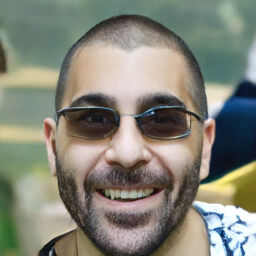} & 
        \includegraphics[width=0.09\textwidth]{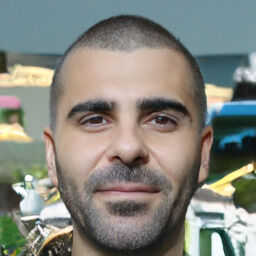} & 
        \includegraphics[width=0.09\textwidth]{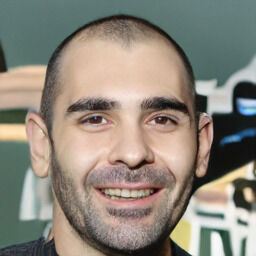} & 
        \includegraphics[width=0.09\textwidth]{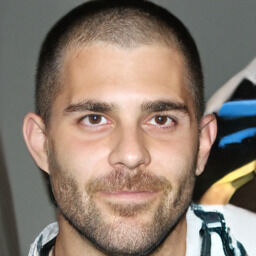} \\
        \\[-0.46cm]
        \includegraphics[width=0.09\textwidth]{images/original/Jakob.jpg} & 
        \includegraphics[width=0.09\textwidth]{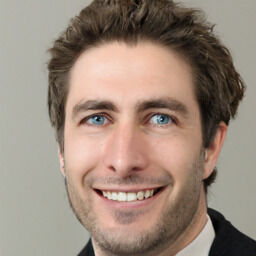} & 
        \includegraphics[width=0.09\textwidth]{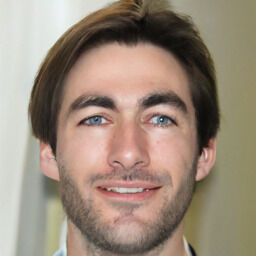} & 
        \includegraphics[width=0.09\textwidth]{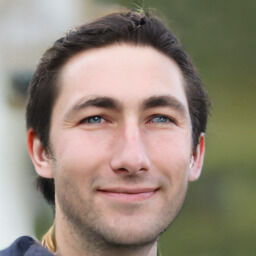} & 
        \includegraphics[width=0.09\textwidth]{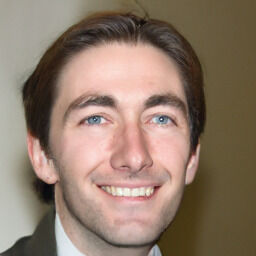} & 
        \includegraphics[width=0.09\textwidth]{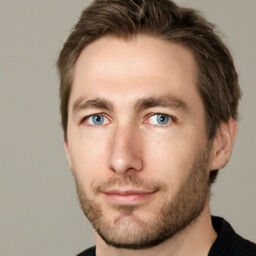} \\
        \\[-0.46cm]
        \includegraphics[width=0.09\textwidth]{images/original/Julia.jpg} & 
        \includegraphics[width=0.09\textwidth]{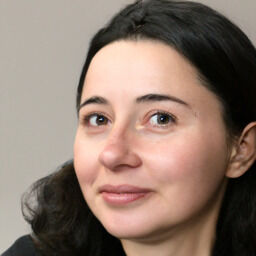} & 
        \includegraphics[width=0.09\textwidth]{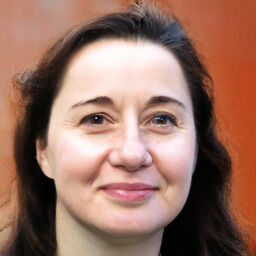} & 
        \includegraphics[width=0.09\textwidth]{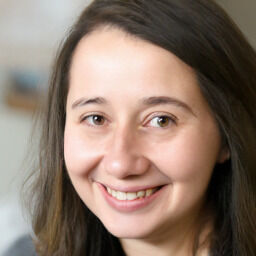} & 
        \includegraphics[width=0.09\textwidth]{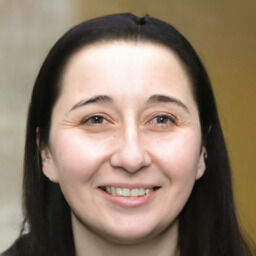} & 
        \includegraphics[width=0.09\textwidth]{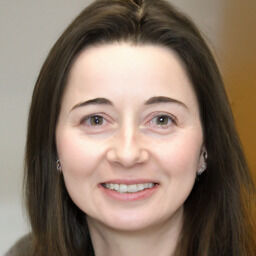} \\
        \\[-0.46cm]
        \includegraphics[width=0.09\textwidth]{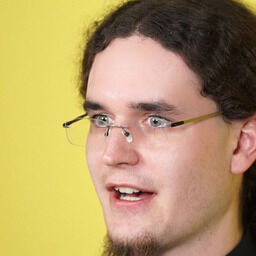} & 
        \includegraphics[width=0.09\textwidth]{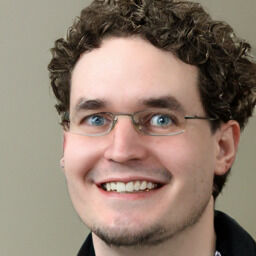} & 
        \includegraphics[width=0.09\textwidth]{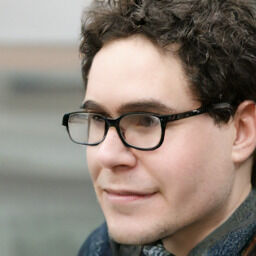} & 
        \includegraphics[width=0.09\textwidth]{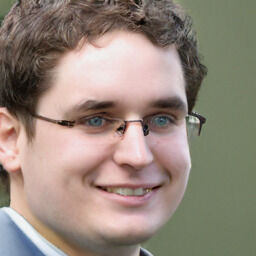} & 
        \includegraphics[width=0.09\textwidth]{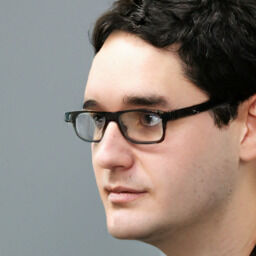} & 
        \includegraphics[width=0.09\textwidth]{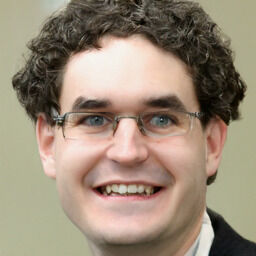} \\
        \\[-0.46cm]
        \includegraphics[width=0.09\textwidth]{images/original/Martina.jpg} & 
        \includegraphics[width=0.09\textwidth]{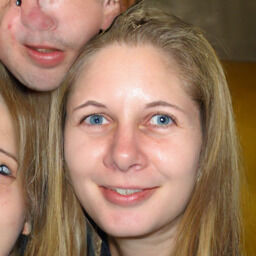} & 
        \includegraphics[width=0.09\textwidth]{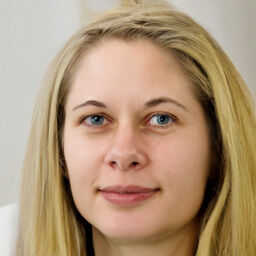} & 
        \includegraphics[width=0.09\textwidth]{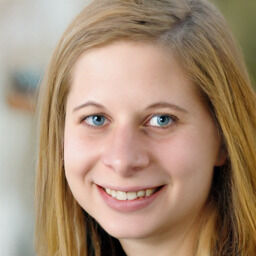} & 
        \includegraphics[width=0.09\textwidth]{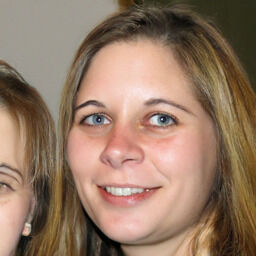} & 
        \includegraphics[width=0.09\textwidth]{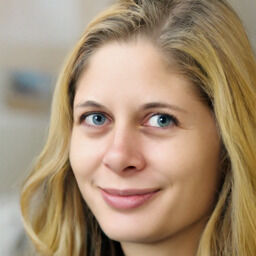} \\
        \\[-0.46cm]
        \includegraphics[width=0.09\textwidth]{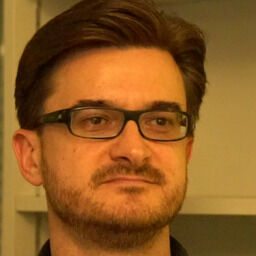} & 
        \includegraphics[width=0.09\textwidth]{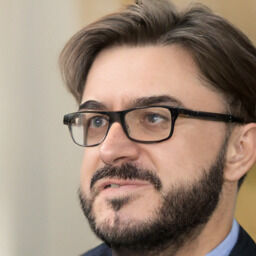} & 
        \includegraphics[width=0.09\textwidth]{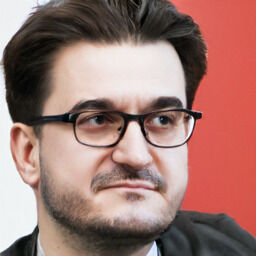} & 
        \includegraphics[width=0.09\textwidth]{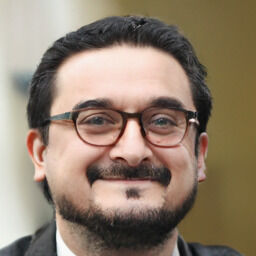} & 
        \includegraphics[width=0.09\textwidth]{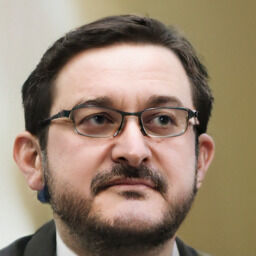} & 
        \includegraphics[width=0.09\textwidth]{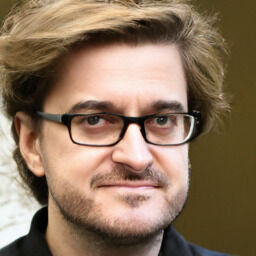} \\
        \\[-0.46cm]
        \includegraphics[width=0.09\textwidth]{images/original/Prashanth.jpg} & 
        \includegraphics[width=0.09\textwidth]{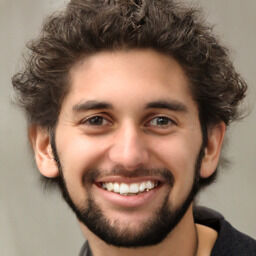} & 
        \includegraphics[width=0.09\textwidth]{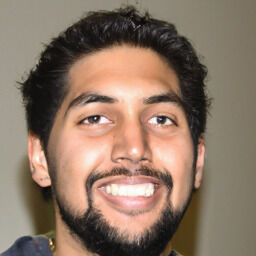} & 
        \includegraphics[width=0.09\textwidth]{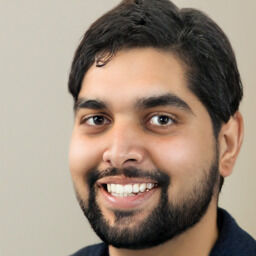} & 
        \includegraphics[width=0.09\textwidth]{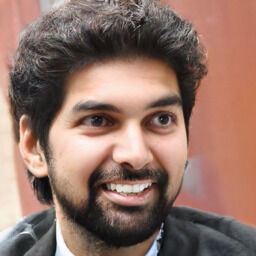} & 
        \includegraphics[width=0.09\textwidth]{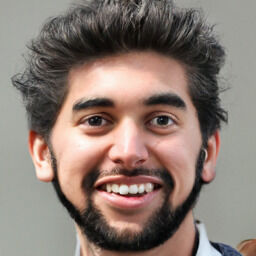} \\
        \\[-0.46cm]
        \includegraphics[width=0.09\textwidth]{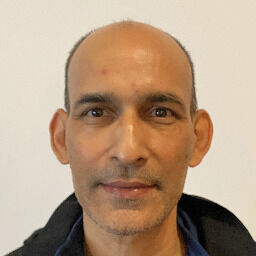} & 
        \includegraphics[width=0.09\textwidth]{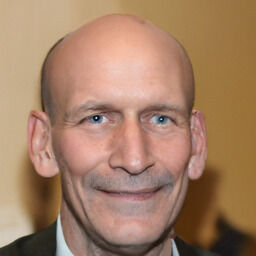} & 
        \includegraphics[width=0.09\textwidth]{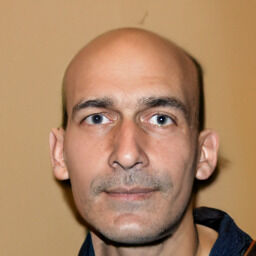} & 
        \includegraphics[width=0.09\textwidth]{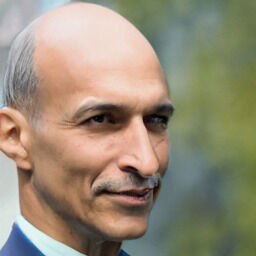} & 
        \includegraphics[width=0.09\textwidth]{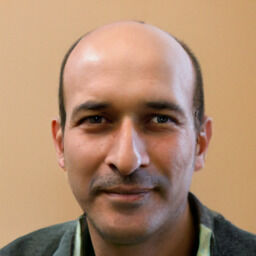} & 
        \includegraphics[width=0.09\textwidth]{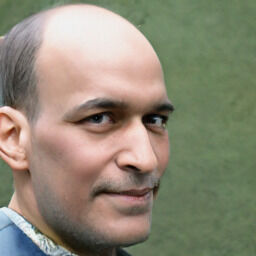} \\
        \\[-0.46cm]
        \includegraphics[width=0.09\textwidth]{images/original/Xianyao.jpg} & 
        \includegraphics[width=0.09\textwidth]{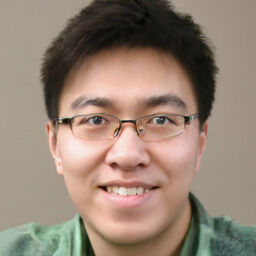} & 
        \includegraphics[width=0.09\textwidth]{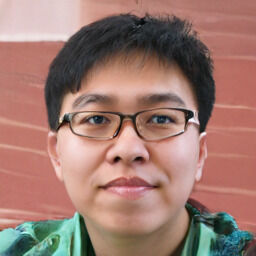} & 
        \includegraphics[width=0.09\textwidth]{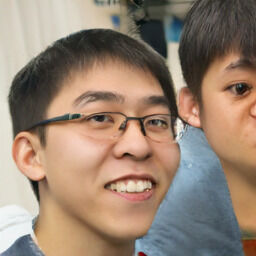} & 
        \includegraphics[width=0.09\textwidth]{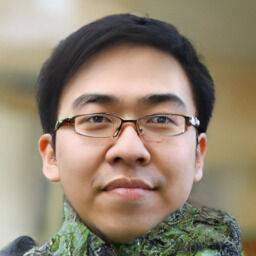} & 
        \includegraphics[width=0.09\textwidth]{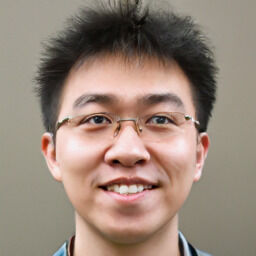} \\
        
        \multirow{2}{*}{\begin{tabular}[c]{@{}c@{}}Original \\ image \end{tabular}} & \multirow{2}{*}{\begin{tabular}[c]{@{}c@{}}Ada- \\ Face~\cite{adaface} \end{tabular}} & \multirow{2}{*}{\begin{tabular}[c]{@{}c@{}}Arc- \\ Face~\cite{arcface, gaussian_sampling} \end{tabular}} & \multirow{2}{*}{\begin{tabular}[c]{@{}c@{}}Face- \\ Net~\cite{facenet, facenet_pytorch} \end{tabular}} &  \multirow{2}{*}{FROM~\cite{from}} & \multirow{2}{*}{\begin{tabular}[c]{@{}c@{}}Insight- \\ Face~\cite{insightface} \end{tabular}} \\
        \\
    \end{tabular}
    }
    \addtolength{\tabcolsep}{5pt}
    \caption{Qualitative evaluation of ID vectors from different state-of-the-art face recognition models. Note that the same seed was used for all images to obtain the most fair results.}
    \label{fig:comp_diff_id_vectors}
\end{figure*}

\Cref{fig:comp_diff_id_vectors} shows uncurated samples of generated images of the considered ID vectors for several identities. While all generated images appear of a similar quality in terms of realism, the identity preservation of different methods behaves quite differently. The relative performance of different ID vectors changes depending on the identity, but the results for FaceNet~\cite{facenet, facenet_pytorch} and InsightFace~\cite{insightface} seem most consistent on average~\footnote{Note that more images than the representative ones shown here were considered to make this statement and other statements in this section.}. As the inversion networks are trained with the same diffusion model architecture and the same data set, we hypothesize that these differences largely boil down to the biases of the respective face recognition methods and the data sets used to train them. 

\subsection{Robustness}

\begin{figure*}[htpb]
\centering
\begin{subfigure}{1\textwidth}
\centering
    \addtolength{\tabcolsep}{-2pt}
    \scriptsize{
    \begin{tabular}{lcc}
    & \raisebox{-.5\height}{\adjincludegraphics[width=.8\textwidth, trim={0 {.835\height} 0 0}, clip]{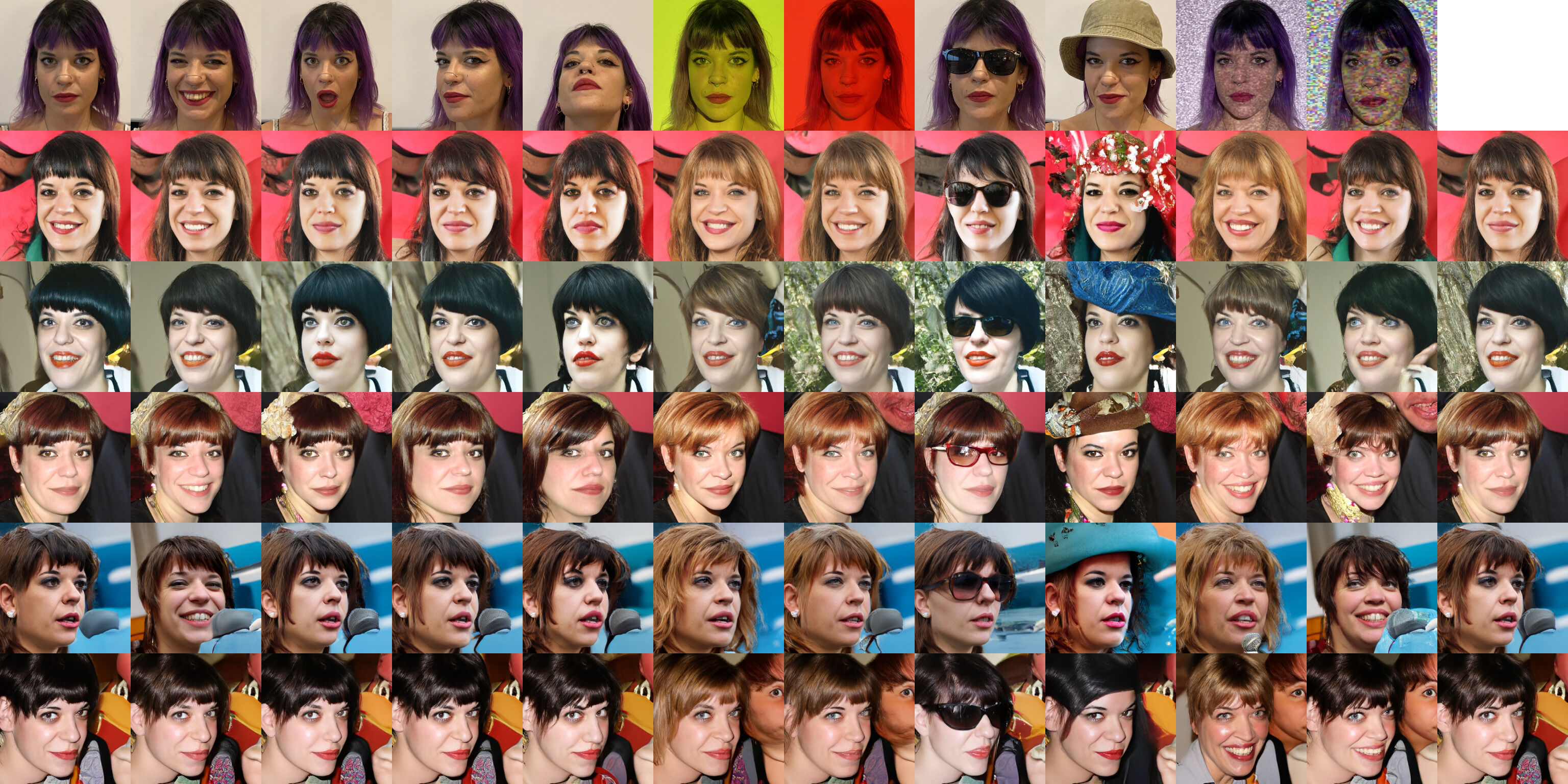}} & \begin{tabular}[c]{@{}c@{}}LFW \\ threshold \end{tabular} \\
    \\[-0.2cm]
    AdaFace~\cite{adaface} & \raisebox{-.5\height}{\adjincludegraphics[width=.8\textwidth, trim={0 {.667\height} 0 {0.167\height}}, clip]{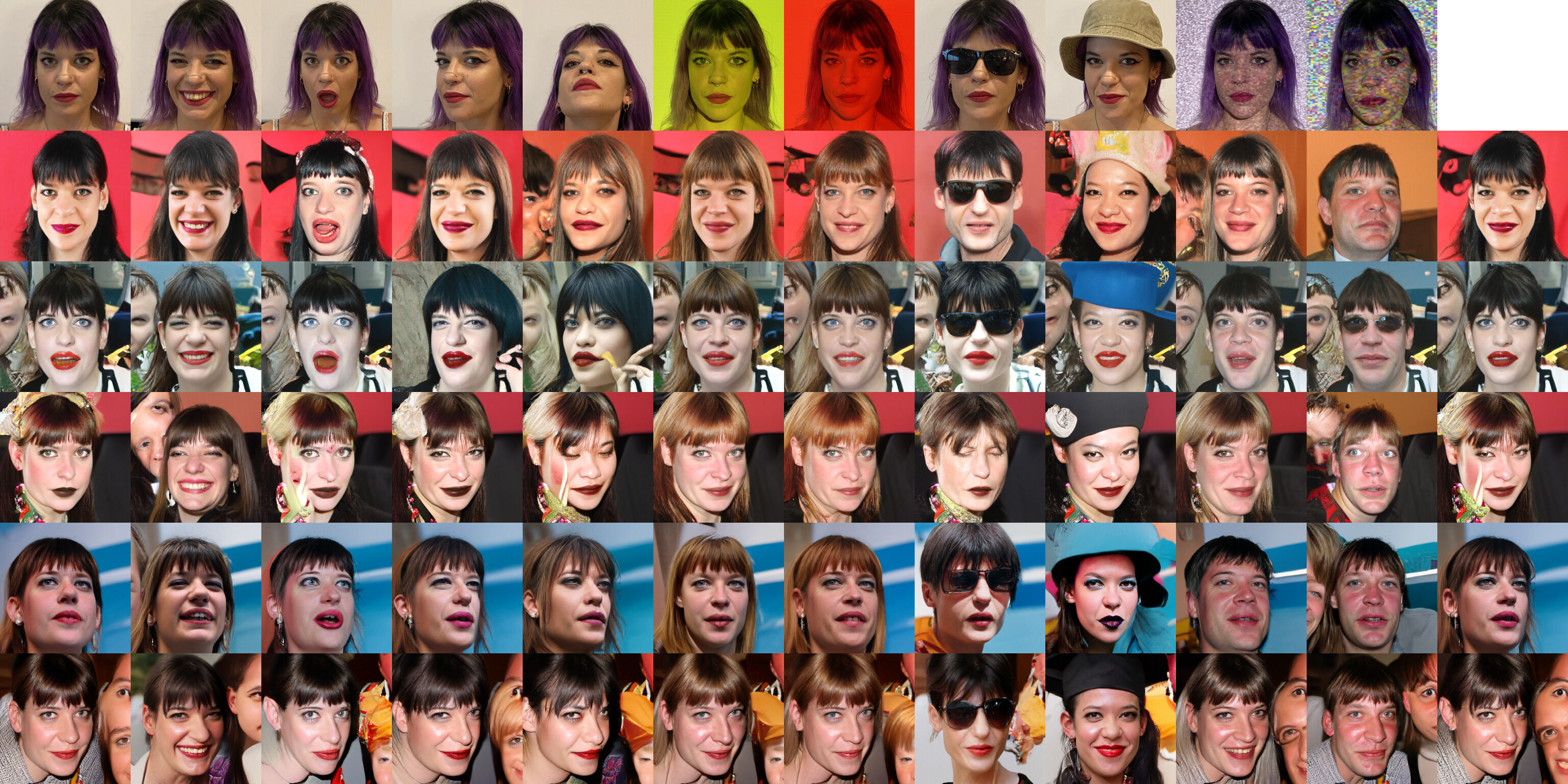}} & \\
    & \begin{tabularx}{0.8\textwidth}{ *{12}{Y} } $0.29$ & $0.24$ & $0.27$ & $0.24$ & $0.27$ & $0.24$ & $0.27$ & $0.35$ & $0.31$ & $0.26$ & $0.35$ & {-} \end{tabularx} & $0.43$ \\
    ArcFace~\cite{arcface, gaussian_sampling} & \raisebox{-.5\height}{\adjincludegraphics[width=.8\textwidth, trim={0 {.667\height} 0 {0.167\height}}, clip]{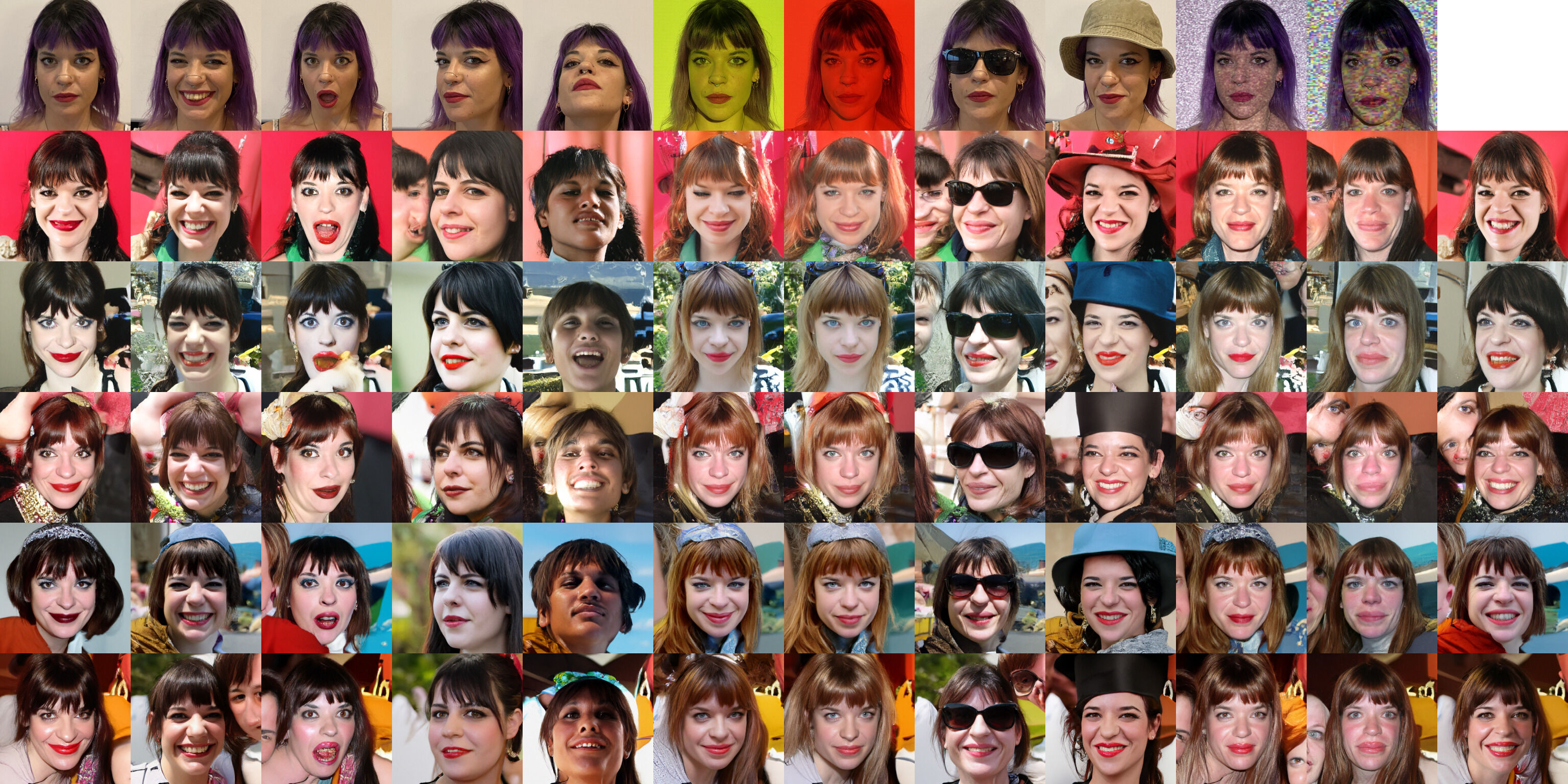}} & \\
    & \begin{tabularx}{0.8\textwidth}{ *{12}{Y} } $0.28$ & $0.25$ & $0.25$ & $0.27$ & $0.30$ & $0.26$ & $0.23$ & $0.26$ & $0.28$ & $0.25$ & $0.28$ & {-} \end{tabularx} & $0.45$ \\
    FaceNet~\cite{facenet, facenet_pytorch} & \raisebox{-.5\height}{\adjincludegraphics[width=.8\textwidth, trim={0 {.667\height} 0 {0.167\height}}, clip]{images/robustness/Melissa_facenet.jpg}} & \\
    & \begin{tabularx}{0.8\textwidth}{ *{12}{Y} } $0.15$ & $0.19$ & $0.16$ & $0.16$ & $0.14$ & $0.12$ & $0.14$ & $0.17$ & $0.14$ & $0.14$ & $0.19$ & {-} \end{tabularx} & $0.37$ \\
    FROM~\cite{from} & \raisebox{-.5\height}{\adjincludegraphics[width=.8\textwidth, trim={0 {.667\height} 0 {0.167\height}}, clip]{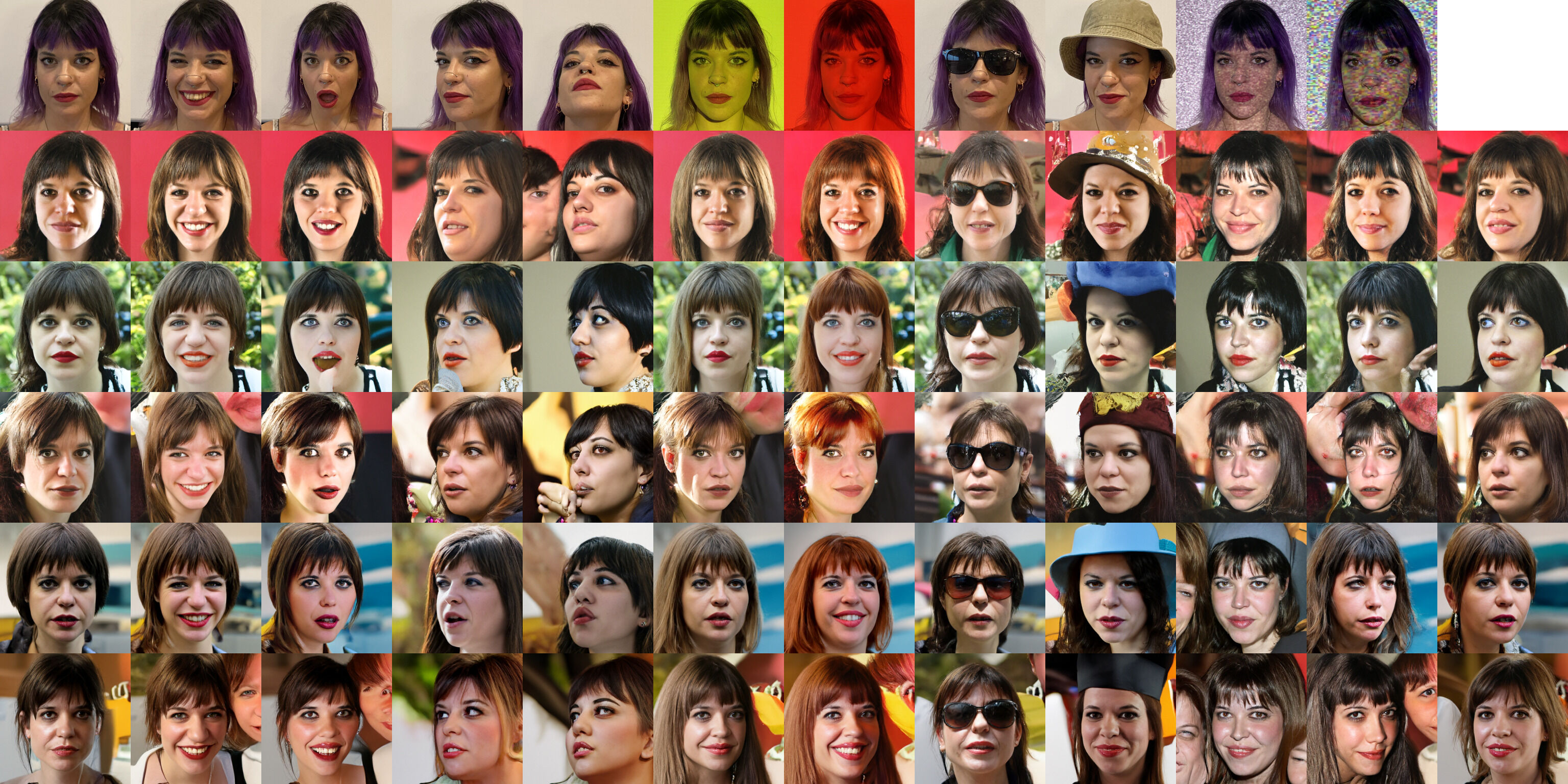}} & \\
    & \begin{tabularx}{0.8\textwidth}{ *{12}{Y} } $0.17$ & $0.16$ & $0.19$ & $0.19$ & $0.23$ & $0.23$ & $0.19$ & $0.18$ & $0.20$ & $0.18$ & $0.22$ & {-} \end{tabularx} & $0.42$ \\
    InsightFace~\cite{insightface} & \raisebox{-.5\height}{\adjincludegraphics[width=.8\textwidth, trim={0 {.667\height} 0 {0.167\height}}, clip]{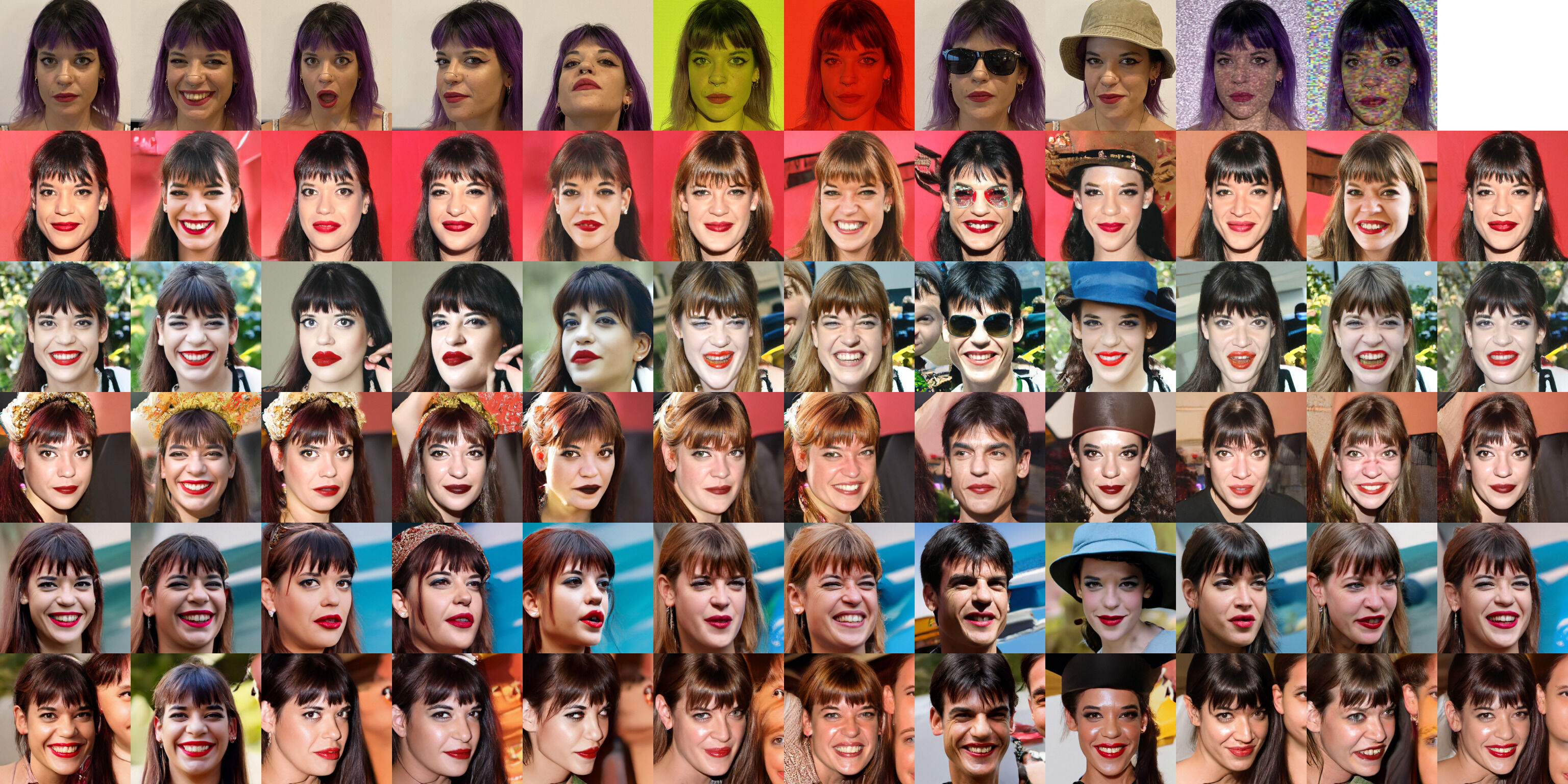}} & \\
    & \begin{tabularx}{0.8\textwidth}{ *{12}{Y} } $0.23$ & $0.24$ & $0.21$ & $0.23$ & $0.24$ & $0.22$ & $0.21$ & $0.25$ & $0.23$ & $0.25$ & $0.24$ & {-} \end{tabularx} & $0.42$ \\
    \end{tabular}
    }
    \addtolength{\tabcolsep}{2pt}
    \caption{Identity 1}
    \vspace{5mm}
\end{subfigure}
\begin{subfigure}{\textwidth}
\centering
    \addtolength{\tabcolsep}{-2pt}
    \scriptsize{
    \begin{tabular}{lcc}
    & \raisebox{-.5\height}{\adjincludegraphics[width=.8\textwidth, trim={0 {.801\height} 0 0}, clip]{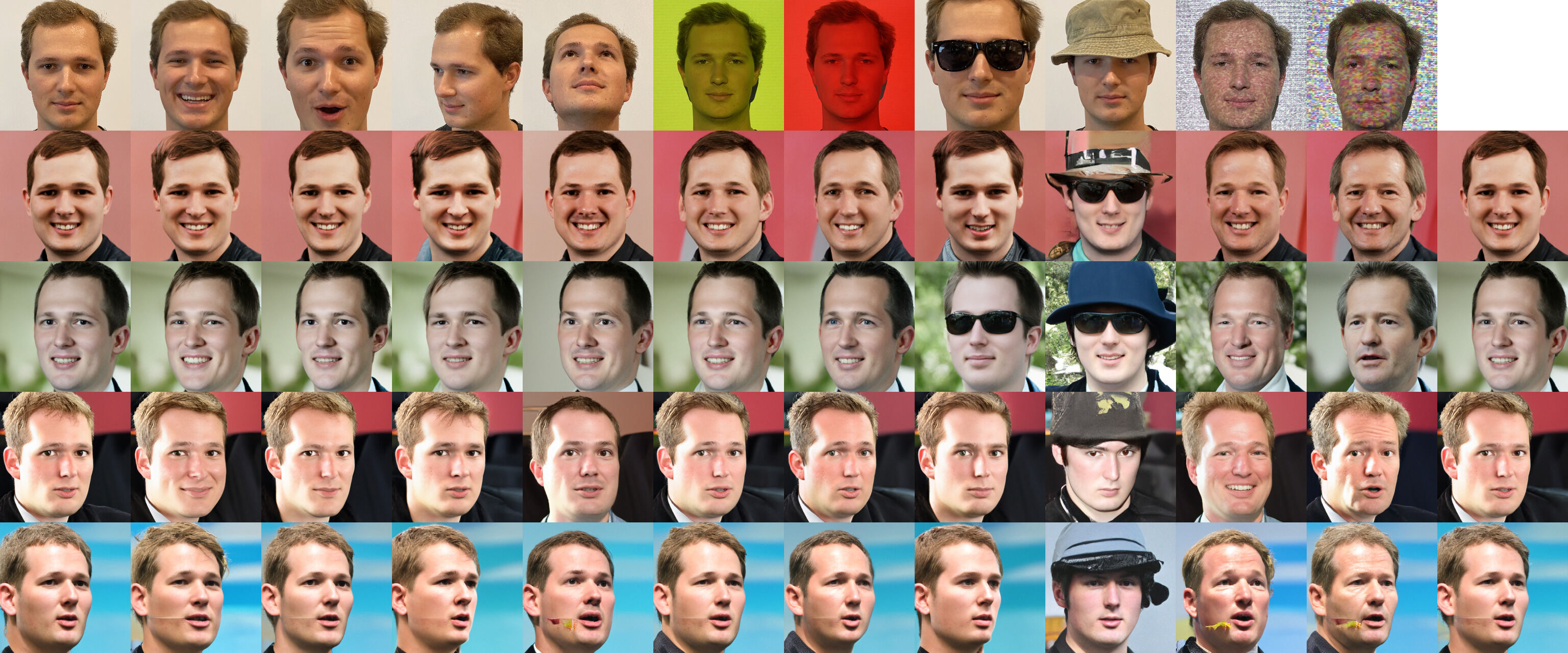}} & \begin{tabular}[c]{@{}c@{}}LFW \\ threshold \end{tabular} \\
    \\[-0.2cm]
    AdaFace~\cite{adaface} & \raisebox{-.5\height}{\adjincludegraphics[width=.8\textwidth, trim={0 {.2\height} 0 {0.6\height}}, clip]{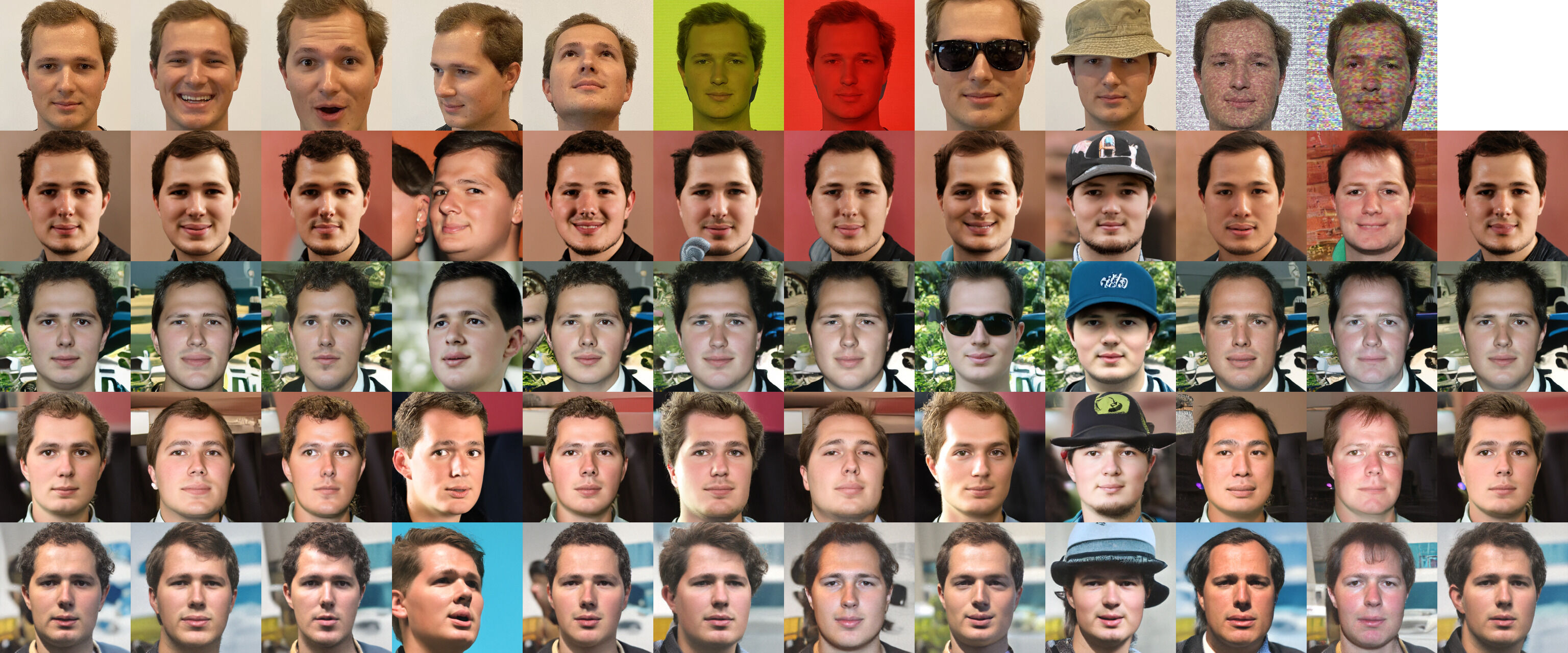}} & \\
    & \begin{tabularx}{0.8\textwidth}{ *{12}{Y} } $0.28$ & $0.29$ & $0.29$ & $0.28$ & $0.27$ & $0.28$ & $0.28$ & $0.26$ & $0.25$ & $0.40$ & $0.32$ & {-} \end{tabularx} & $0.43$ \\
    ArcFace~\cite{arcface, gaussian_sampling} & \raisebox{-.5\height}{\adjincludegraphics[width=.8\textwidth, trim={0 {.2\height} 0 {0.6\height}}, clip]{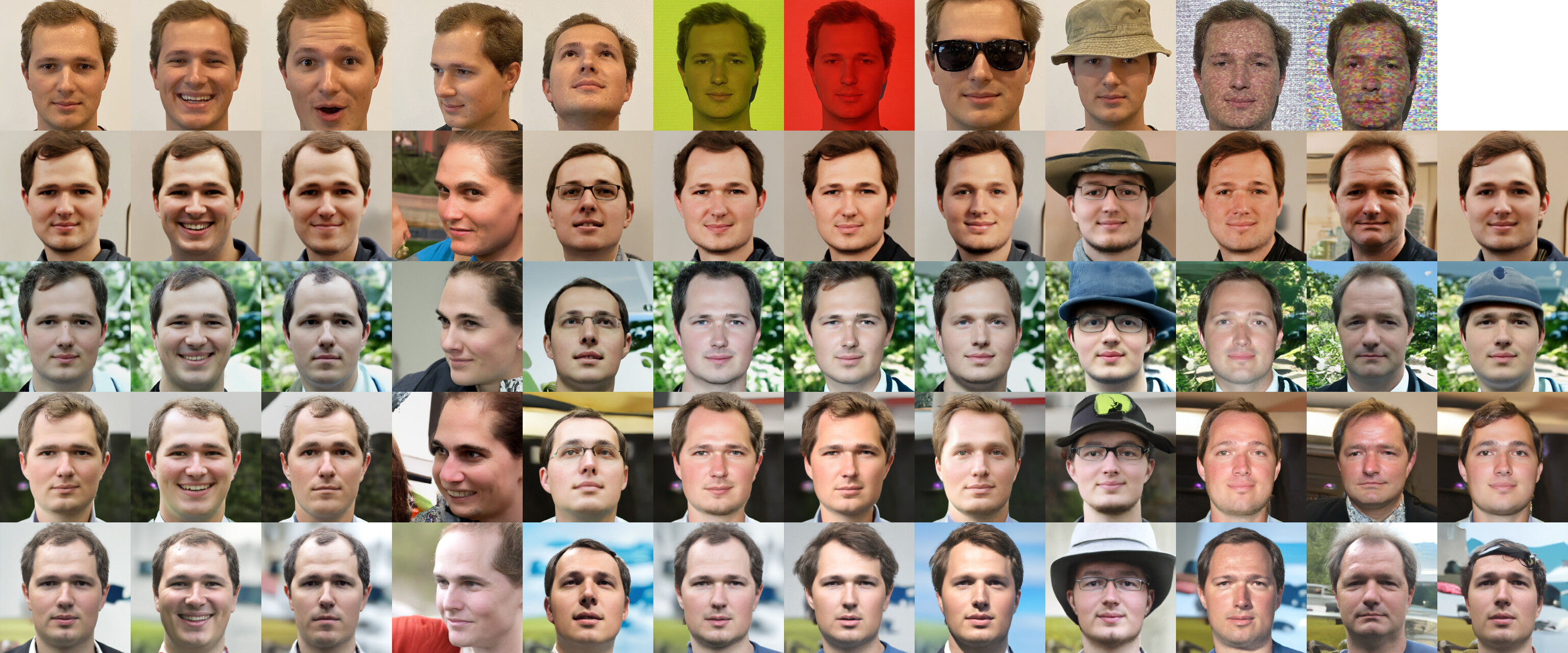}} & \\
    & \begin{tabularx}{0.8\textwidth}{ *{12}{Y} } $0.23$ & $0.25$ & $0.22$ & $0.30$ & $0.24$ & $0.22$ & $0.24$ & $0.23$ & $0.27$ & $0.24$ & $0.23$ & {-} \end{tabularx} & $0.45$ \\
    FaceNet~\cite{facenet, facenet_pytorch} & \raisebox{-.5\height}{\adjincludegraphics[width=.8\textwidth, trim={0 {.2\height} 0 {0.6\height}}, clip]{images/robustness/Lucas_facenet.jpg}} & \\
    & \begin{tabularx}{0.8\textwidth}{ *{12}{Y} } $0.18$ & $0.17$ & $0.19$ & $0.16$ & $0.15$ & $0.17$ & $0.17$ & $0.17$ & $0.18$ & $0.13$ & $0.21$ & {-} \end{tabularx} & $0.37$ \\
    FROM~\cite{from} & \raisebox{-.5\height}{\adjincludegraphics[width=.8\textwidth, trim={0 {.2\height} 0 {0.6\height}}, clip]{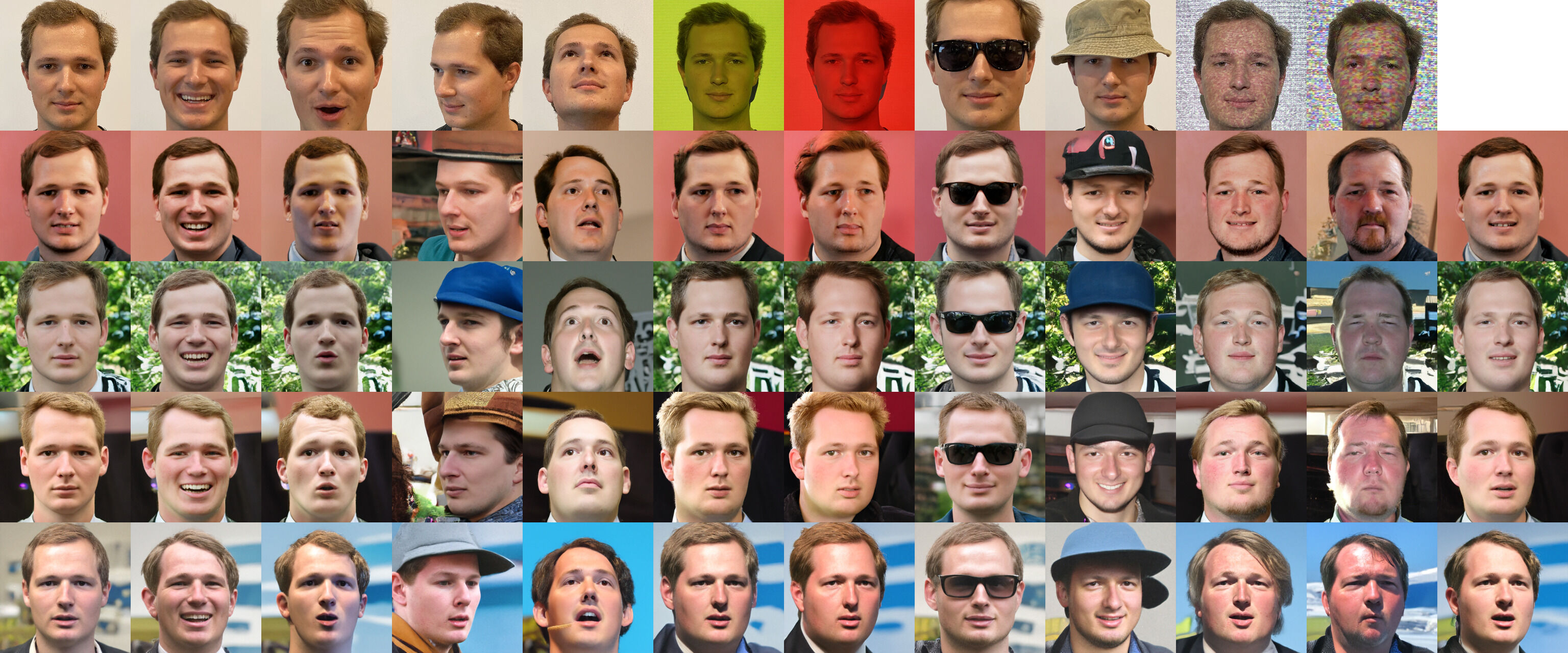}} & \\
    & \begin{tabularx}{0.8\textwidth}{ *{12}{Y} } $0.15$ & $0.12$ & $0.17$ & $0.23$ & $0.21$ & $0.15$ & $0.16$ & $0.17$ & $0.22$ & $0.16$ & $0.18$ & {-} \end{tabularx} & $0.42$ \\
    InsightFace~\cite{insightface} & \raisebox{-.5\height}{\adjincludegraphics[width=.8\textwidth, trim={0 {.2\height} 0 {0.6\height}}, clip]{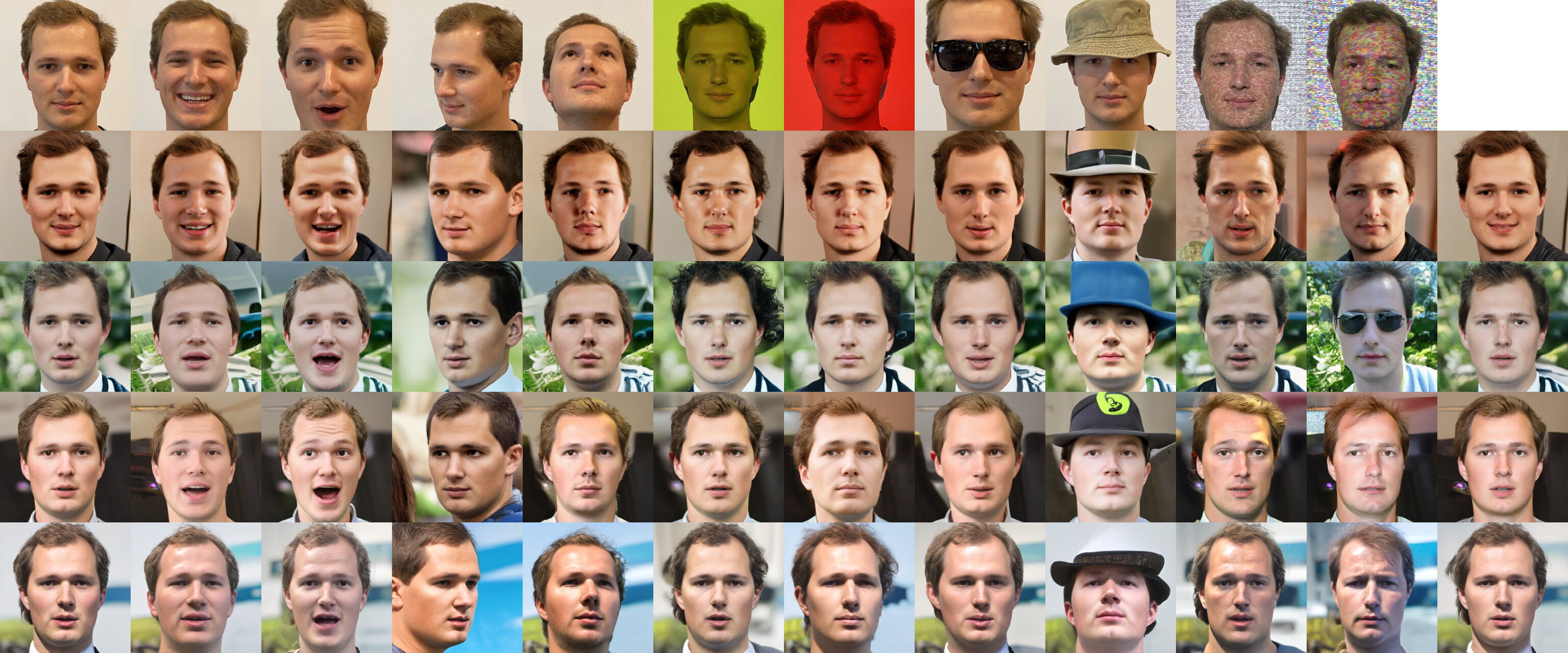}} & \\
    & \begin{tabularx}{0.8\textwidth}{ *{12}{Y} } $0.21$ & $0.22$ & $0.21$ & $0.21$ & $0.22$ & $0.23$ & $0.22$ & $0.23$ & $0.23$ & $0.23$ & $0.24$ & {-} \end{tabularx} & $0.42$ \\
    \end{tabular}
    }
    \addtolength{\tabcolsep}{2pt}
    \caption{Identity 2}
\end{subfigure}
    \caption{Robustness experiment. The first row shows images of a source identity in challenging scenarios whereas the remaining rows show the results when using different ID vectors. The last image column shows images generated from the mean ID vector of all of the source identity images (for which no source image exists). The numbers under each line are the angular distances between the identity of the generated images and the target identity for the same face recognition method. The numbers in the last column are the optimal thresholds for that method calculated using real images of the LFW~\cite{lfw} data set and official protocol.}
    \label{fig:robustness_2}
\end{figure*}

Our method can also be used to analyze and visualize the robustness of face recognition models in difficult scenarios such as varying expressions, poses, lighting, occlusions, and noise as seen in \cref{fig:robustness_2}. In line with our previous observations, FaceNet~\cite{facenet, facenet_pytorch} and InsightFace~\cite{insightface} appear the most robust. 

In this analysis, it is also relatively easy to tell which features are extracted by observing which features of a target identity's image are preserved. For example, ArcFace~\cite{arcface, gaussian_sampling} and FROM~\cite{from} seem to contain pose information as the generated images in the fourth and fifth columns have similar poses as the target identities' images for both identities. Similarly, AdaFace~\cite{adaface} and ArcFace~\cite{arcface, gaussian_sampling} seem to copy the expression for the third column of the first identity. InsightFace~\cite{insightface} also seems to contain expressions and pose for the second identity as seen in columns two to four. Another feature that is commonly copied is whether a person is wearing a hat or not even though this should arguably not be considered part of a person's identity. Interestingly, FROM, a method specifically aimed to mask out corrupted features, does not appear more robust for the tested occlusions (sunglasses, hat). Lastly, noise seems to affect most face recognition methods significantly.

\Cref{fig:robustness_2} also lists the angular distances of the identities for each generated image to the target identity for the same FR method. The distances for generated images that can be considered failure cases are in general higher than those of the images that worked better. However, they are all still below the optimal threshold\footnote{It is actually the mean threshold over the $10$ different splits. Note that the standard deviation across the splits is smaller than $0.005$.} calculated for each FR method for real images of the LFW~\cite{lfw} data set using the official protocol -- meaning that all generated images are considered to be of the same person. Therefore, we argue that the wrong ID reconstructions are mostly due to problems in the ID vectors rather than the inversion of our model.

\begin{figure*}[htpb]
    \centering
    \addtolength{\tabcolsep}{-5pt}
    \small{
    \begin{tabular}{cccccc}
        \includegraphics[width=0.13\textwidth]{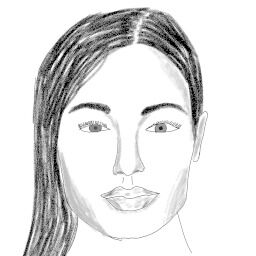} & 
        \includegraphics[width=0.13\textwidth]{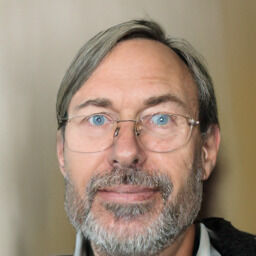} & 
        \includegraphics[width=0.13\textwidth]{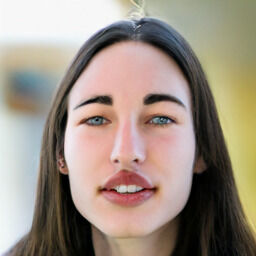} & 
        \includegraphics[width=0.13\textwidth]{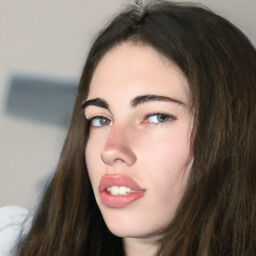} & 
        \includegraphics[width=0.13\textwidth]{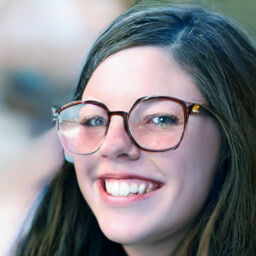} & 
        \includegraphics[width=0.13\textwidth]{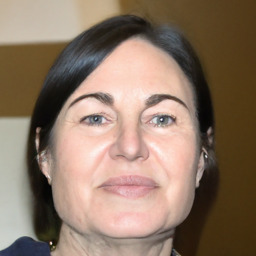} \\
        \\[-0.46cm]
        \includegraphics[width=0.13\textwidth]{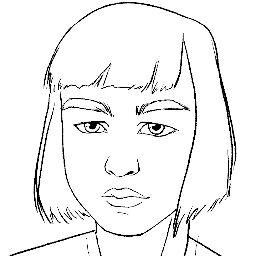} & 
        \includegraphics[width=0.13\textwidth]{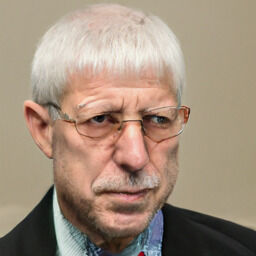} & 
        \includegraphics[width=0.13\textwidth]{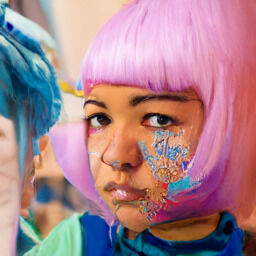} & 
        \includegraphics[width=0.13\textwidth]{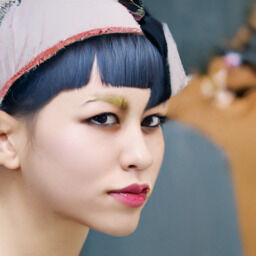} & 
        \includegraphics[width=0.13\textwidth]{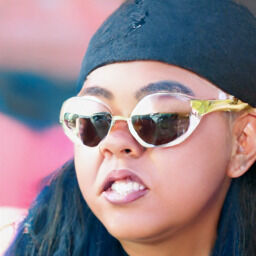} & 
        \includegraphics[width=0.13\textwidth]{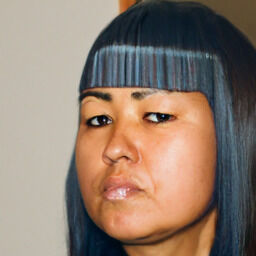} \\
        \\[-0.46cm]
        \includegraphics[width=0.13\textwidth]{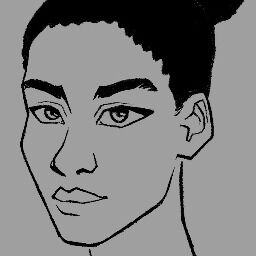} & 
        \includegraphics[width=0.13\textwidth]{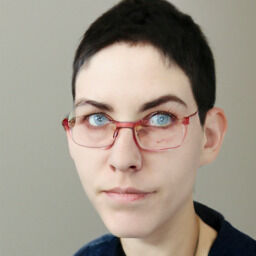} & 
        \includegraphics[width=0.13\textwidth]{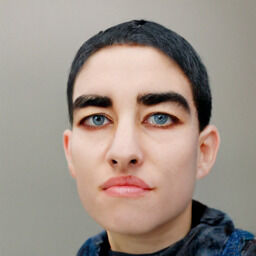} & 
        \includegraphics[width=0.13\textwidth]{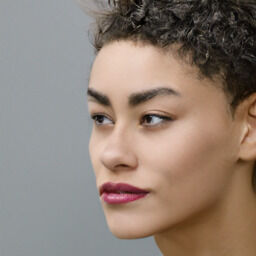} & 
        \includegraphics[width=0.13\textwidth]{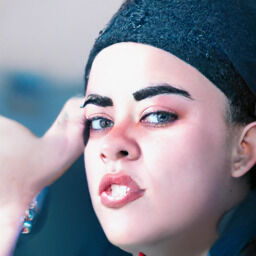} & 
        \includegraphics[width=0.13\textwidth]{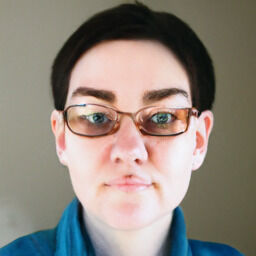} \\
        \\[-0.46cm]
        \includegraphics[width=0.13\textwidth]{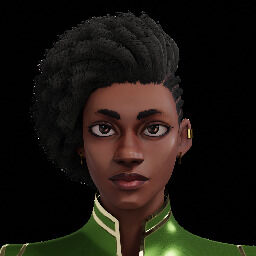} & 
        \includegraphics[width=0.13\textwidth]{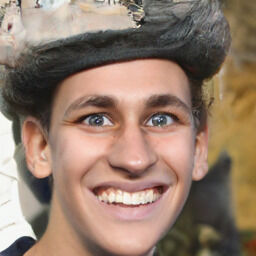} & 
        \includegraphics[width=0.13\textwidth]{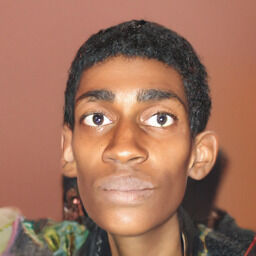} & 
        \includegraphics[width=0.13\textwidth]{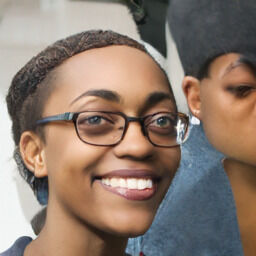} & 
        \includegraphics[width=0.13\textwidth]{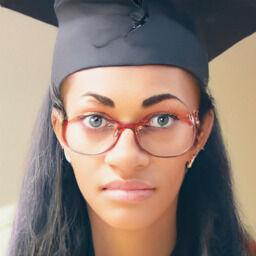} & 
        \includegraphics[width=0.13\textwidth]{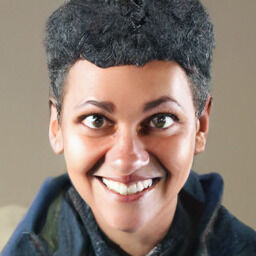} \\
        \\[-0.46cm]
        \includegraphics[width=0.13\textwidth]{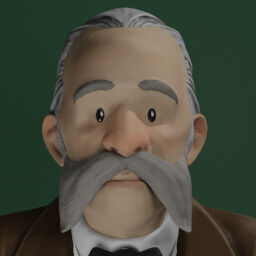} & 
        \includegraphics[width=0.13\textwidth]{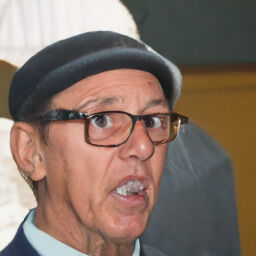} & 
        \includegraphics[width=0.13\textwidth]{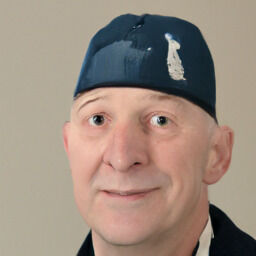} & 
        \includegraphics[width=0.13\textwidth]{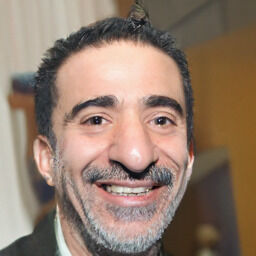} & 
        \includegraphics[width=0.13\textwidth]{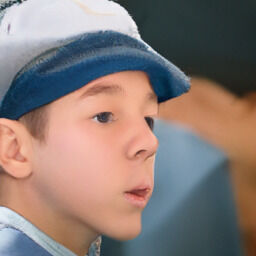} & 
        \includegraphics[width=0.13\textwidth]{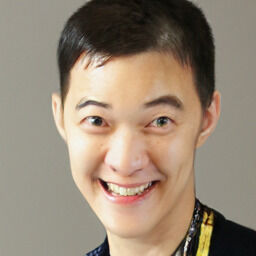} \\
        \\[-0.46cm]
        \includegraphics[width=0.13\textwidth]{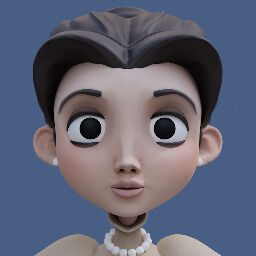} & 
        \includegraphics[width=0.13\textwidth]{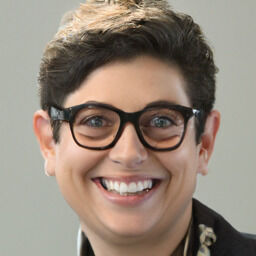} & 
        \includegraphics[width=0.13\textwidth]{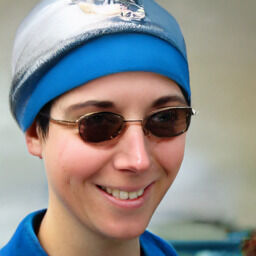} & 
        \includegraphics[width=0.13\textwidth]{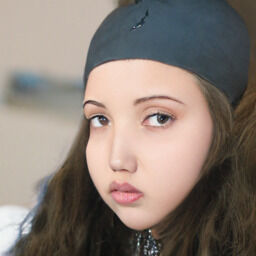} & 
        \includegraphics[width=0.13\textwidth]{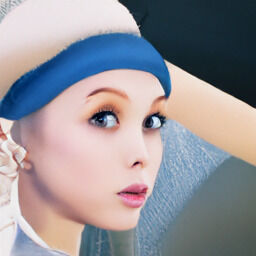} & 
        \includegraphics[width=0.13\textwidth]{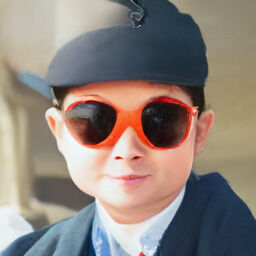} \\
        
        \multirow{2}{*}{\begin{tabular}[c]{@{}c@{}}Original \\ image \end{tabular}} & \multirow{2}{*}{AdaFace~\cite{adaface}} & \multirow{2}{*}{ArcFace~\cite{arcface, gaussian_sampling}} & \multirow{2}{*}{FaceNet~\cite{facenet, facenet_pytorch}} &  \multirow{2}{*}{FROM~\cite{from}} & \multirow{2}{*}{\begin{tabular}[c]{@{}c@{}}Insight- \\ Face~\cite{insightface} \end{tabular}} \\
        \\
    \end{tabular}
    }
    \addtolength{\tabcolsep}{5pt}
    \caption{Robustness of ID vectors from different state-of-the-art face recognition models for out-of-distribution samples. Note that the same seed was used for all images to obtain the most fair results.}
    \label{fig:out_of_distr}
\end{figure*}

We further experiment with out-of-distribution samples such as drawings and digital renders as shown in \cref{fig:out_of_distr}. Interestingly, despite the extremely difficult setup, some of the resulting images resemble the identity fairly well, especially for FaceNet~\cite{facenet, facenet_pytorch}, demonstrating that some face recognition models can extract reasonable identity-specific features even from faces that are out of distribution. Furthermore, this experiment shows that our method can generate extreme features, such as the large nose of the man in the fifth row with FaceNet~\cite{facenet, facenet_pytorch}, that are likely not in the training data set.

\clearpage

\subsection{Identity interpolations}

\begin{figure*}[htpb]
\centering
\begin{subfigure}{0.95\textwidth}
\centering
    \addtolength{\tabcolsep}{-2pt}
    \small{
    \begin{tabular}{lc}
    AdaFace~\cite{adaface} & \raisebox{-.5\height}{\adjincludegraphics[width=.85\textwidth, trim={0 {.75\height} 0 0}, clip]{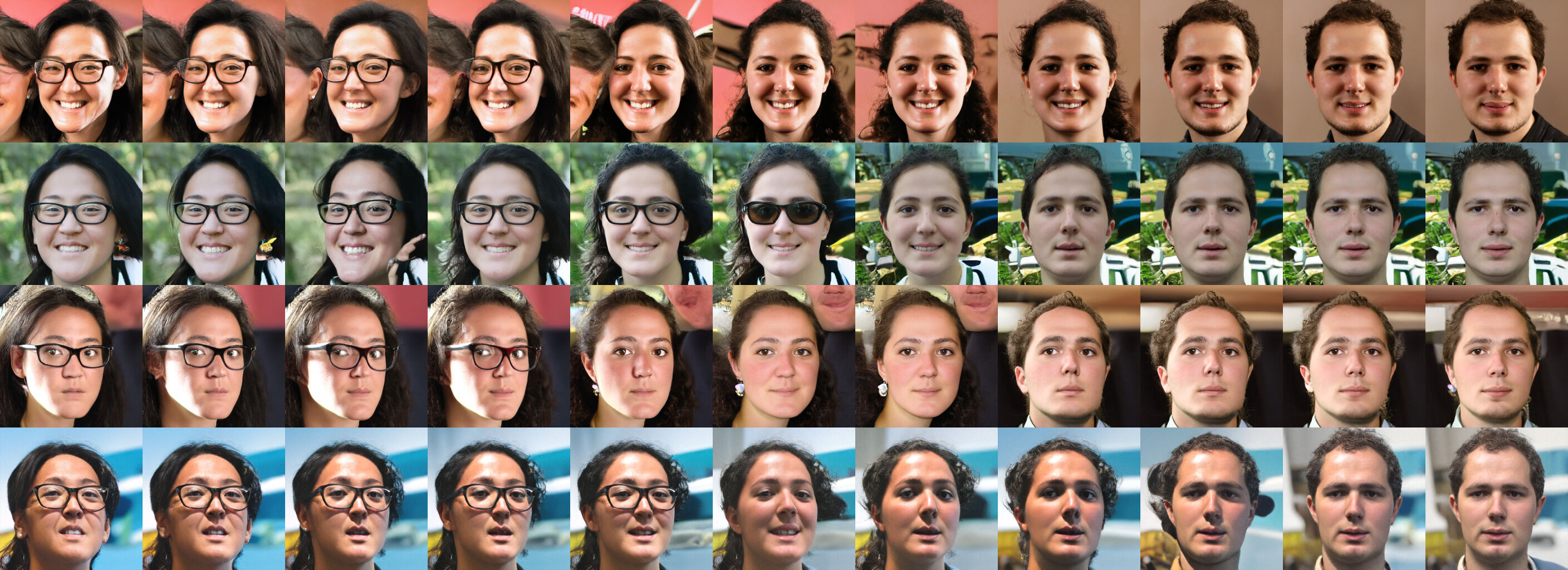}} \\
    \\[-0.35cm]
    ArcFace~\cite{arcface, gaussian_sampling} & \raisebox{-.5\height}{\adjincludegraphics[width=.85\textwidth, trim={0 {.75\height} 0 0}, clip]{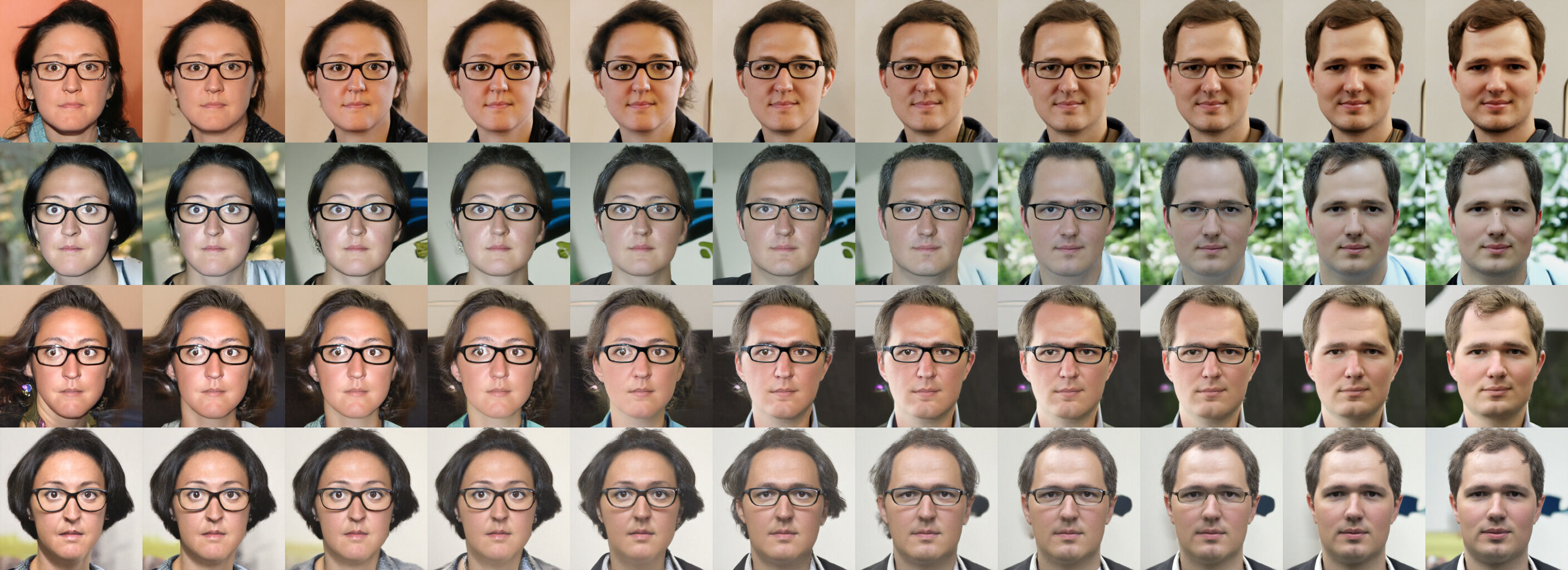}} \\
    \\[-0.35cm]
    FaceNet~\cite{facenet, facenet_pytorch} & \raisebox{-.5\height}{\adjincludegraphics[width=.85\textwidth, trim={0 {.75\height} 0 0}, clip]{images/interpolations/sally_lucas/facenet.jpg}} \\
    \\[-0.35cm]
    FROM~\cite{from} & \raisebox{-.5\height}{\adjincludegraphics[width=.85\textwidth, trim={0 {.75\height} 0 0}, clip]{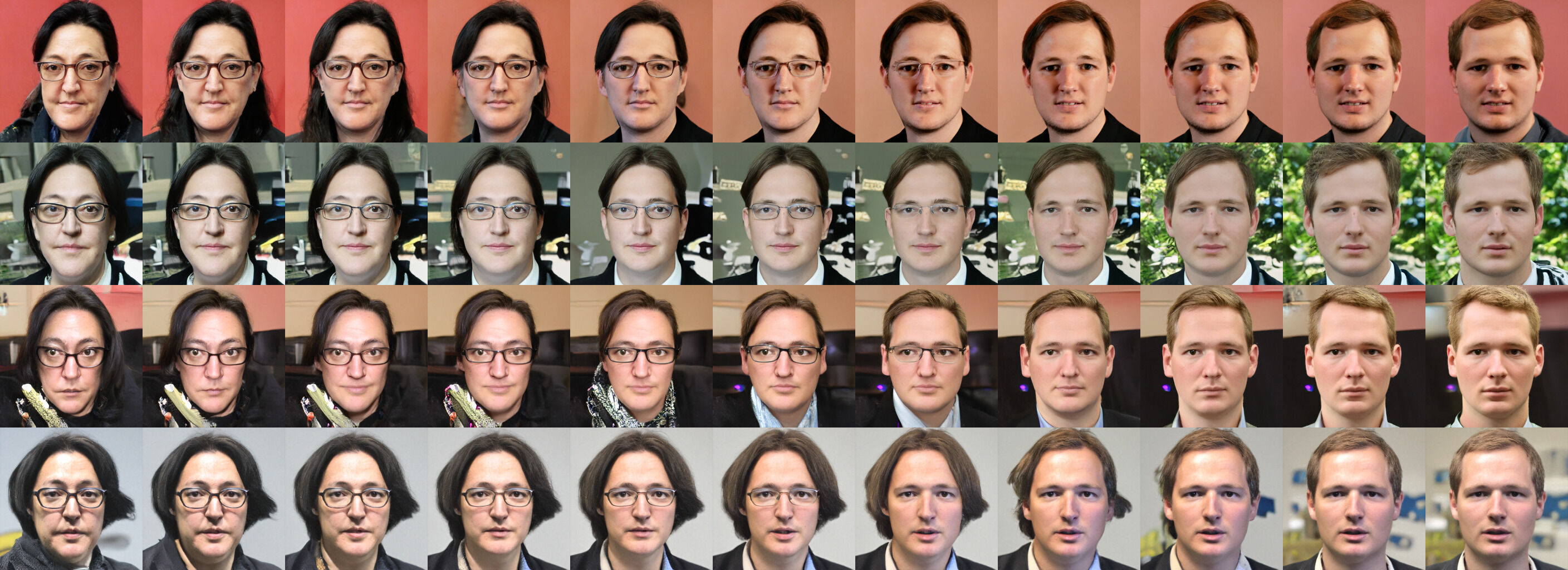}} \\
    \\[-0.35cm]
    InsightFace~\cite{insightface} & \raisebox{-.5\height}{\adjincludegraphics[width=.85\textwidth, trim={0 {.75\height} 0 0}, clip]{images/interpolations/sally_lucas/insightface.jpg}} \\
    \end{tabular}
    }
    \addtolength{\tabcolsep}{2pt}
    \caption{Identity 1 $\longleftrightarrow$ Identity 2}
    \vspace{5mm}
\end{subfigure}
\begin{subfigure}{0.95\textwidth}
\centering
    \addtolength{\tabcolsep}{-2pt}
    \small{
    \begin{tabular}{lc}
    AdaFace~\cite{adaface} & \raisebox{-.5\height}{\adjincludegraphics[width=.85\textwidth, trim={0 {.75\height} 0 0}, clip]{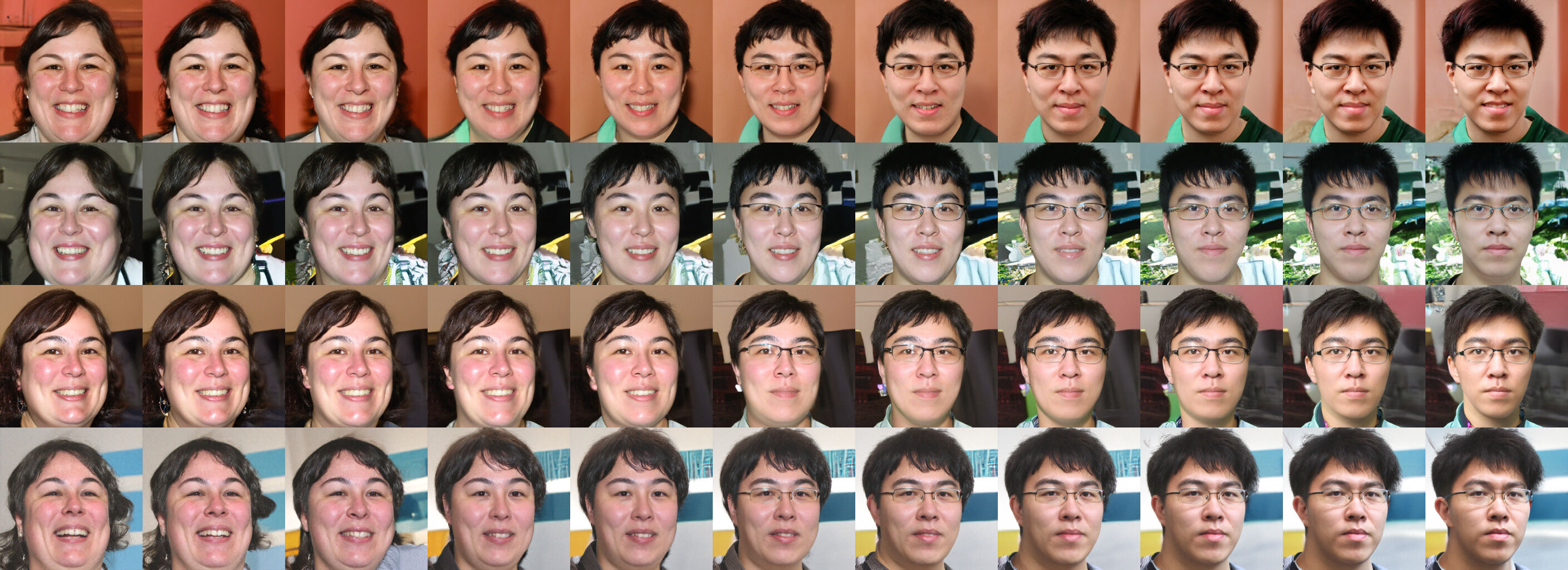}} \\
    \\[-0.35cm]
    ArcFace~\cite{arcface, gaussian_sampling} & \raisebox{-.5\height}{\adjincludegraphics[width=.85\textwidth, trim={0 {.75\height} 0 0}, clip]{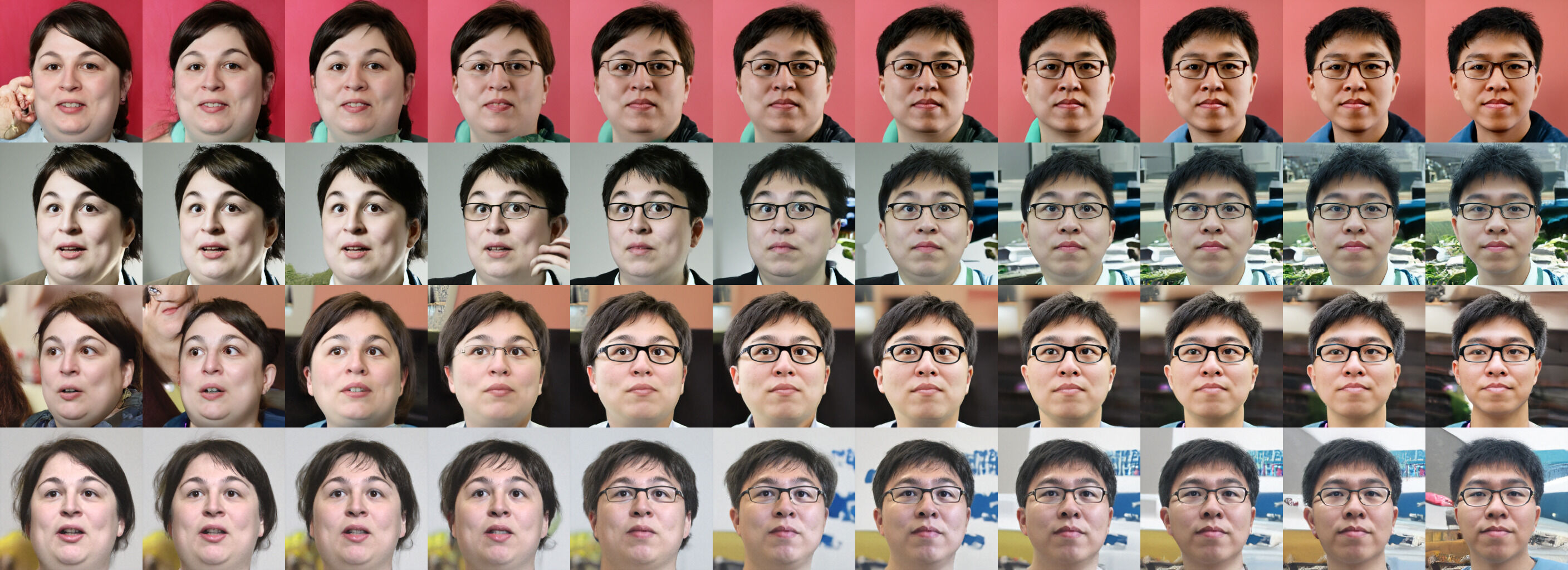}} \\
    \\[-0.35cm]
    FaceNet~\cite{facenet, facenet_pytorch} & \raisebox{-.5\height}{\adjincludegraphics[width=.85\textwidth, trim={0 {.75\height} 0 0}, clip]{images/interpolations/sandra_xianyao/facenet.jpg}} \\
    \\[-0.35cm]
    FROM~\cite{from} & \raisebox{-.5\height}{\adjincludegraphics[width=.85\textwidth, trim={0 {.75\height} 0 0}, clip]{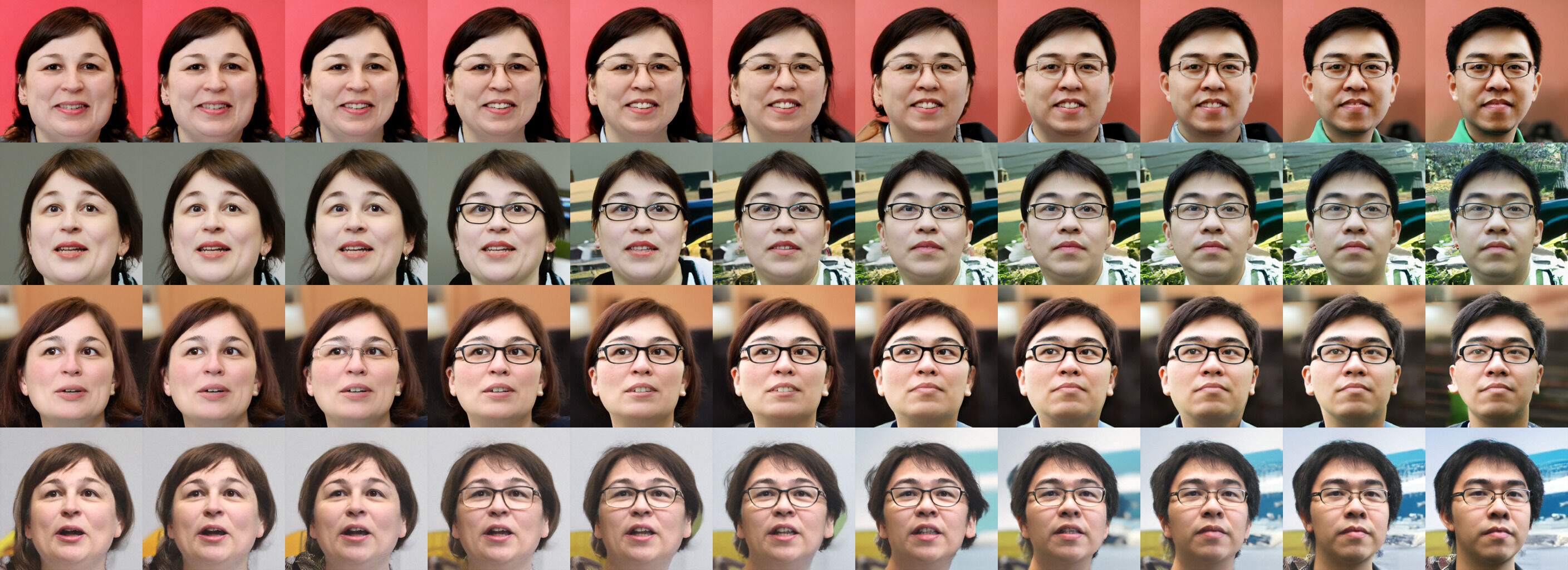}} \\
    \\[-0.35cm]
    InsightFace~\cite{insightface} & \raisebox{-.5\height}{\adjincludegraphics[width=.85\textwidth, trim={0 {.75\height} 0 0}, clip]{images/interpolations/sandra_xianyao/insightface.jpg}} \\
    \end{tabular}
    }
    \addtolength{\tabcolsep}{2pt}
    \caption{Identity 3 $\longleftrightarrow$ Identity 4}
\end{subfigure}
    \caption{Identity interpolations for two pairs of identities using different ID vectors.}
    \label{fig:interpolation}
\end{figure*}

By interpolating between two ID vectors, our method can produce new, intermediate identities, as shown in the main paper and in \cref{fig:interpolation}. This empirically demonstrates that the latent spaces of most face recognition methods are fairly well-structured. Note that we use spherical linear interpolation because it produces slightly smoother results compared to linear interpolation.

\clearpage

\subsection{Principal component analysis}

\begin{figure*}[htpb]
\centering
\begin{subfigure}{0.45\textwidth}
    \addtolength{\tabcolsep}{-4pt}
    \small{
    \begin{tabular}{llc}
    AdaFace~\cite{adaface} & \raisebox{-.5\height}{\adjincludegraphics[width=0.7\textwidth, trim={0 {0.7\height} 0 0}, clip]{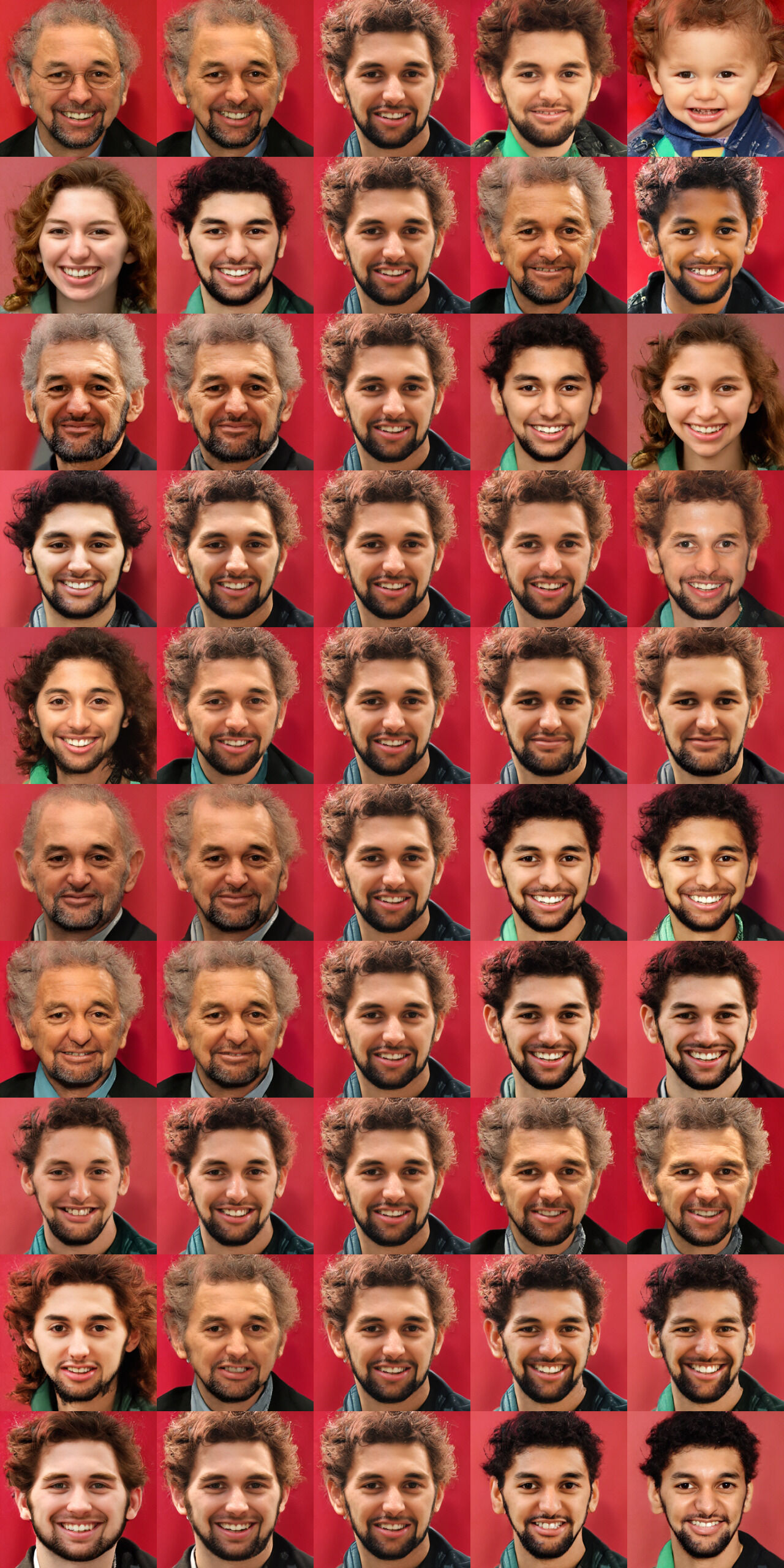}} & \begin{tabular}[c]{@{}l@{}} \\[-0.2cm] 1 \\[0.7cm] 2 \\[0.7cm] 3\end{tabular} \\
    \\[-0.32cm]
    ArcFace~\cite{arcface, gaussian_sampling} & \raisebox{-.5\height}{\adjincludegraphics[width=0.7\textwidth, trim={0 {0.7\height} 0 0}, clip]{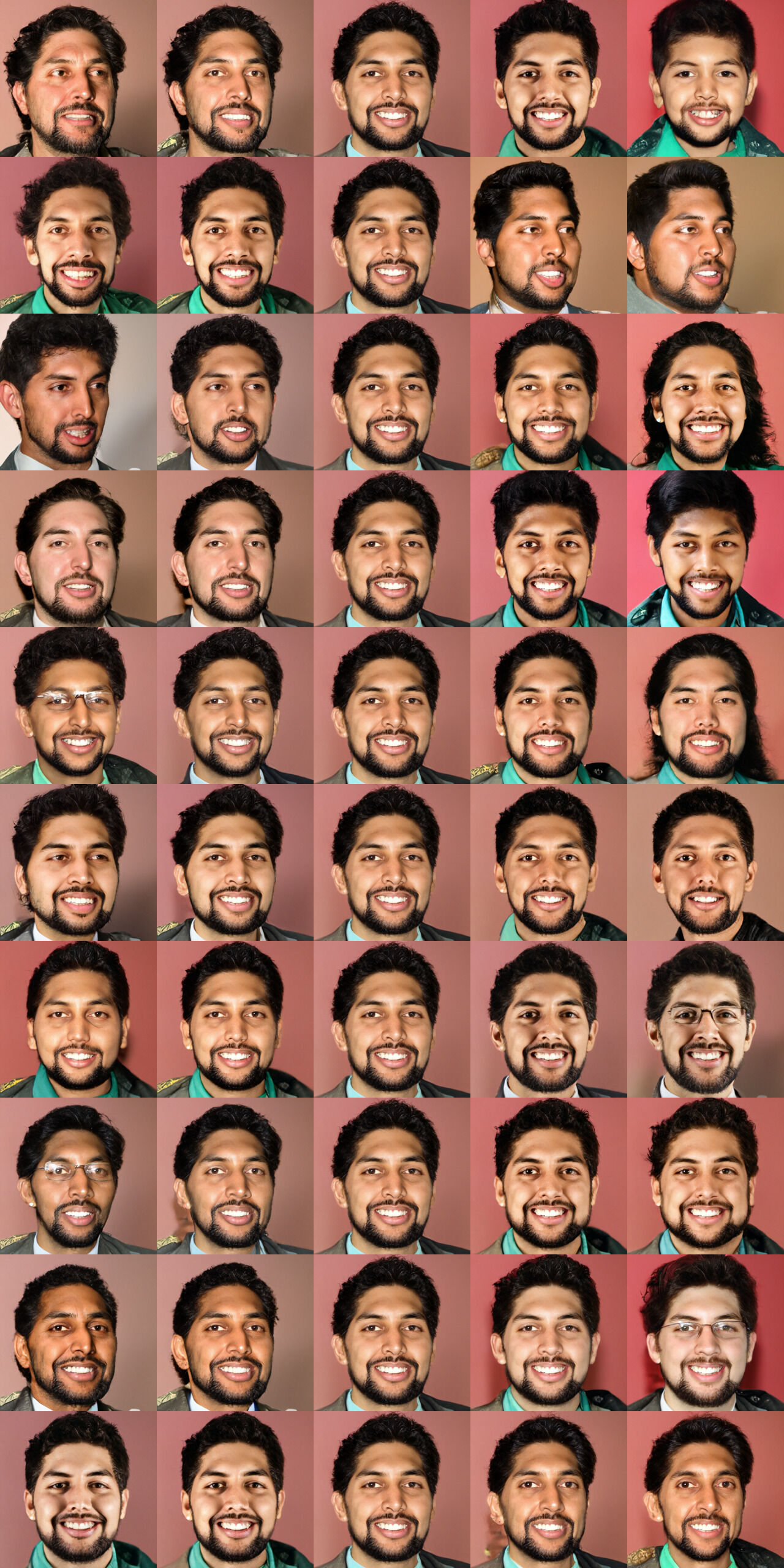}} & \begin{tabular}[c]{@{}l@{}} \\[-0.2cm] 1 \\[0.7cm] 2 \\[0.7cm] 3\end{tabular} \\
    \\[-0.32cm]
    FaceNet~\cite{facenet, facenet_pytorch} & \raisebox{-.5\height}{\adjincludegraphics[width=0.7\textwidth, trim={0 {0.7\height} 0 0}, clip]{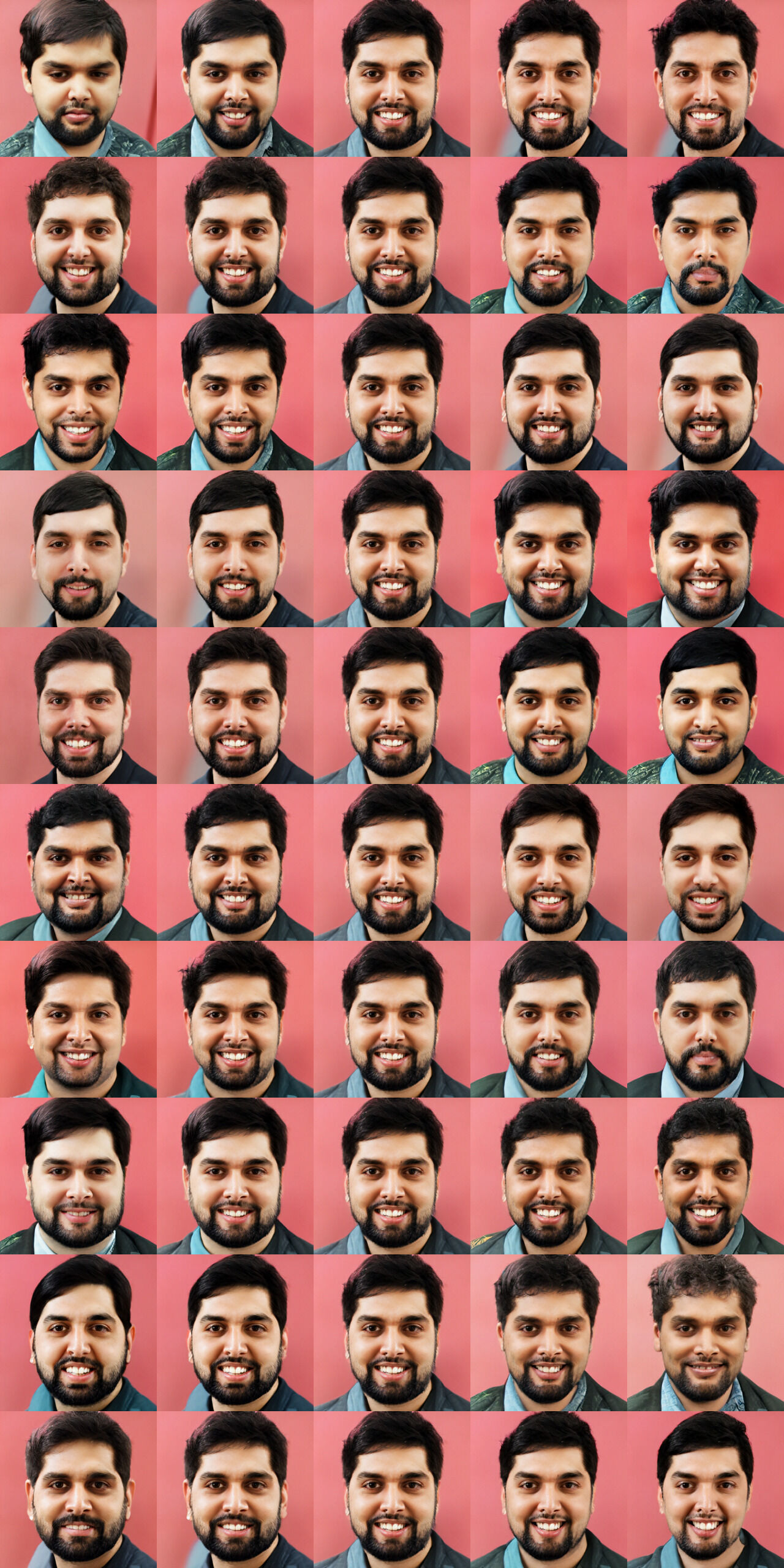}} & \begin{tabular}[c]{@{}l@{}} \\[-0.2cm] 1 \\[0.7cm] 2 \\[0.7cm] 3\end{tabular} \\
    \\[-0.32cm]
    FROM~\cite{from} & \raisebox{-.5\height}{\adjincludegraphics[width=0.7\textwidth, trim={0 {0.7\height} 0 0}, clip]{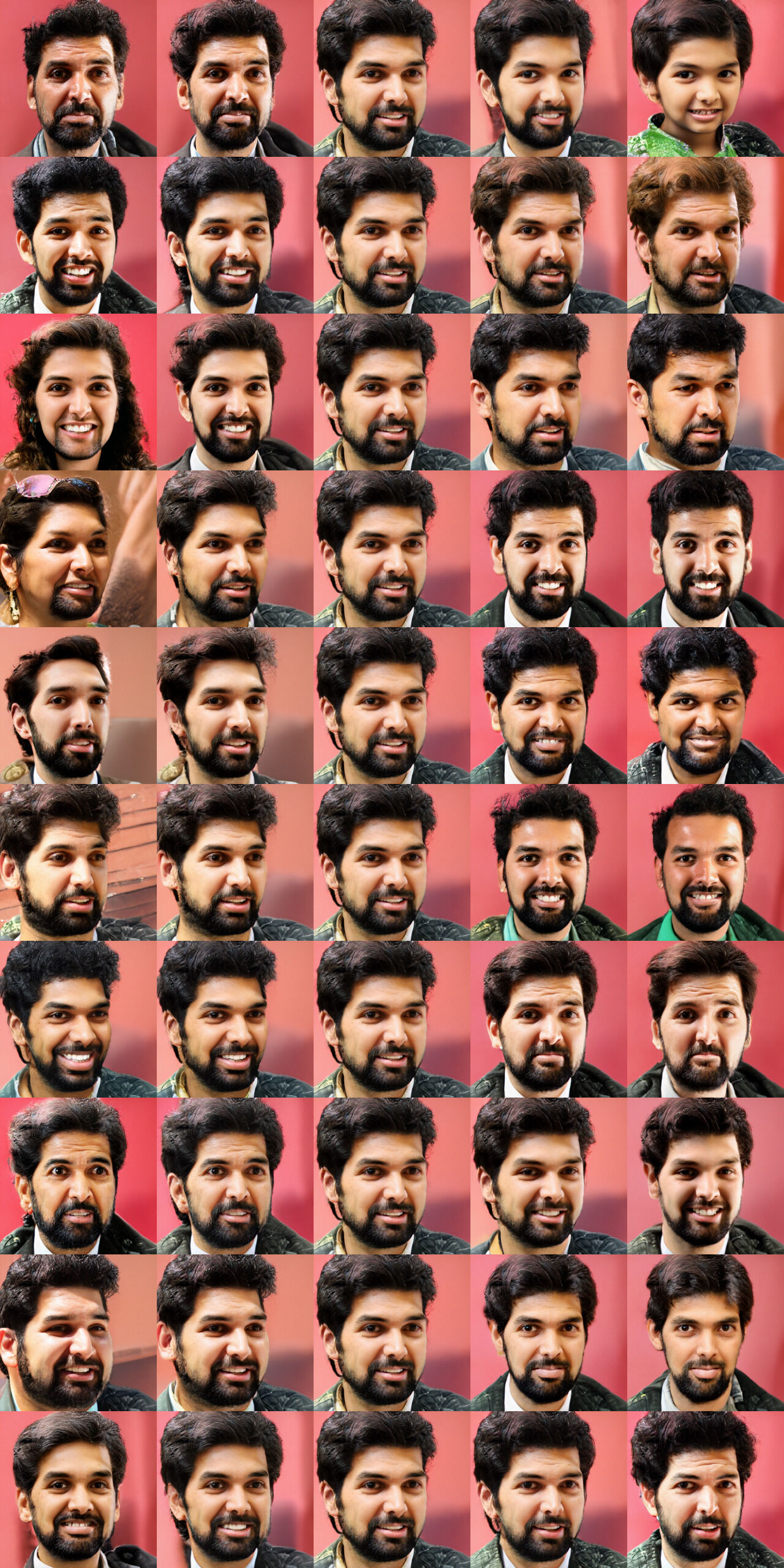}} & \begin{tabular}[c]{@{}l@{}} \\[-0.2cm] 1 \\[0.7cm] 2 \\[0.7cm] 3\end{tabular} \\
    \\[-0.32cm]
    InsightFace~\cite{insightface} \hspace{0.2cm} & \raisebox{-.5\height}{\adjincludegraphics[width=0.7\textwidth, trim={0 {0.7\height} 0 0}, clip]{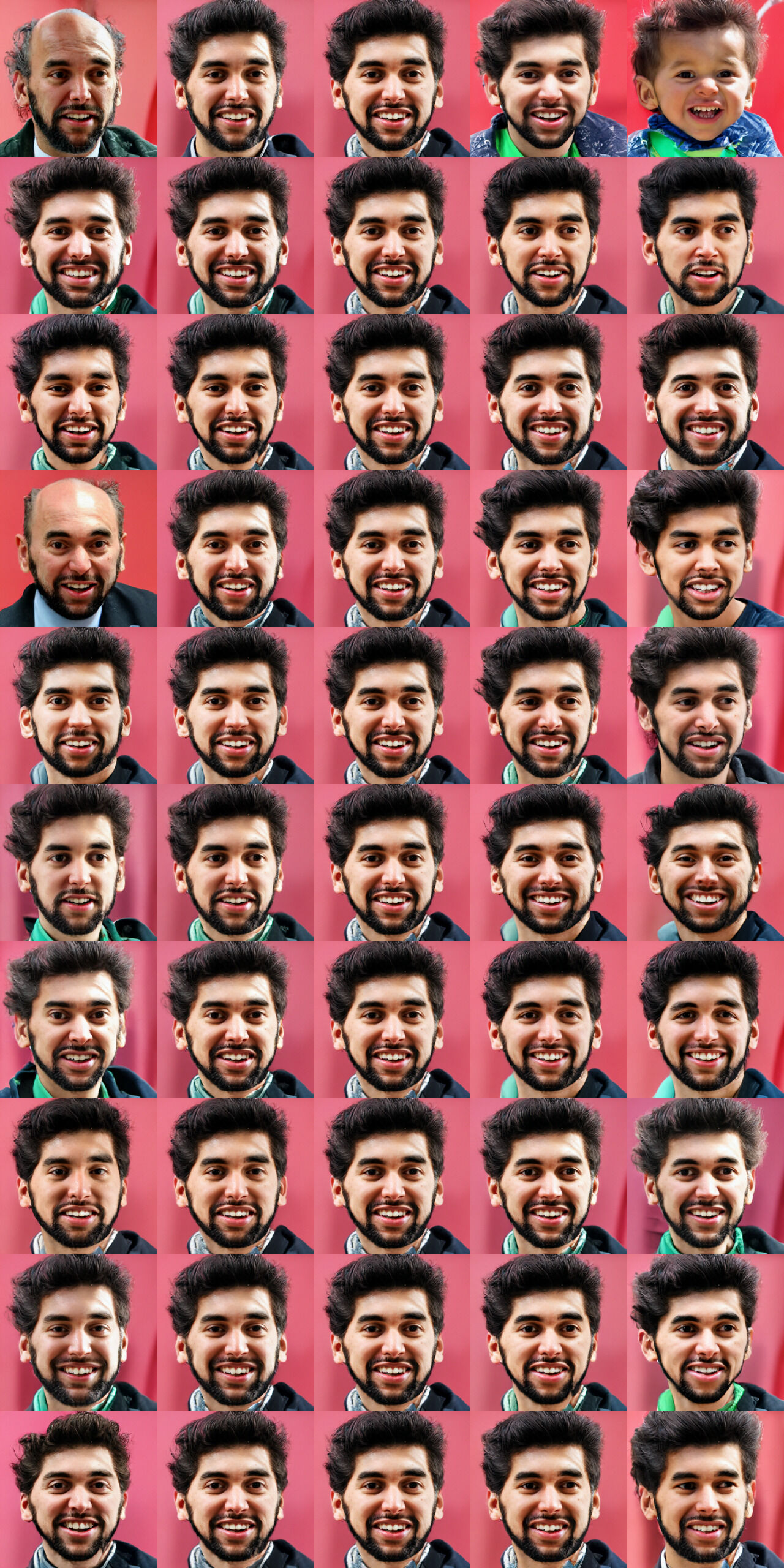}} & \begin{tabular}[c]{@{}l@{}} \\[-0.2cm] 1 \\[0.7cm] 2 \\[0.7cm] 3\end{tabular} \\
    \end{tabular}
    }
    \addtolength{\tabcolsep}{4pt}
    \caption*{\hspace{2.3cm} (a) Identity 1}
\end{subfigure}
\hspace{1cm}
\begin{subfigure}{0.45\textwidth}
    \addtolength{\tabcolsep}{-4pt}
    \small{
    \begin{tabular}{lc}
    \raisebox{-.5\height}{\adjincludegraphics[width=0.7\textwidth, trim={0 {0.7\height} 0 0}, clip]{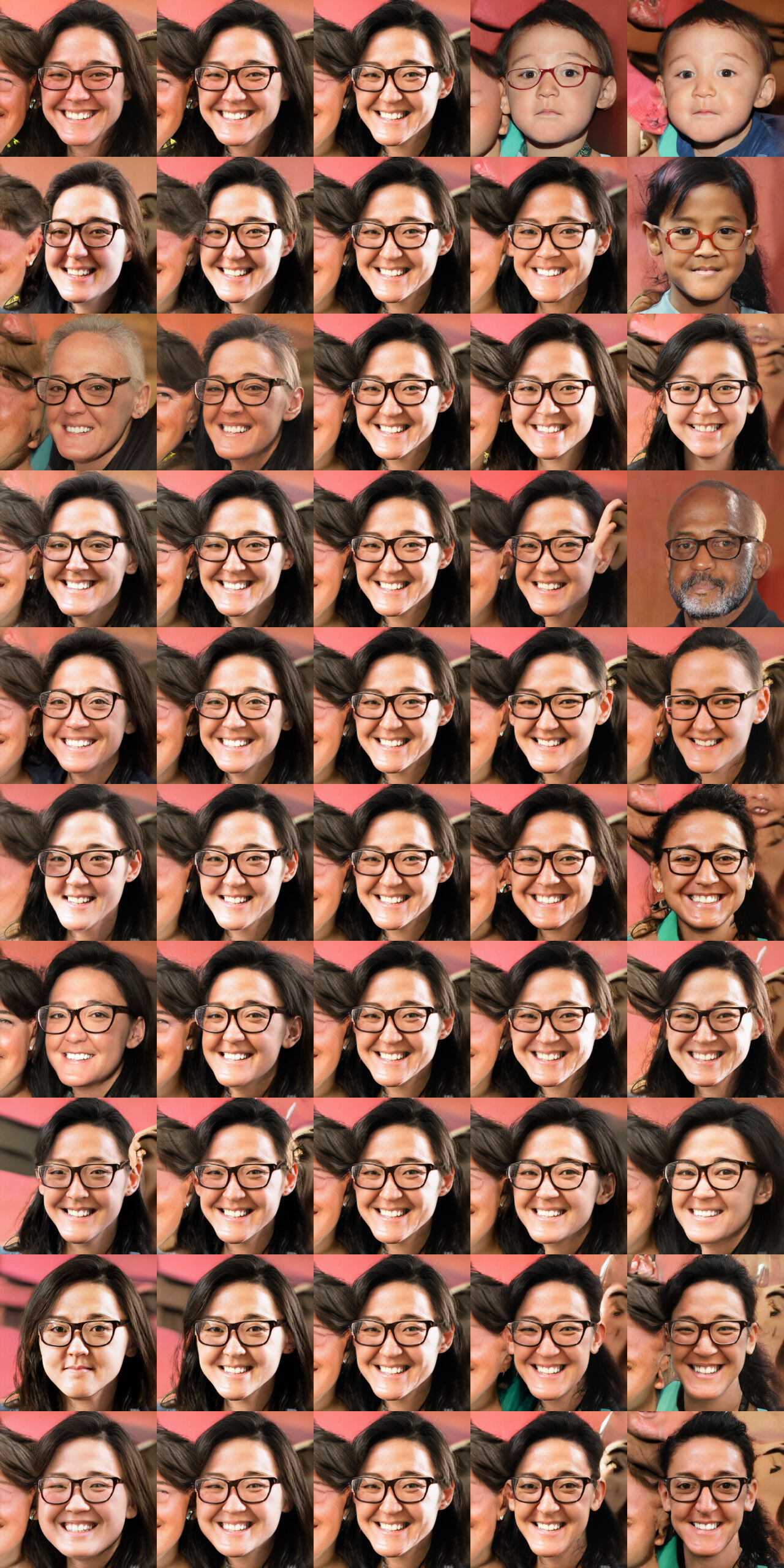}} & \begin{tabular}[c]{@{}l@{}} \\[-0.2cm] 1 \\[0.7cm] 2 \\[0.7cm] 3\end{tabular} \\
    \\[-0.32cm]
    \raisebox{-.5\height}{\adjincludegraphics[width=0.7\textwidth, trim={0 {0.7\height} 0 0}, clip]{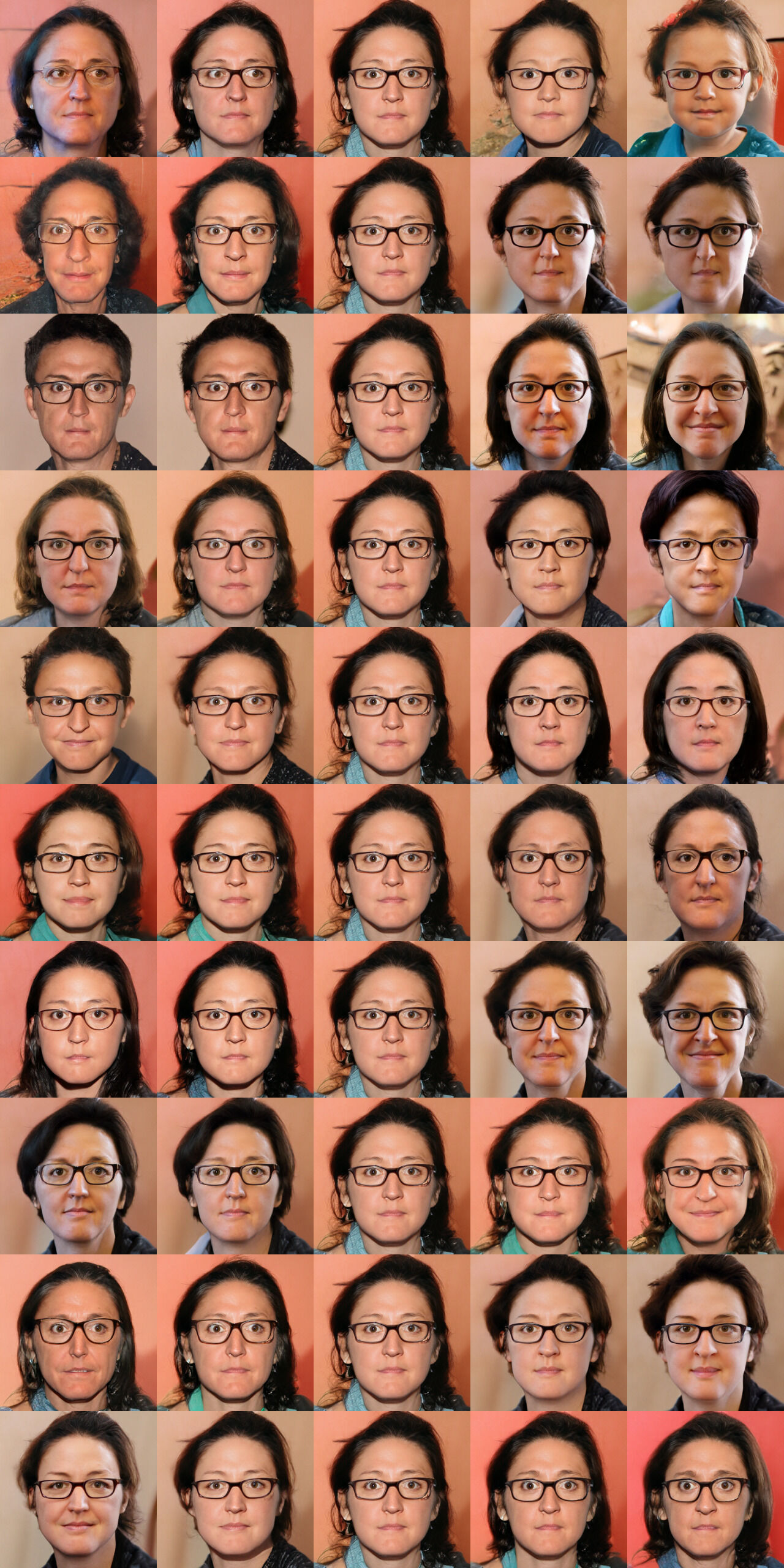}} & \begin{tabular}[c]{@{}l@{}} \\[-0.2cm] 1 \\[0.7cm] 2 \\[0.7cm] 3\end{tabular} \\
    \\[-0.32cm]
    \raisebox{-.5\height}{\adjincludegraphics[width=0.7\textwidth, trim={0 {0.7\height} 0 0}, clip]{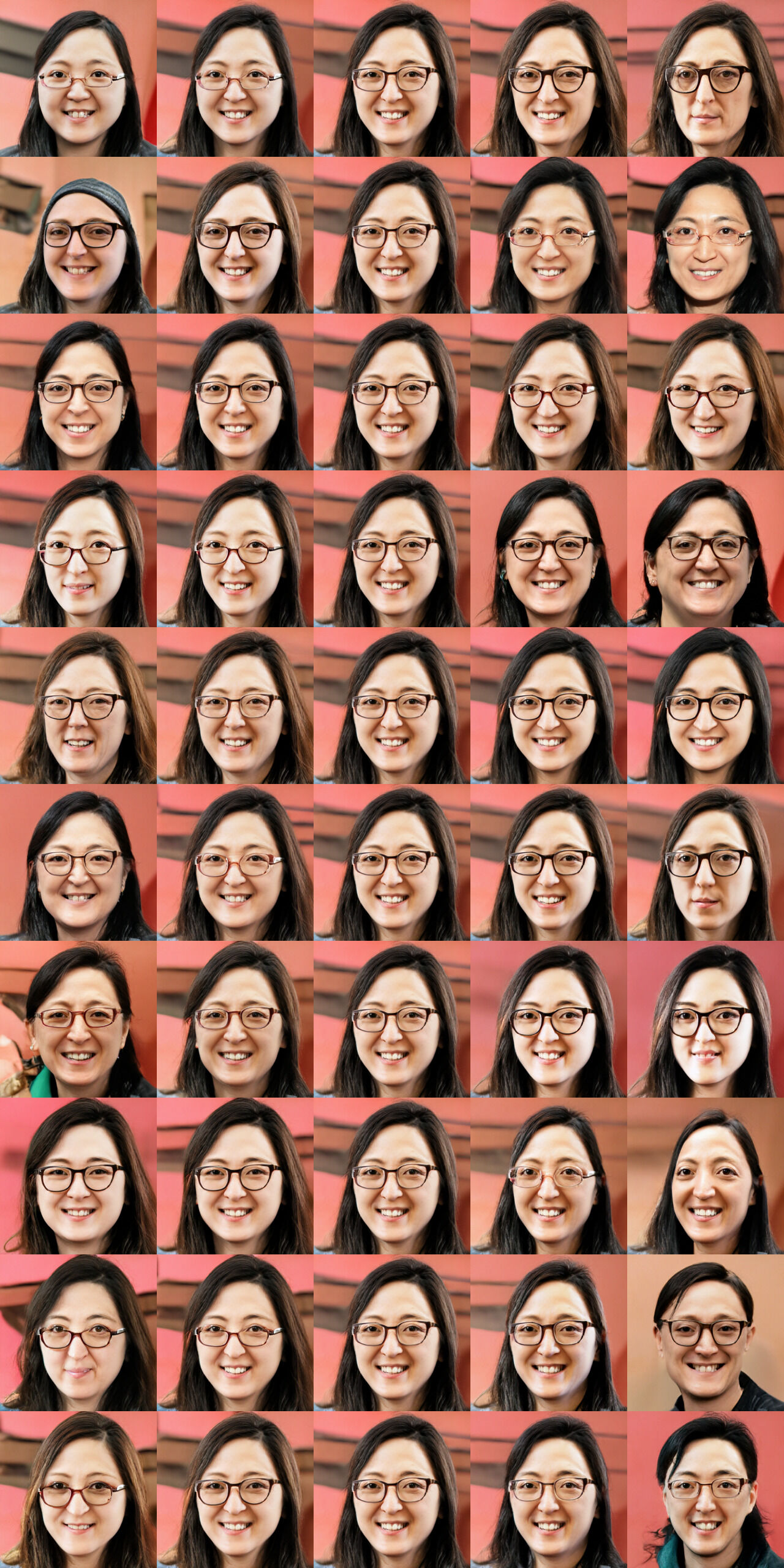}} & \begin{tabular}[c]{@{}l@{}} \\[-0.2cm] 1 \\[0.7cm] 2 \\[0.7cm] 3\end{tabular} \\
    \\[-0.32cm]
    \raisebox{-.5\height}{\adjincludegraphics[width=0.7\textwidth, trim={0 {0.7\height} 0 0}, clip]{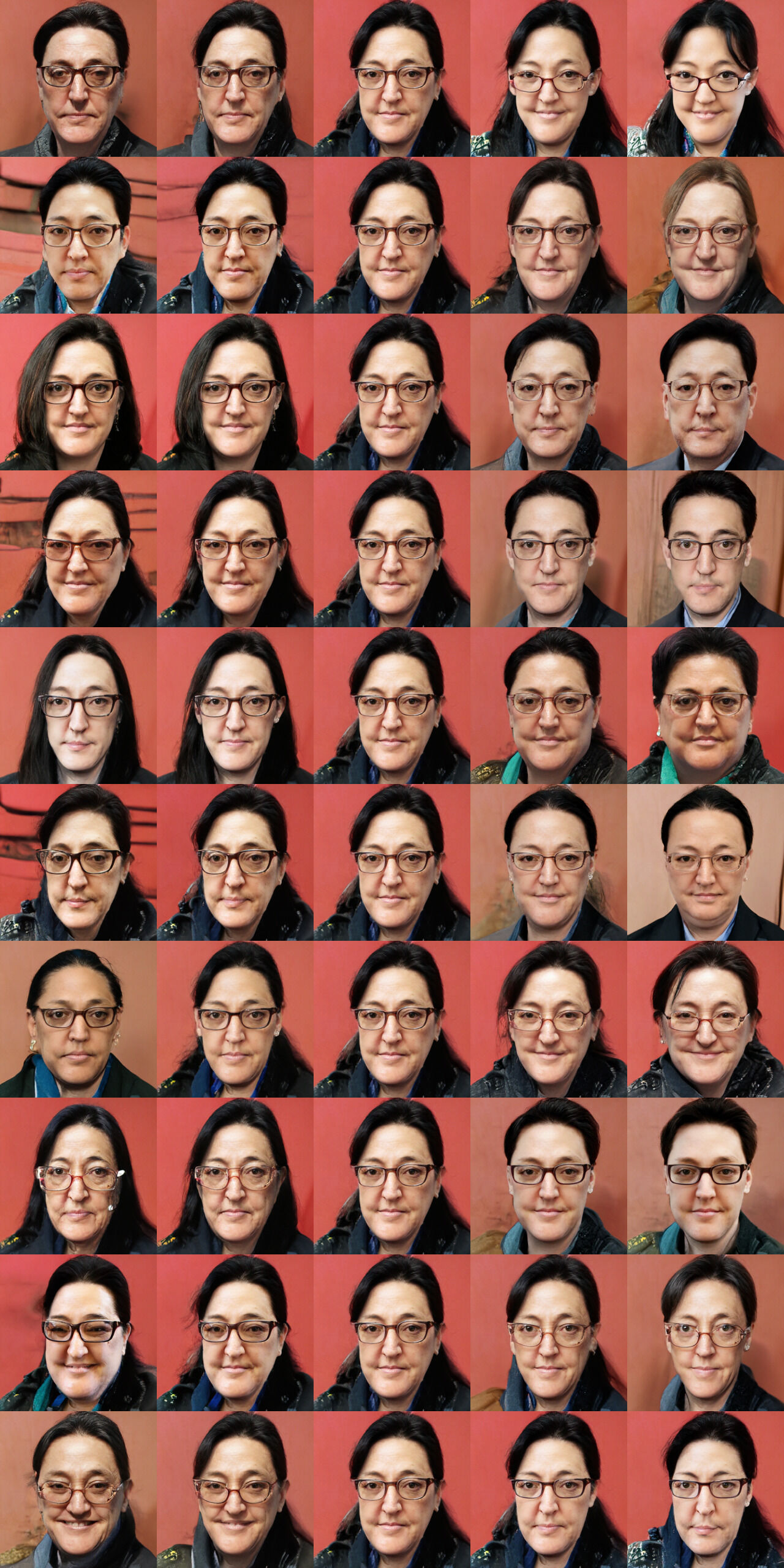}} & \begin{tabular}[c]{@{}l@{}} \\[-0.2cm] 1 \\[0.7cm] 2 \\[0.7cm] 3\end{tabular} \\
    \\[-0.32cm]
    \raisebox{-.5\height}{\adjincludegraphics[width=0.7\textwidth, trim={0 {0.7\height} 0 0}, clip]{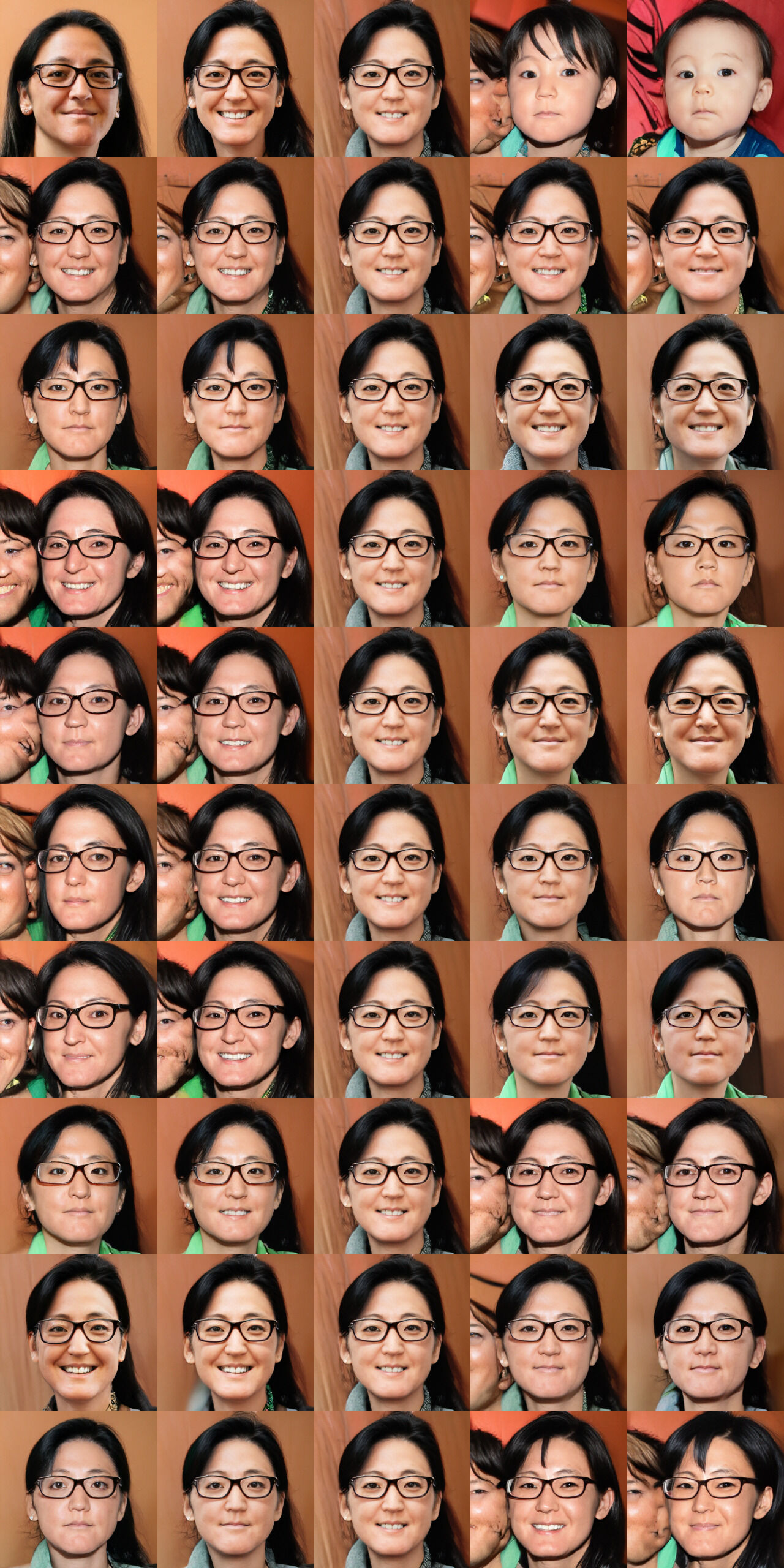}} & \begin{tabular}[c]{@{}l@{}} \\[-0.2cm] 1 \\[0.7cm] 2 \\[0.7cm] 3\end{tabular} \\
    \end{tabular}
    }
    \addtolength{\tabcolsep}{4pt}
    \caption*{\hspace{-2.3cm} (b) Identity 2}
\end{subfigure}
    \caption{Visualization of the first three principal component analysis axes for two identities using different ID vectors.}
    \label{fig:pca}
\end{figure*}

To analyze the most prominent axes within the latent space, we perform principal component analysis (PCA) on the ID vectors of all $70000$ images of the FFHQ data set for all considered face recognition models. As seen in \cref{fig:pca}, the first principal component appears to mainly encode a person's age while the subsequent components are more entangled and thus less interpretable. Note that we normalize the size of the steps along the PCA directions by the $L_2$ norm of the ID vectors to ensure a similar relative step size for the different ID vectors. Further note that large steps along any direction can cause the resulting latent vector to leave the distribution of plausible ID vectors and can cause artifacts, which is expected.

\clearpage

\begin{figure*}[htpb]
\centering
\begin{subfigure}{0.9\textwidth}
\centering
    \addtolength{\tabcolsep}{-2pt}
    \small{
    \begin{tabular}{lc}
    AdaFace~\cite{adaface} & \raisebox{-.5\height}{\adjincludegraphics[width=.85\textwidth, trim={0 {.5\height} 0 {0.25\height}}, clip]{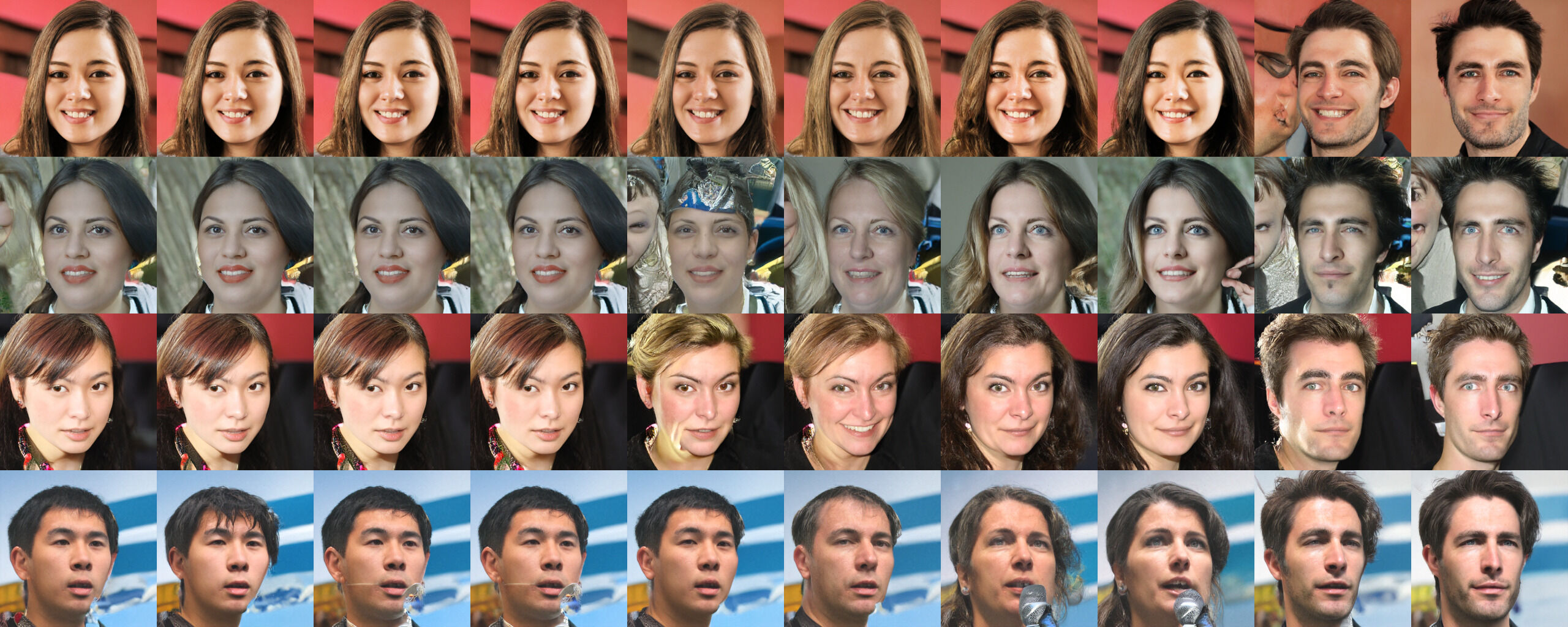}} \\
    \\[-0.32cm]
    ArcFace~\cite{arcface, gaussian_sampling} & \raisebox{-.5\height}{\adjincludegraphics[width=.85\textwidth, trim={0 {.5\height} 0 {0.25\height}}, clip]{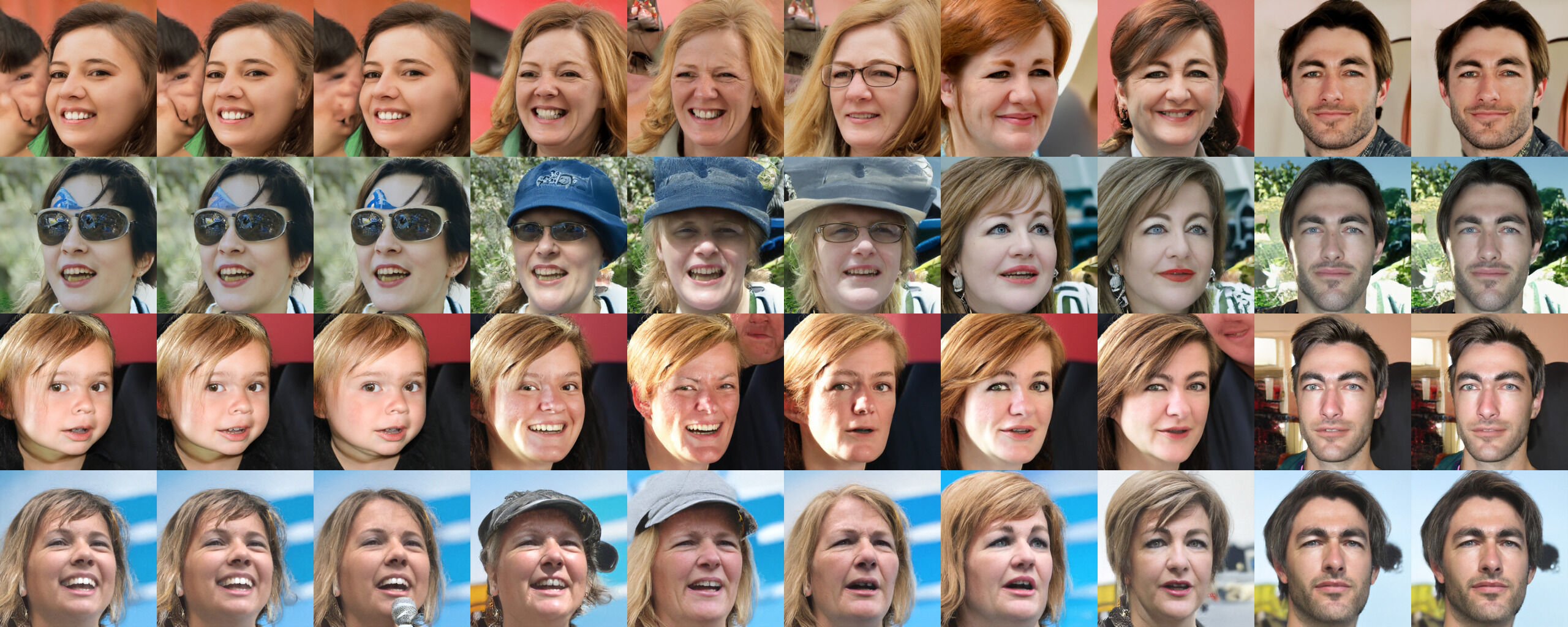}} \\
    \\[-0.32cm]
    FaceNet~\cite{facenet, facenet_pytorch} & \raisebox{-.5\height}{\adjincludegraphics[width=.85\textwidth, trim={0 {.5\height} 0 {0.25\height}}, clip]{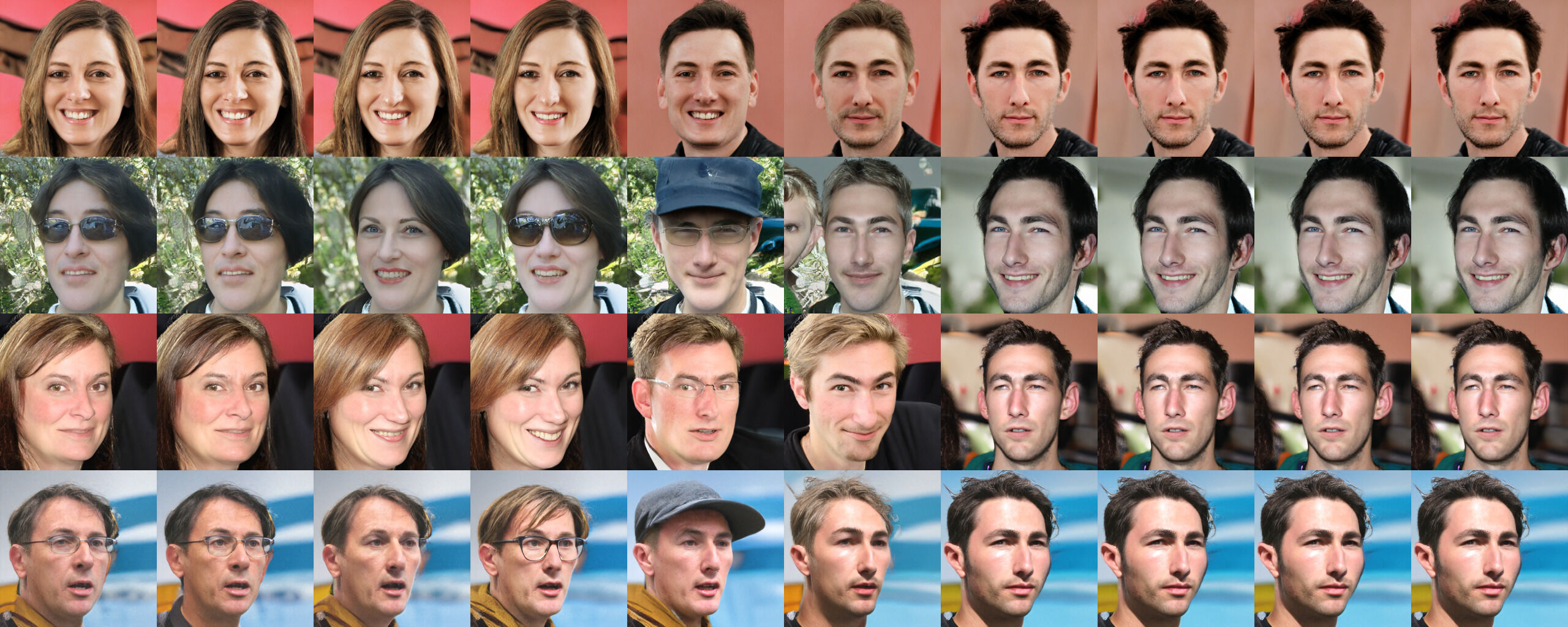}} \\
    \\[-0.32cm]
    FROM~\cite{from} & \raisebox{-.5\height}{\adjincludegraphics[width=.85\textwidth, trim={0 {.5\height} 0 {0.25\height}}, clip]{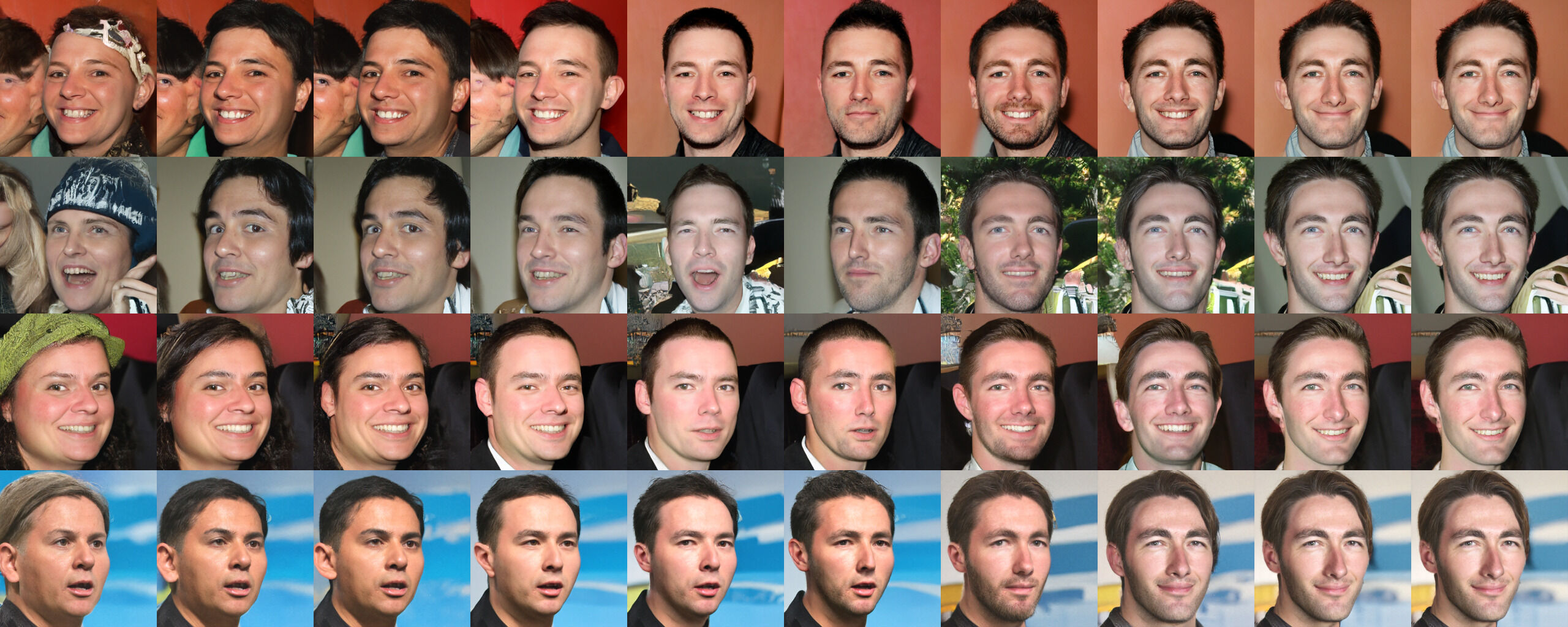}} \\
    \\[-0.32cm]
    InsightFace~\cite{insightface} & \raisebox{-.5\height}{\adjincludegraphics[width=.85\textwidth, trim={0 {.5\height} 0 {0.25\height}}, clip]{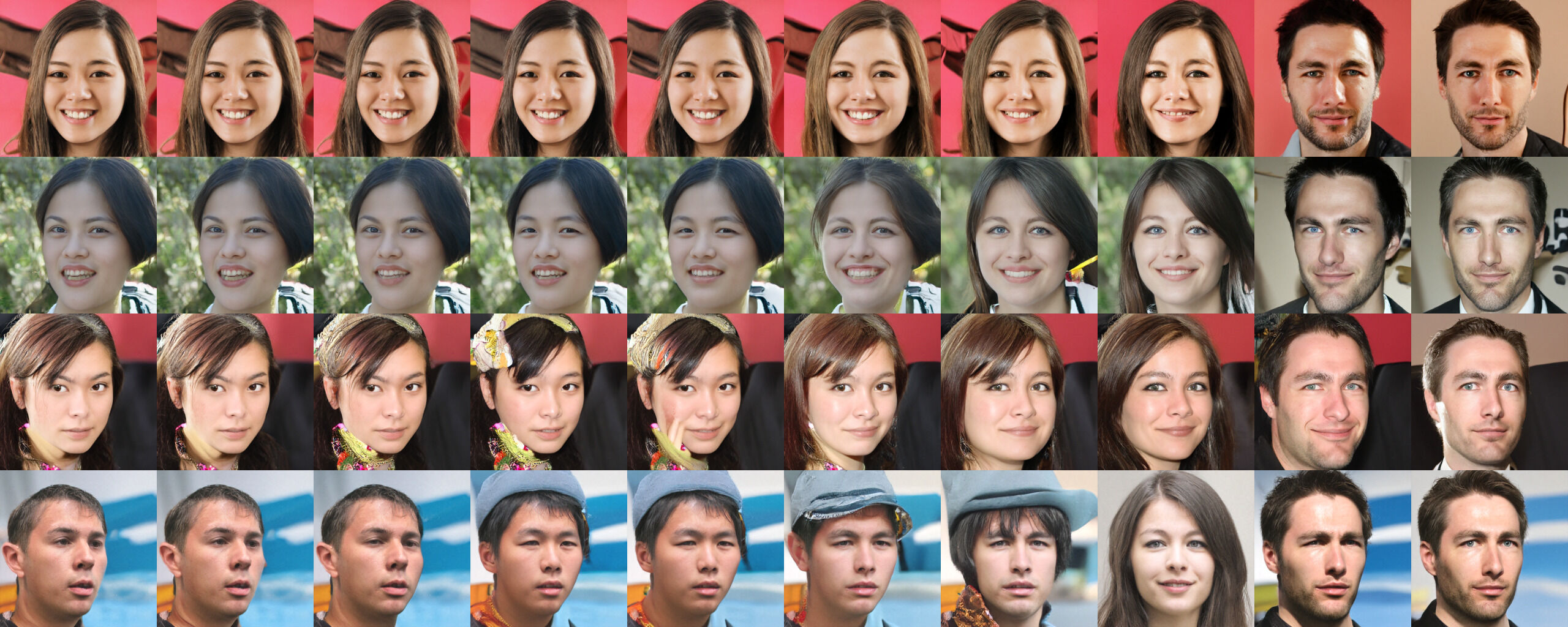}} \\
    Number of axes $\rightarrow$ & \begin{tabularx}{0.85\textwidth}{ *{10}{Y} } $1$ & $2$ & $4$ & $8$ & $16$ & $32$ & $64$ & $128$ & $256$ & $512$ \end{tabularx} \\
    \end{tabular}
    }
    \addtolength{\tabcolsep}{2pt}
    \caption{Identity 1}
    \vspace{5mm}
\end{subfigure}
\begin{subfigure}{0.9\textwidth}
\centering
    \addtolength{\tabcolsep}{-2pt}
    \small{
    \begin{tabular}{lc}
    AdaFace~\cite{adaface} & \raisebox{-.5\height}{\adjincludegraphics[width=.85\textwidth, trim={0 {.5\height} 0 {0.25\height}}, clip]{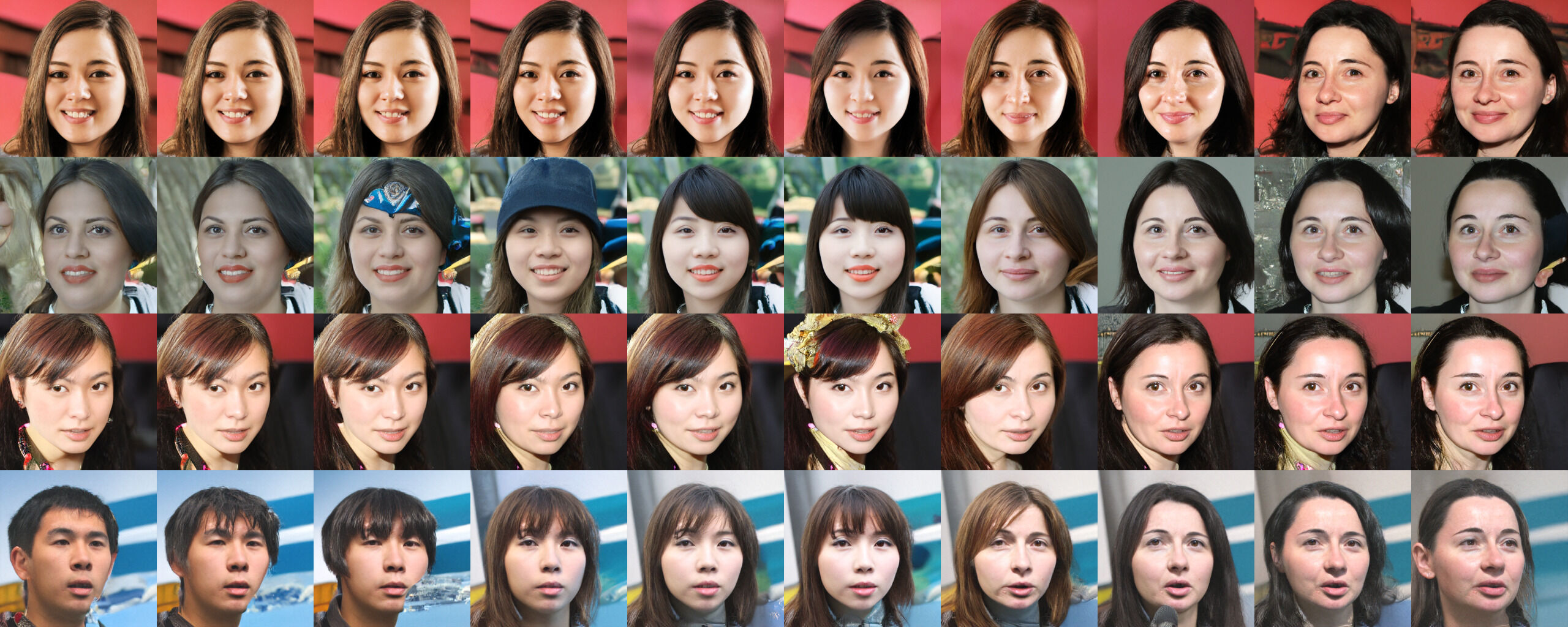}} \\
    \\[-0.32cm]
    ArcFace~\cite{arcface, gaussian_sampling} & \raisebox{-.5\height}{\adjincludegraphics[width=.85\textwidth, trim={0 {.5\height} 0 {0.25\height}}, clip]{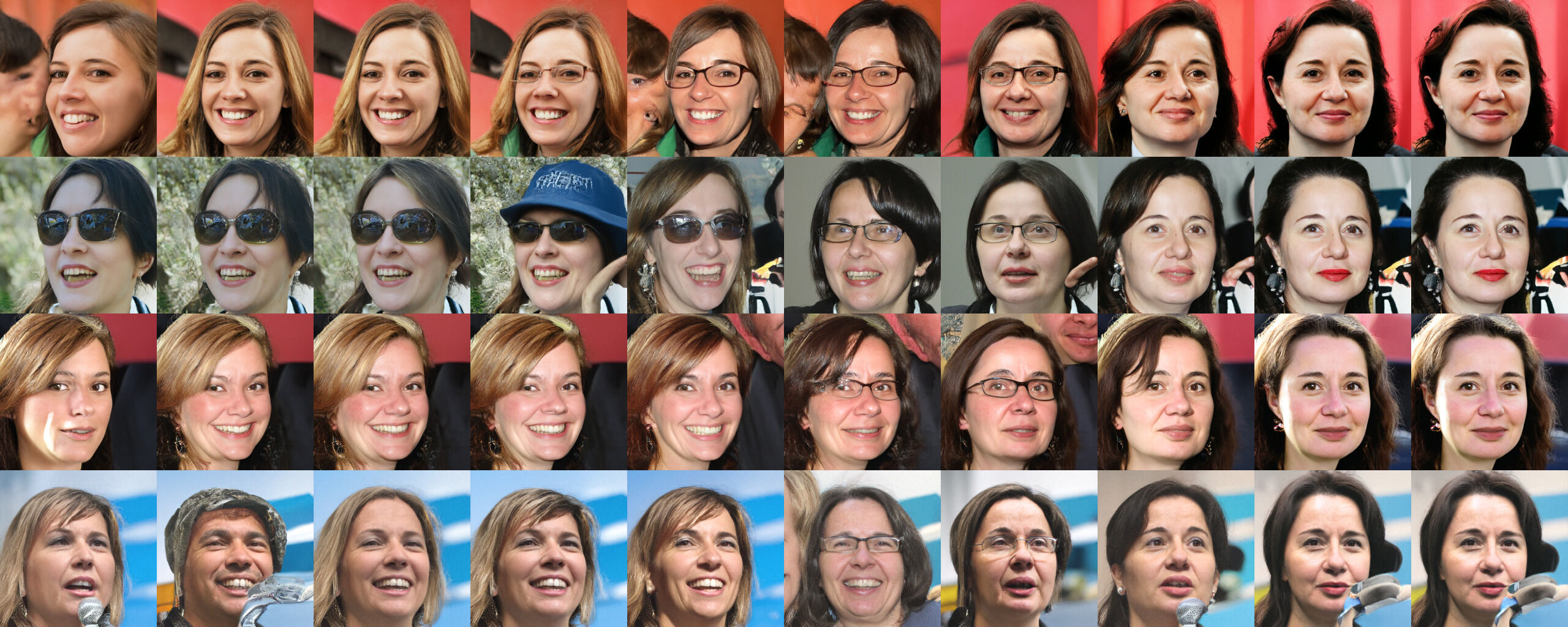}} \\
    \\[-0.32cm]
    FaceNet~\cite{facenet, facenet_pytorch} & \raisebox{-.5\height}{\adjincludegraphics[width=.85\textwidth, trim={0 {.5\height} 0 {0.25\height}}, clip]{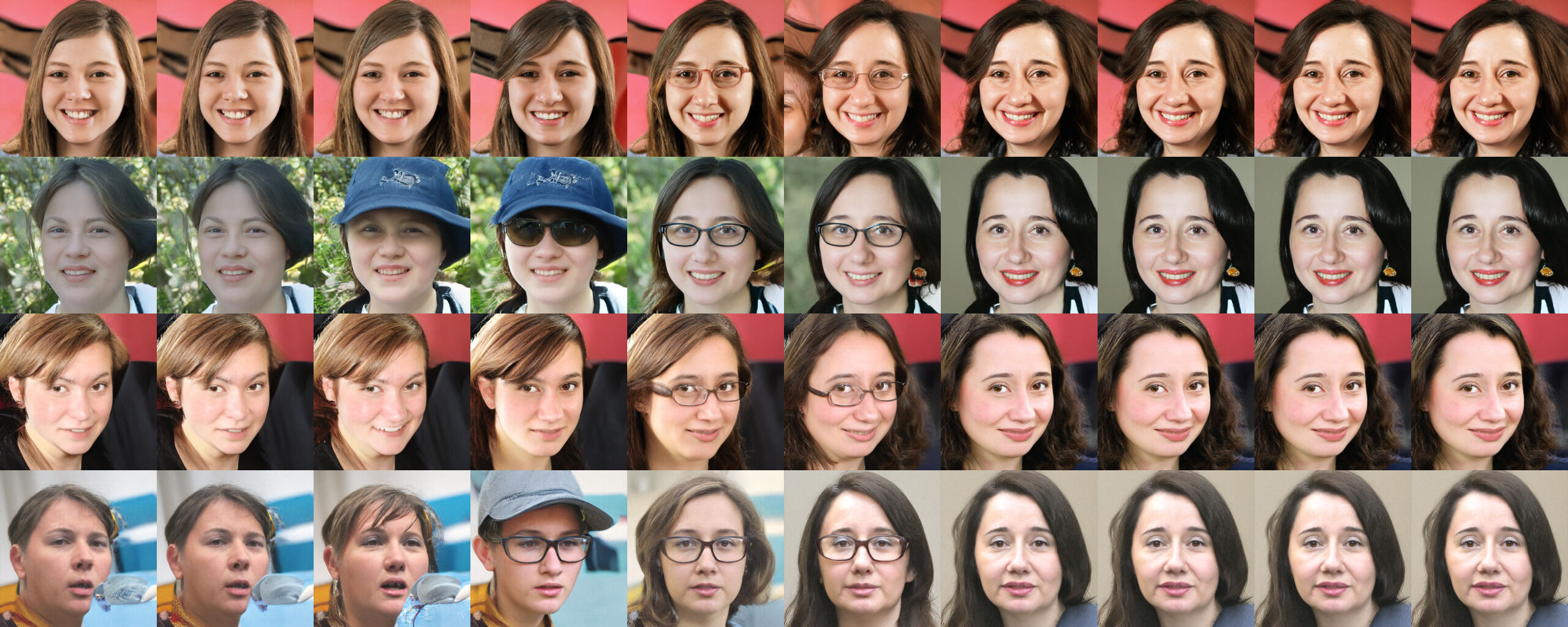}} \\
    \\[-0.32cm]
    FROM~\cite{from} & \raisebox{-.5\height}{\adjincludegraphics[width=.85\textwidth, trim={0 {.5\height} 0 {0.25\height}}, clip]{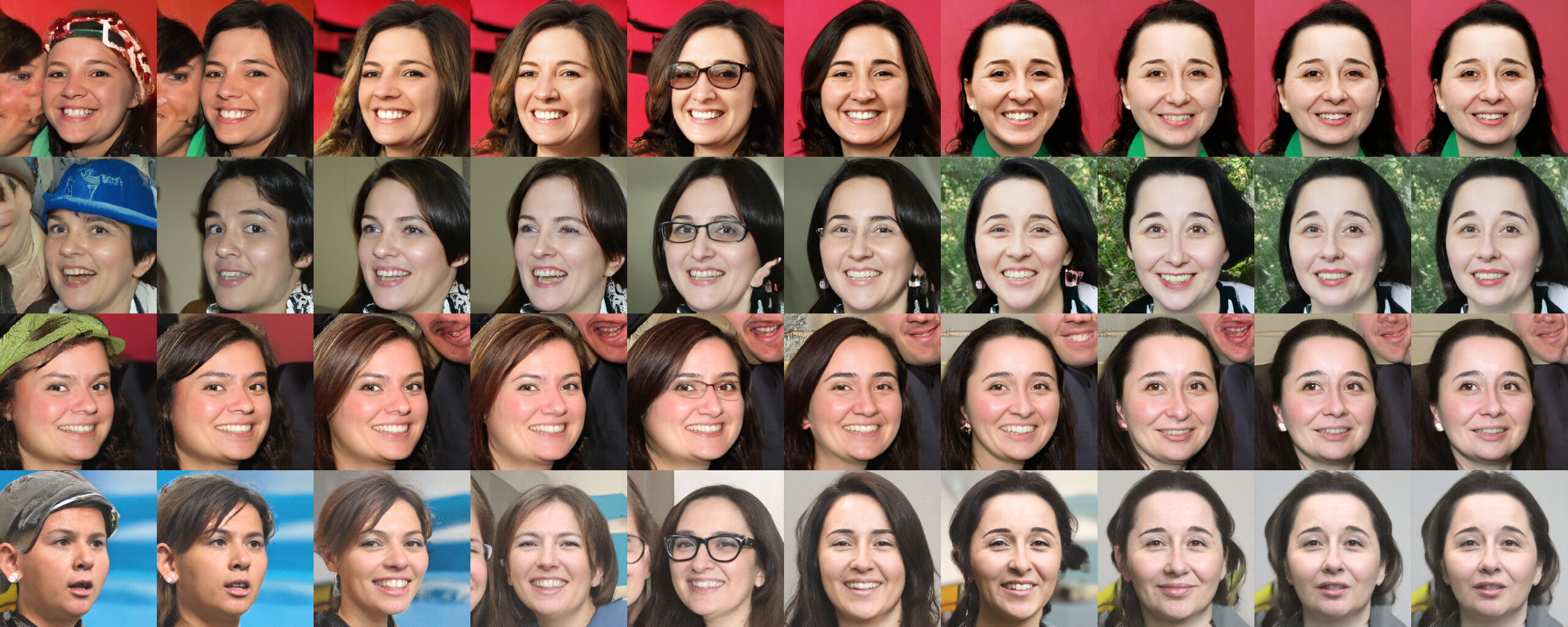}} \\
    \\[-0.32cm]
    InsightFace~\cite{insightface} & \raisebox{-.5\height}{\adjincludegraphics[width=.85\textwidth, trim={0 {.5\height} 0 {0.25\height}}, clip]{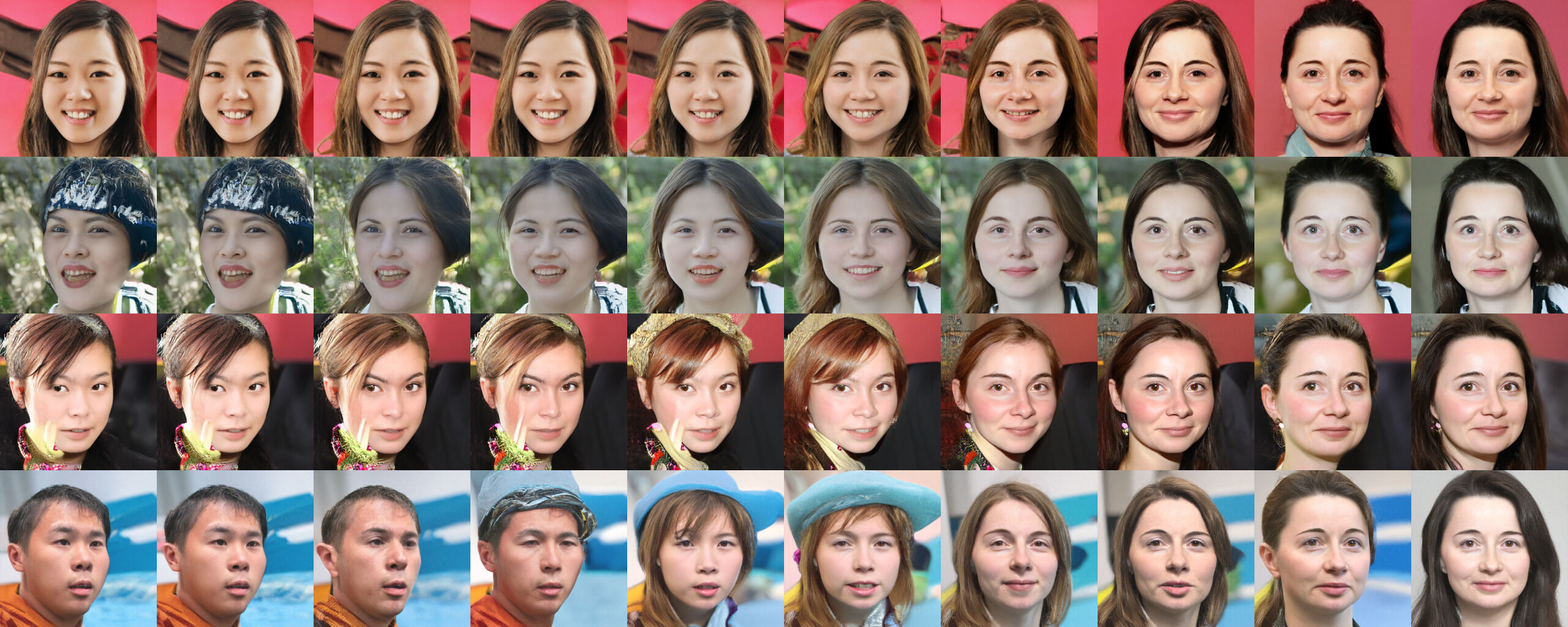}} \\
    Number of axes $\rightarrow$ & \begin{tabularx}{0.85\textwidth}{ *{10}{Y} } $1$ & $2$ & $4$ & $8$ & $16$ & $32$ & $64$ & $128$ & $256$ & $512$ \end{tabularx} \\
    \end{tabular}
    }
    \addtolength{\tabcolsep}{2pt}
    \caption{Identity 2}
\end{subfigure}
    \caption{Visualization of the projections onto the first $\{1, 2, 4, 8, 16, 32, 64, 128, 256, 512\}$ principal component analysis (PCA) axes for two identities using different ID vectors.}
    \label{fig:pca_red}
\end{figure*}

Rather than traversing along PCA directions, \cref{fig:pca_red} shows how the images change when projecting the ID vectors onto the first $\{1, 2, 4, 8, 16, 32, 64, 128, 256, 512\}$ PCA axes. The main insight from this experiment is that the ID vectors from some face recognition models, such as FaceNet~\cite{facenet, facenet_pytorch}, can be compressed to as few as 64 dimensions without changing the perceived identity while others, such as AdaFace~\cite{adaface}, require all 512 dimensions.

\clearpage

\subsection{Custom directions}

\begin{figure*}[htpb]
\centering
\begin{subfigure}{0.55\textwidth}
\centering
    \addtolength{\tabcolsep}{-4pt}
    \small{
    \begin{tabular}{lc}
    AdaFace~\cite{adaface} & \raisebox{-.5\height}{\adjincludegraphics[width=0.7\textwidth, trim={0 {0.875\height} 0 0}, clip]{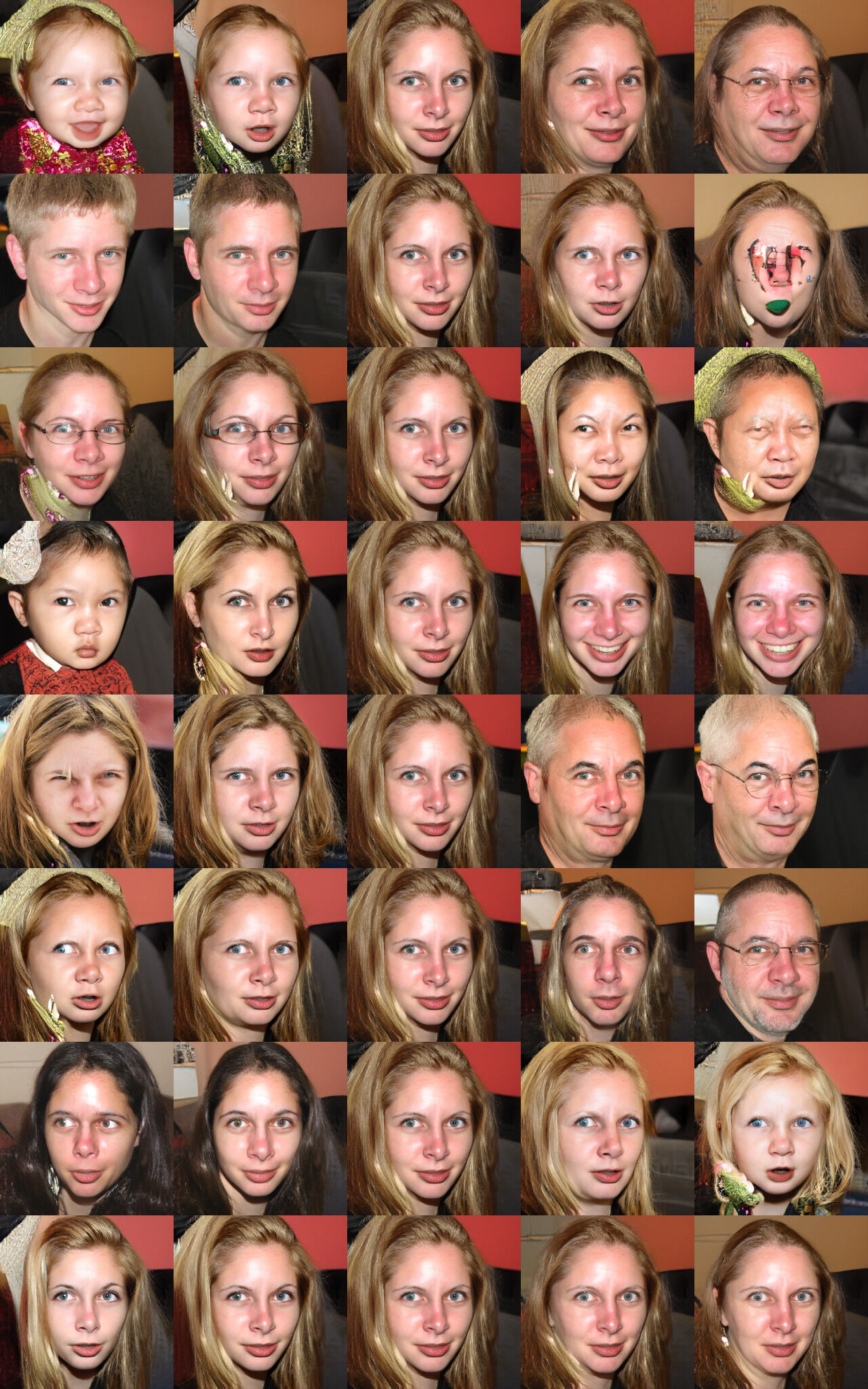}} \\
    \\[-0.35cm]
    ArcFace~\cite{arcface, gaussian_sampling} & \raisebox{-.5\height}{\adjincludegraphics[width=0.7\textwidth, trim={0 {0.875\height} 0 0}, clip]{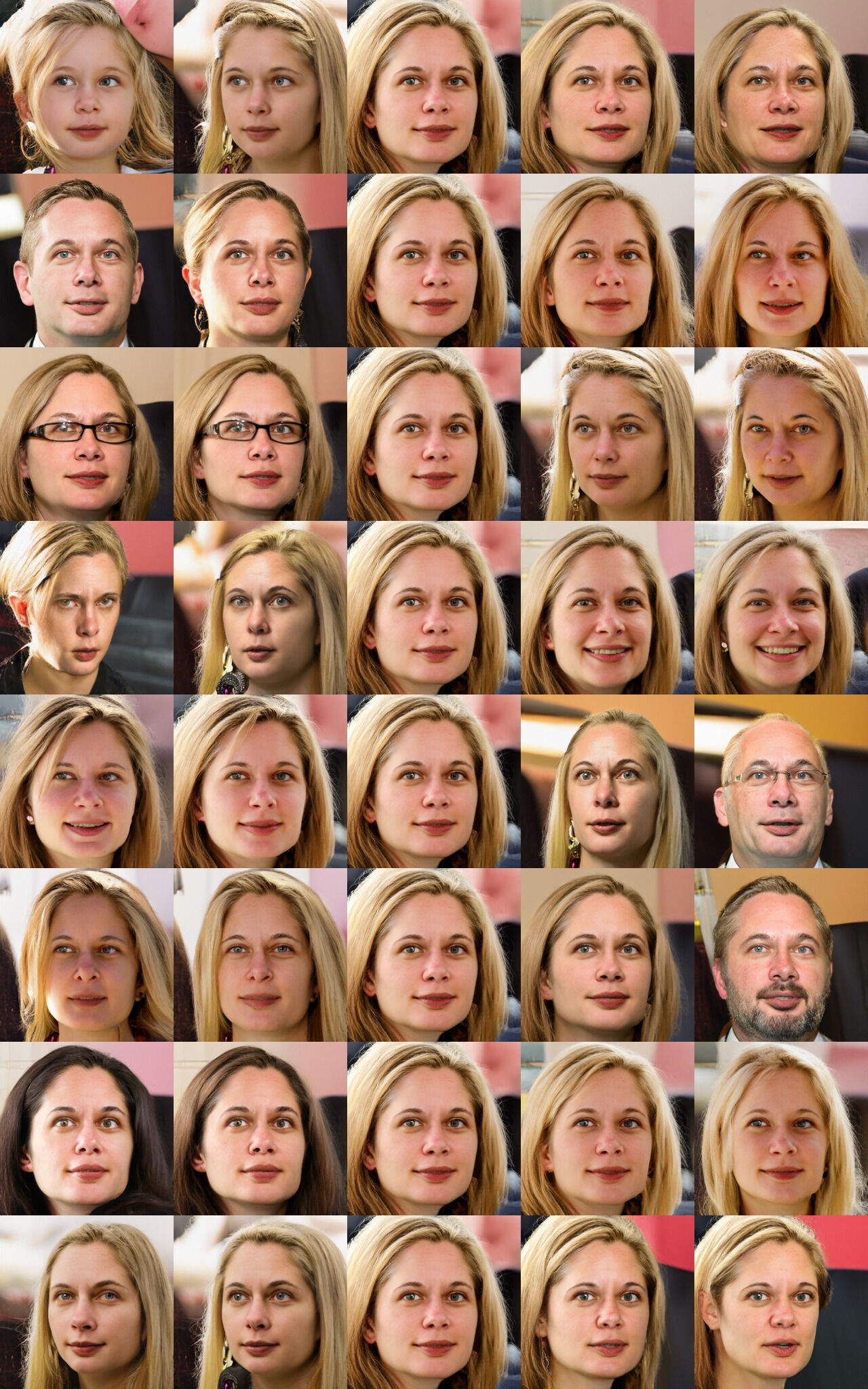}} \\
    \\[-0.35cm]
    FaceNet~\cite{facenet, facenet_pytorch} & \raisebox{-.5\height}{\adjincludegraphics[width=0.7\textwidth, trim={0 {0.875\height} 0 0}, clip]{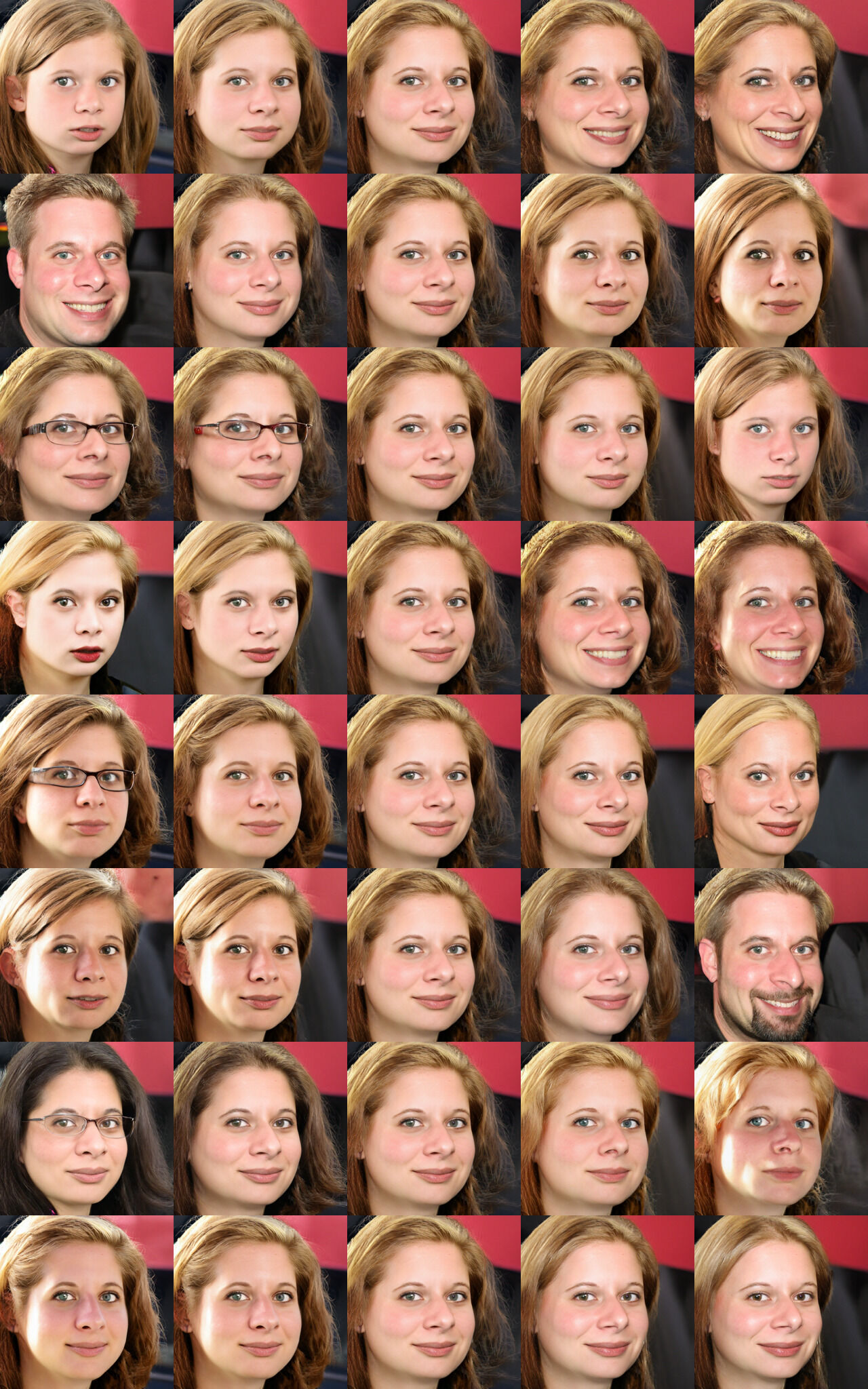}} \\
    \\[-0.35cm]
    FROM~\cite{from} & \raisebox{-.5\height}{\adjincludegraphics[width=0.7\textwidth, trim={0 {0.875\height} 0 0}, clip]{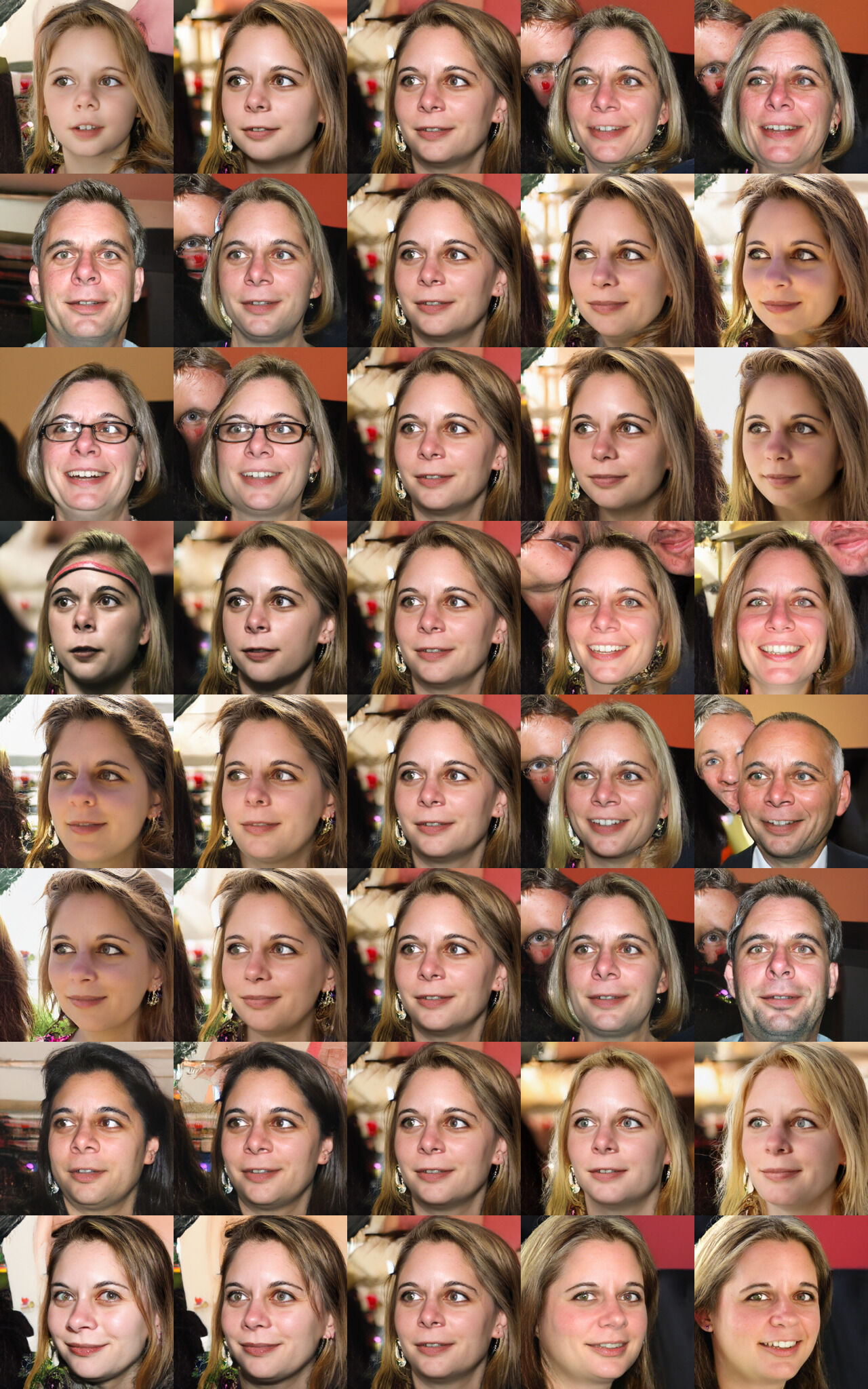}} \\
    \\[-0.35cm]
    InsightFace~\cite{insightface} \hspace{0.2cm} & \raisebox{-.5\height}{\adjincludegraphics[width=0.7\textwidth, trim={0 {0.875\height} 0 0}, clip]{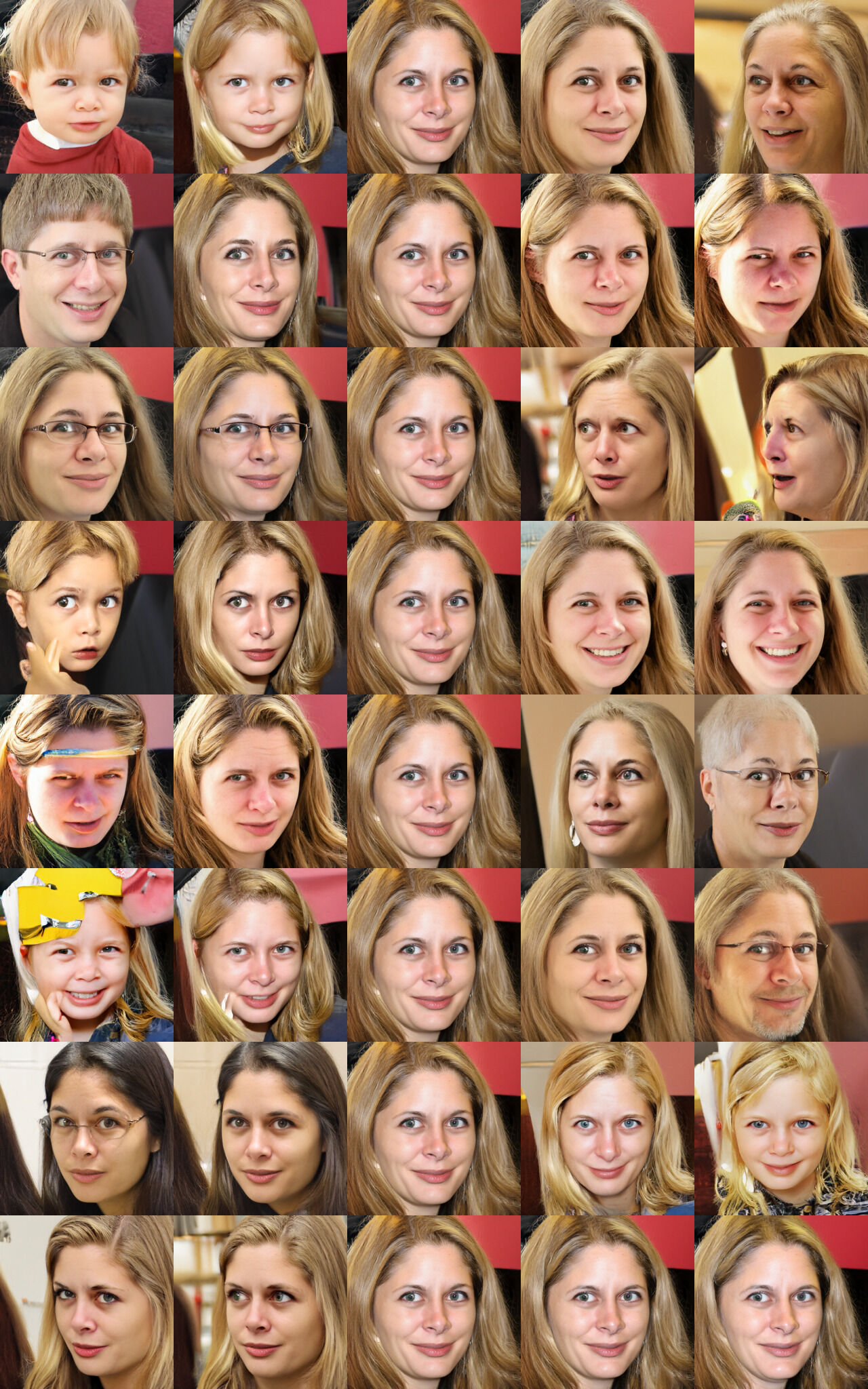}} \\
    \end{tabular}
    }
    \addtolength{\tabcolsep}{4pt}
    \caption*{\hspace{2.0cm} (a) Identity 1 $|$ Age}
    \vspace{5mm}
\end{subfigure}
\hspace{0.5cm}
\begin{subfigure}{0.385\textwidth}
\centering
    \addtolength{\tabcolsep}{-4pt}
    \small{
    \begin{tabular}{c}
    \raisebox{-.5\height}{\adjincludegraphics[width=\textwidth, trim={0 {.125\height} 0 {0.75\height}}, clip]{images/custom_dir/martina/AdaFace.jpg}} \\
    \\[-0.35cm]
    \raisebox{-.5\height}{\adjincludegraphics[width=\textwidth, trim={0 {.125\height} 0 {0.75\height}}, clip]{images/custom_dir/martina/arcface.jpg}} \\
    \\[-0.35cm]
    \raisebox{-.5\height}{\adjincludegraphics[width=\textwidth, trim={0 {.125\height} 0 {0.75\height}}, clip]{images/custom_dir/martina/facenet.jpg}} \\
    \\[-0.35cm]
    \raisebox{-.5\height}{\adjincludegraphics[width=\textwidth, trim={0 {.125\height} 0 {0.75\height}}, clip]{images/custom_dir/martina/FROM.jpg}} \\
    \\[-0.35cm]
    \raisebox{-.5\height}{\adjincludegraphics[width=\textwidth, trim={0 {.125\height} 0 {0.75\height}}, clip]{images/custom_dir/martina/insightface.jpg}} \\
    \end{tabular}
    }
    \addtolength{\tabcolsep}{4pt}
    \caption*{(b) Identity 1 $|$ Blond hair color}
    \vspace{5mm}
\end{subfigure}
\begin{subfigure}{0.55\textwidth}
\centering
    \addtolength{\tabcolsep}{-4pt}
    \small{
    \begin{tabular}{lc}
    AdaFace~\cite{adaface} & \raisebox{-.5\height}{\adjincludegraphics[width=0.7\textwidth, trim={0 {0.875\height} 0 0}, clip]{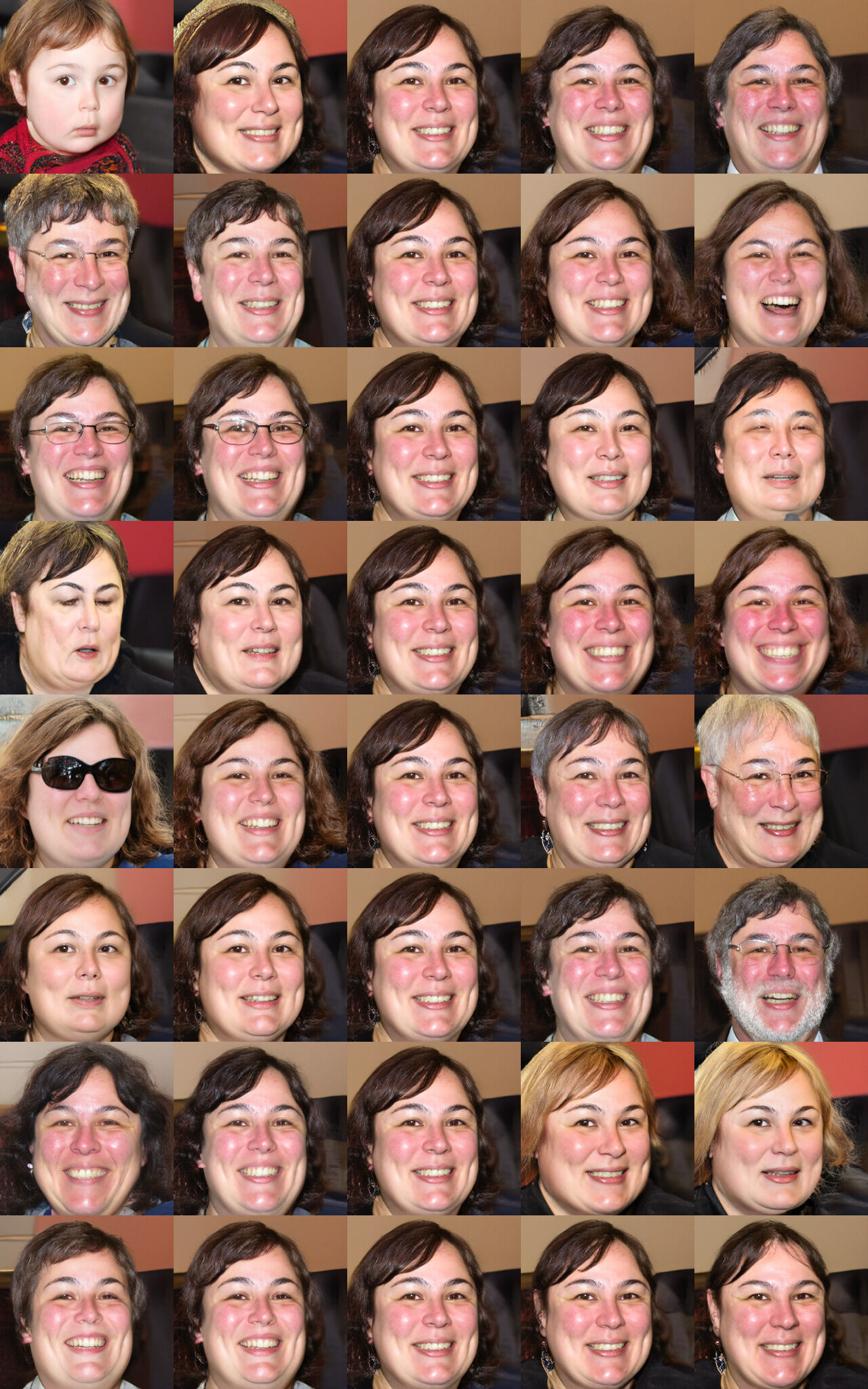}} \\
    \\[-0.35cm]
    ArcFace~\cite{arcface, gaussian_sampling} & \raisebox{-.5\height}{\adjincludegraphics[width=0.7\textwidth, trim={0 {0.875\height} 0 0}, clip]{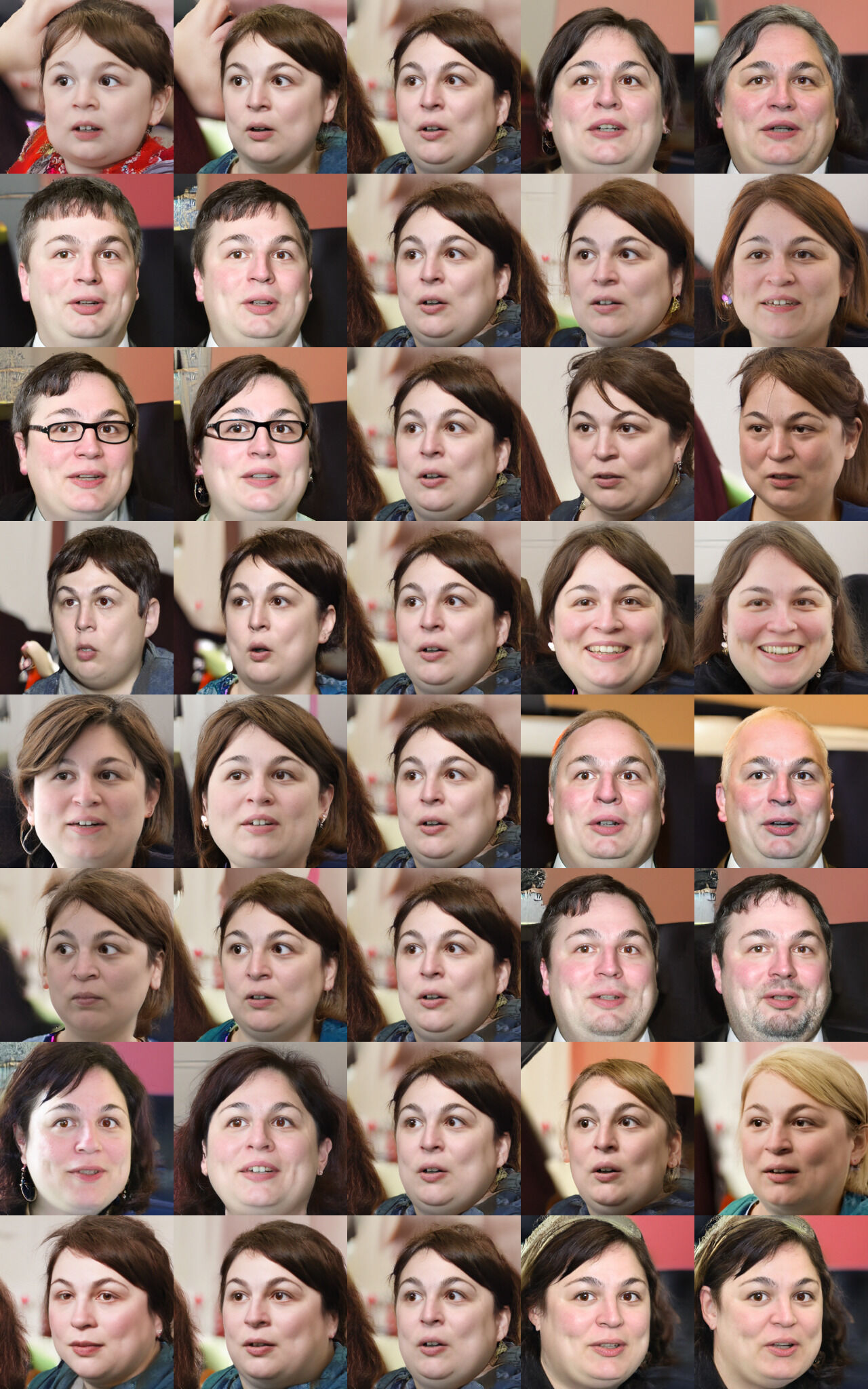}} \\
    \\[-0.35cm]
    FaceNet~\cite{facenet, facenet_pytorch} & \raisebox{-.5\height}{\adjincludegraphics[width=0.7\textwidth, trim={0 {0.875\height} 0 0}, clip]{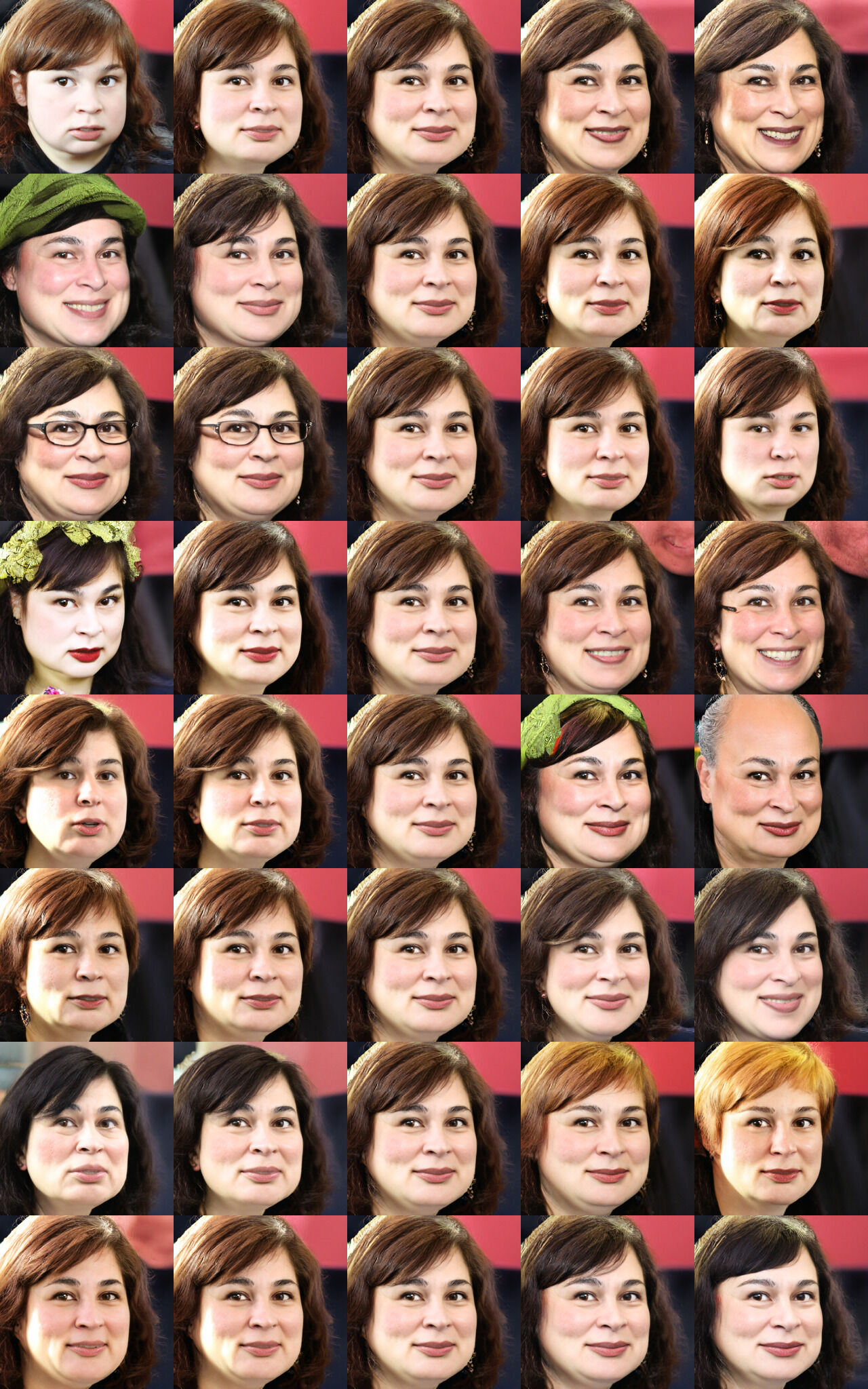}} \\
    \\[-0.35cm]
    FROM~\cite{from} & \raisebox{-.5\height}{\adjincludegraphics[width=0.7\textwidth, trim={0 {0.875\height} 0 0}, clip]{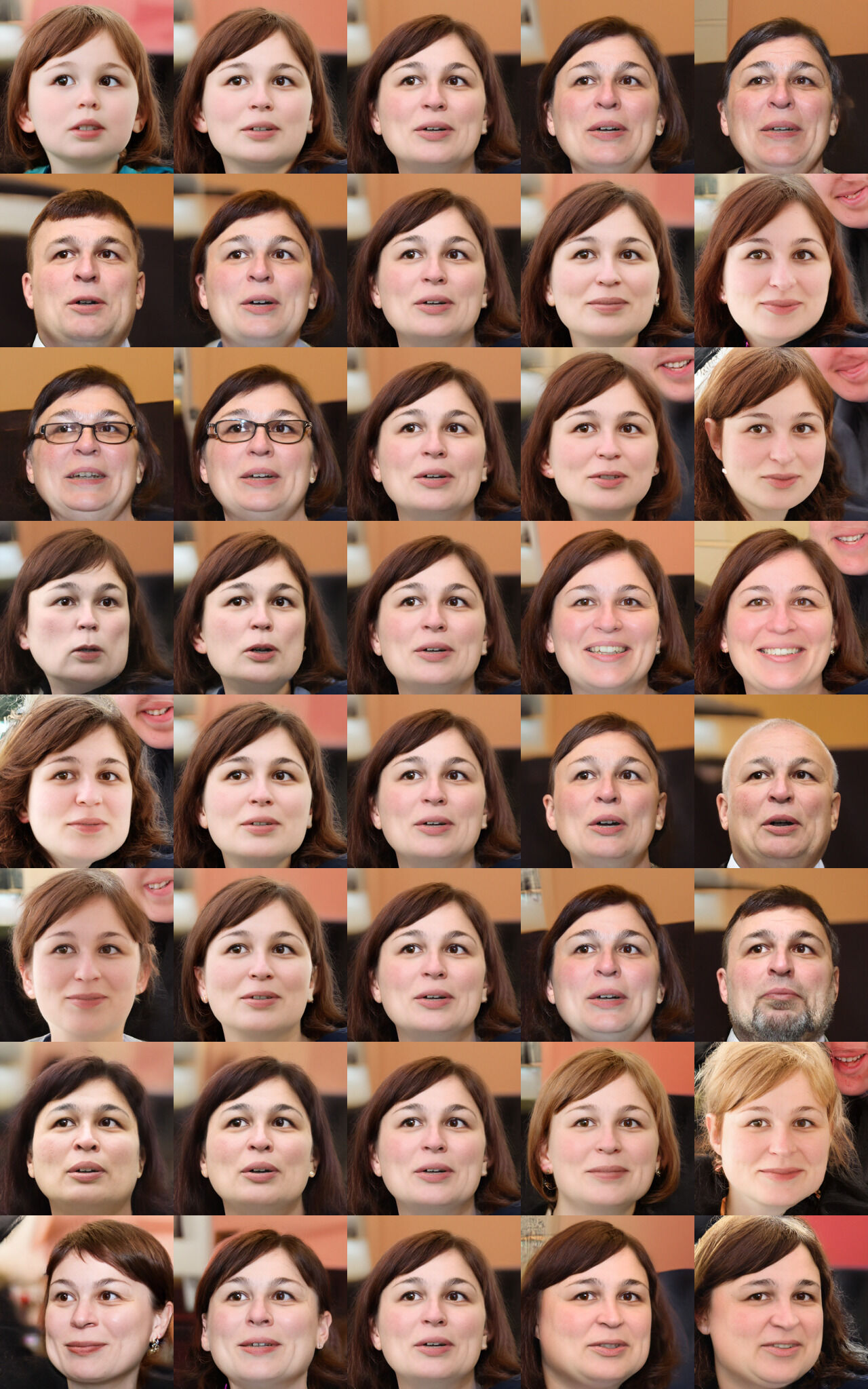}} \\
    \\[-0.35cm]
    InsightFace~\cite{insightface} \hspace{0.2cm} & \raisebox{-.5\height}{\adjincludegraphics[width=0.7\textwidth, trim={0 {0.875\height} 0 0}, clip]{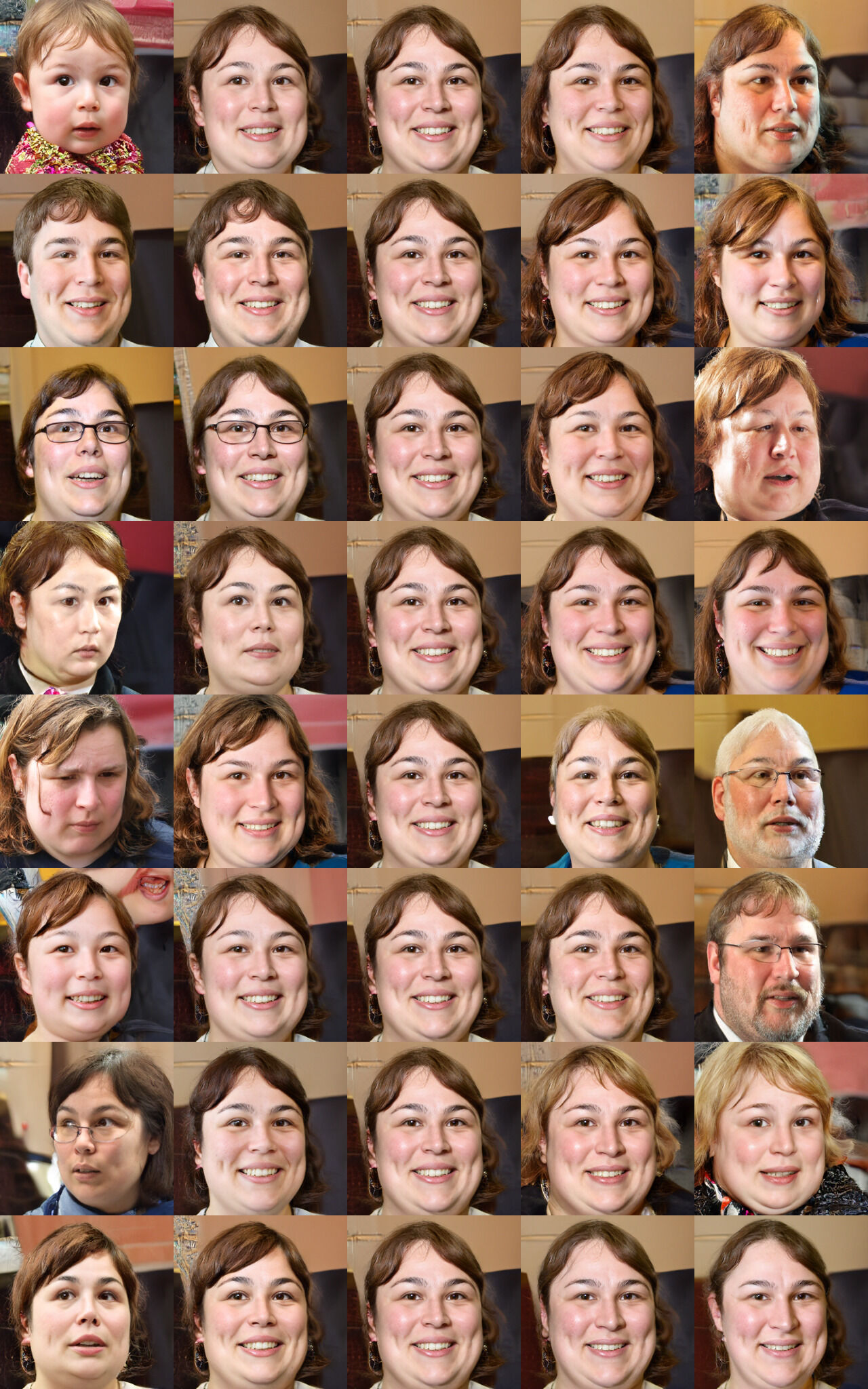}} \\
    \end{tabular}
    }
    \addtolength{\tabcolsep}{4pt}
    \caption*{\hspace{2.0cm} (c) Identity 2 $|$ Age}
\end{subfigure}
\hspace{0.5cm}
\begin{subfigure}{0.385\textwidth}
\centering
    \addtolength{\tabcolsep}{-4pt}
    \small{
    \begin{tabular}{c}
    \raisebox{-.5\height}{\adjincludegraphics[width=\textwidth, trim={0 {.125\height} 0 {0.75\height}}, clip]{images/custom_dir/sandra/AdaFace.jpg}} \\
    \\[-0.35cm]
    \raisebox{-.5\height}{\adjincludegraphics[width=\textwidth, trim={0 {.125\height} 0 {0.75\height}}, clip]{images/custom_dir/sandra/arcface.jpg}} \\
    \\[-0.35cm]
    \raisebox{-.5\height}{\adjincludegraphics[width=\textwidth, trim={0 {.125\height} 0 {0.75\height}}, clip]{images/custom_dir/sandra/facenet.jpg}} \\
    \\[-0.35cm]
    \raisebox{-.5\height}{\adjincludegraphics[width=\textwidth, trim={0 {.125\height} 0 {0.75\height}}, clip]{images/custom_dir/sandra/FROM.jpg}} \\
    \\[-0.35cm]
    \raisebox{-.5\height}{\adjincludegraphics[width=\textwidth, trim={0 {.125\height} 0 {0.75\height}}, clip]{images/custom_dir/sandra/insightface.jpg}} \\
    \end{tabular}
    }
    \addtolength{\tabcolsep}{4pt}
    \caption*{(d) Identity 2 $|$ Blond hair color}
\end{subfigure}
    \caption{Visualization of custom direction modifications for two identities using different ID vectors for two directions that can be considered to belong to a person's identity.}
    \label{fig:custom_dir_identity}
\end{figure*}

Since the PCA axes are difficult to interpret, we calculate custom directions for each face recognition model as described in the main paper. As the biases of the FFHQ data set used to train our inversion models are the same for all ID vectors (\eg glasses appearing when increasing the age direction), the presence or absence of certain directions in the latent space along which a given feature can be changed give insights about what information is extracted by a given face recognition model. As seen in the main paper and \cref{fig:custom_dir_identity}, directions that are expected to be contained in the ID vector, such as the age and the gender, can be traversed smoothly. Furthermore, directions that may or may not be considered as part of the identity, such as the current look of a person (\eg glasses, hair style, facial hair style), are also commonly contained as seen in the examples with the blond hair color in \cref{fig:custom_dir_identity}.

\clearpage

\begin{figure*}[htpb]
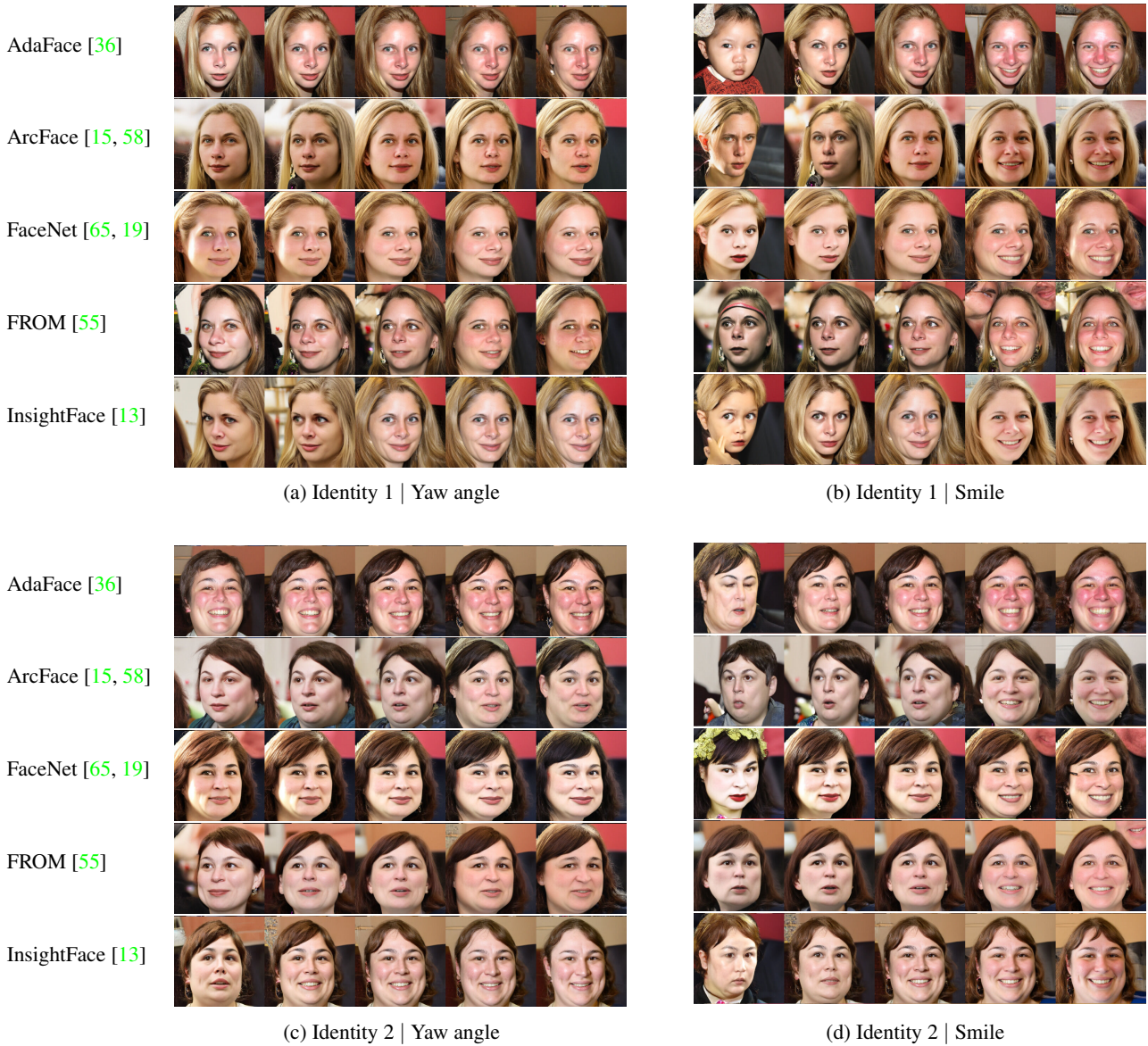

\centering
\begin{subfigure}{0.55\textwidth}
\centering
    \addtolength{\tabcolsep}{-4pt}
    \small{
    \begin{tabular}{lc}
    AdaFace~\cite{adaface} & \raisebox{-.5\height}{\adjincludegraphics[width=0.7\textwidth, trim={0 0 0 {0.875\height}}, clip]{images/custom_dir/martina/AdaFace.jpg}} \\
    \\[-0.35cm]
    ArcFace~\cite{arcface, gaussian_sampling} & \raisebox{-.5\height}{\adjincludegraphics[width=0.7\textwidth, trim={0 0 0 {0.875\height}}, clip]{images/custom_dir/martina/arcface.jpg}} \\
    \\[-0.35cm]
    FaceNet~\cite{facenet, facenet_pytorch} & \raisebox{-.5\height}{\adjincludegraphics[width=0.7\textwidth, trim={0 0 0 {0.875\height}}, clip]{images/custom_dir/martina/facenet.jpg}} \\
    \\[-0.35cm]
    FROM~\cite{from} & \raisebox{-.5\height}{\adjincludegraphics[width=0.7\textwidth, trim={0 0 0 {0.875\height}}, clip]{images/custom_dir/martina/FROM.jpg}} \\
    \\[-0.35cm]
    InsightFace~\cite{insightface} \hspace{0.2cm} & \raisebox{-.5\height}{\adjincludegraphics[width=0.7\textwidth, trim={0 0 0 {0.875\height}}, clip]{images/custom_dir/martina/insightface.jpg}} \\
    \end{tabular}
    }
    \addtolength{\tabcolsep}{4pt}
    \caption*{\hspace{2.0cm} (a) Identity 1 $|$ Yaw angle}
    \vspace{5mm}
\end{subfigure}
\hspace{0.5cm}
\begin{subfigure}{0.385\textwidth}
\centering
    \addtolength{\tabcolsep}{-4pt}
    \small{
    \begin{tabular}{c}
    \raisebox{-.5\height}{\adjincludegraphics[width=\textwidth, trim={0 {.5\height} 0 {0.375\height}}, clip]{images/custom_dir/martina/AdaFace.jpg}} \\
    \\[-0.35cm]
    \raisebox{-.5\height}{\adjincludegraphics[width=\textwidth, trim={0 {.5\height} 0 {0.375\height}}, clip]{images/custom_dir/martina/arcface.jpg}} \\
    \\[-0.35cm]
    \raisebox{-.5\height}{\adjincludegraphics[width=\textwidth, trim={0 {.5\height} 0 {0.375\height}}, clip]{images/custom_dir/martina/facenet.jpg}} \\
    \\[-0.35cm]
    \raisebox{-.5\height}{\adjincludegraphics[width=\textwidth, trim={0 {.5\height} 0 {0.375\height}}, clip]{images/custom_dir/martina/FROM.jpg}} \\
    \\[-0.35cm]
    \raisebox{-.5\height}{\adjincludegraphics[width=\textwidth, trim={0 {.5\height} 0 {0.375\height}}, clip]{images/custom_dir/martina/insightface.jpg}} \\
    \end{tabular}
    }
    \addtolength{\tabcolsep}{4pt}
    \caption*{(b) Identity 1 $|$ Smile}
    \vspace{5mm}
\end{subfigure}
\begin{subfigure}{0.55\textwidth}
\centering
    \addtolength{\tabcolsep}{-4pt}
    \small{
    \begin{tabular}{lc}
    AdaFace~\cite{adaface} & \raisebox{-.5\height}{\adjincludegraphics[width=0.7\textwidth, trim={0 0 0 {0.875\height}}, clip]{images/custom_dir/sandra/AdaFace.jpg}} \\
    \\[-0.35cm]
    ArcFace~\cite{arcface, gaussian_sampling} & \raisebox{-.5\height}{\adjincludegraphics[width=0.7\textwidth, trim={0 0 0 {0.875\height}}, clip]{images/custom_dir/sandra/arcface.jpg}} \\
    \\[-0.35cm]
    FaceNet~\cite{facenet, facenet_pytorch} & \raisebox{-.5\height}{\adjincludegraphics[width=0.7\textwidth, trim={0 0 0 {0.875\height}}, clip]{images/custom_dir/sandra/facenet.jpg}} \\
    \\[-0.35cm]
    FROM~\cite{from} & \raisebox{-.5\height}{\adjincludegraphics[width=0.7\textwidth, trim={0 0 0 {0.875\height}}, clip]{images/custom_dir/sandra/FROM.jpg}} \\
    \\[-0.35cm]
    InsightFace~\cite{insightface} \hspace{0.2cm} & \raisebox{-.5\height}{\adjincludegraphics[width=0.7\textwidth, trim={0 0 0 {0.875\height}}, clip]{images/custom_dir/sandra/insightface.jpg}} \\
    \end{tabular}
    }
    \addtolength{\tabcolsep}{4pt}
    \caption*{\hspace{2.0cm} (c) Identity 2 $|$ Yaw angle}
\end{subfigure}
\hspace{0.5cm}
\begin{subfigure}{0.385\textwidth}
\centering
    \addtolength{\tabcolsep}{-4pt}
    \small{
    \begin{tabular}{lc}
    \raisebox{-.5\height}{\adjincludegraphics[width=\textwidth, trim={0 {.5\height} 0 {0.375\height}}, clip]{images/custom_dir/sandra/AdaFace.jpg}} \\
    \\[-0.35cm]
    \raisebox{-.5\height}{\adjincludegraphics[width=\textwidth, trim={0 {.5\height} 0 {0.375\height}}, clip]{images/custom_dir/sandra/arcface.jpg}} \\
    \\[-0.35cm]
    \raisebox{-.5\height}{\adjincludegraphics[width=\textwidth, trim={0 {.5\height} 0 {0.375\height}}, clip]{images/custom_dir/sandra/facenet.jpg}} \\
    \\[-0.35cm]
    \raisebox{-.5\height}{\adjincludegraphics[width=\textwidth, trim={0 {.5\height} 0 {0.375\height}}, clip]{images/custom_dir/sandra/FROM.jpg}} \\
    \\[-0.35cm]
    \raisebox{-.5\height}{\adjincludegraphics[width=\textwidth, trim={0 {.5\height} 0 {0.375\height}}, clip]{images/custom_dir/sandra/insightface.jpg}} \\
    \end{tabular}
    }
    \addtolength{\tabcolsep}{4pt}
    \caption*{(d) Identity 2 $|$ Smile}
\end{subfigure}
    \caption{Visualization of custom direction modifications for two identities using different ID vectors for two directions that arguably do not belong to a person's identity. Our method reveals that many face recognition methods inadvertently extract identity-agnostic information such as the pose and emotion.}
    \label{fig:custom_dir_nonidentity}
\end{figure*}

Most interestingly, our method reveals that some directions, such as the pose and emotion of a person, that arguably do not belong to a person's identity can be found for some face recognition models as seen in \cref{fig:custom_dir_nonidentity}. For example, ArcFace~\cite{arcface, gaussian_sampling}, FROM~\cite{from}, and InsightFace~\cite{insightface} seem to (inadvertently) extract pose information as the yaw angle can be controlled somewhat by moving along the corresponding direction in the ID vector latent space. Similarly, the smile appears to be controllable in some small region for all considered ID vectors. Note that the goal of looking at these identity-agnostic directions in the ID vector latent space is not necessarily to control this specific dimension cleanly (this can be achieved with attribute conditioning), but rather to analyze what information is extracted by a given FR method. Thus, our method can be used as a tool to reveal and visualize problems of FR methods that we might not even have been aware of and thus suggest hypotheses for further quantitative experiments.

\clearpage

\end{document}